\documentclass[twoside,11pt]{article}

\usepackage{jmlr2e}
\usepackage{bm}
\usepackage{microtype, amsfonts, color, amsmath}
\usepackage{algorithm}
\usepackage{algorithmic}

\ifx\assump\undefined
\newtheorem{assump}{Assumption}
\fi
\newcommand{\bF}{\bm{F}}
\newcommand{\dist}{\textrm{dist}}

\newtheorem{rmk}{Remark}

\ShortHeadings{Learning state and action representations via tensor decomposition}{Ni, Duan, Dahleh, Wang, and Zhang}
\firstpageno{1}

\begin{document}

\title{Learning Good State and Action Representations for Markov Decision Process via Tensor Decomposition}

\author{\name Chengzhuo Ni \email chengzhuo.ni@princeton.edu \\
       \addr Department of Electrical \& Computer Engineering\\       
       Princeton University
       \AND
	\name Yaqi Duan \email yaqid@mit.edu\\
Laboratory for Information \& Decision Systems\\ 
Massachusetts Institute of Technology
       \AND
\name Munther Dahleh \email dahleh@mit.edu \\
\addr Electrical Engineering \& Computer Science Department\\
Massachusetts Institute of Technology
\AND
\name Mengdi Wang$^1$ \email mengdiw@princeton.edu \\
\addr Department of Electrical \& Computer Engineering\\
Princeton University
\AND       
\name Anru R. Zhang$^1$ \email anru.zhang@duke.edu \\
\addr Departments of Biostatistics \& Bioinformatics, Computer Science, Mathematics, and Statistical Science\\
Duke University
}
\editor{}
\maketitle

\footnotetext[1]{To whom the correspondence should be addressed to.}

\begin{abstract}
	The transition kernel of a continuous-state-action Markov decision process (MDP) admits a natural tensor structure. This paper proposes a tensor-inspired unsupervised learning method to identify meaningful low-dimensional state and action representations from empirical trajectories.  The method exploits the MDP's tensor structure by kernelization, importance sampling and low-Tucker-rank approximation. This method can be further used to cluster states and actions respectively and find the best discrete MDP abstraction. We provide sharp statistical error bounds for tensor concentration and the preservation of diffusion distance after embedding. We further prove that the learned state/action abstractions provide accurate approximations to latent block structures if they exist, enabling function approximation in downstream tasks such as policy evaluation.
\end{abstract}

\def\bP{\mathbf{P}}
\def\bQ{\mathbf{Q}}
\def\bK{\mathbf{K}}
\def\red#1{\textcolor{red}{#1}}
\def\blue#1{\textcolor{blue}{#1}}
\def\cH{\mathcal{H}}
\def\cS{\mathcal{S}}
\def\cA{\mathcal{A}}

\section{Introduction}

State abstraction is a core problem at the heart of control and reinforcement learning (RL). In high-dimension RL, a naive grid discretization of the continuous state space often leads to exponentially many discrete states - an open challenge known as the curse of dimensionality. Having good state representations will significantly improve the efficiency of RL, by enabling the use of function approximation to better generalize knowledge from seen states to unseen states.

We say a state/action representation is ``good", if it enables the use of function approximation to extrapolate and predict future value of unseen states. Suppose there is a representation allowing exact linear parametrization of the transition and value functions, then the sample complexity of RL reduces to depend linearly on $d$ - the representation's dimension \citep{lagoudakis2003least,zanette2019limiting,yang2019sample,jin2019provably}. Even if exact parametrization is not possible, a good representation can be still useful for solving RL with approximation error guarantee (see discussions in \cite{du2019good,lattimore2019learning}). 
An important related problem is to strategically explore in online RL while learning state abstractions \citep{du2019provably,misra2019kinematic}.
Motivated by these advances, we desire methods that can learn good representations, for RL with high-dimensional state and action spaces, automatically from empirical data.

What further complicates the problem is the large action space. An action can be either a one-step decision or a  sequence of multi-step decisions (known as {\it option}). States under different actions lead to very different dynamics. Although states and actions may admit separate low-dimensional structures, they are entangled with each other in sample trajectories.  
This necessitates the tensor approach to decouple actions from states, so that we can learn their abstractions respectively.

\begin{figure*}[t]	
	\centering
	\includegraphics[width = \linewidth]{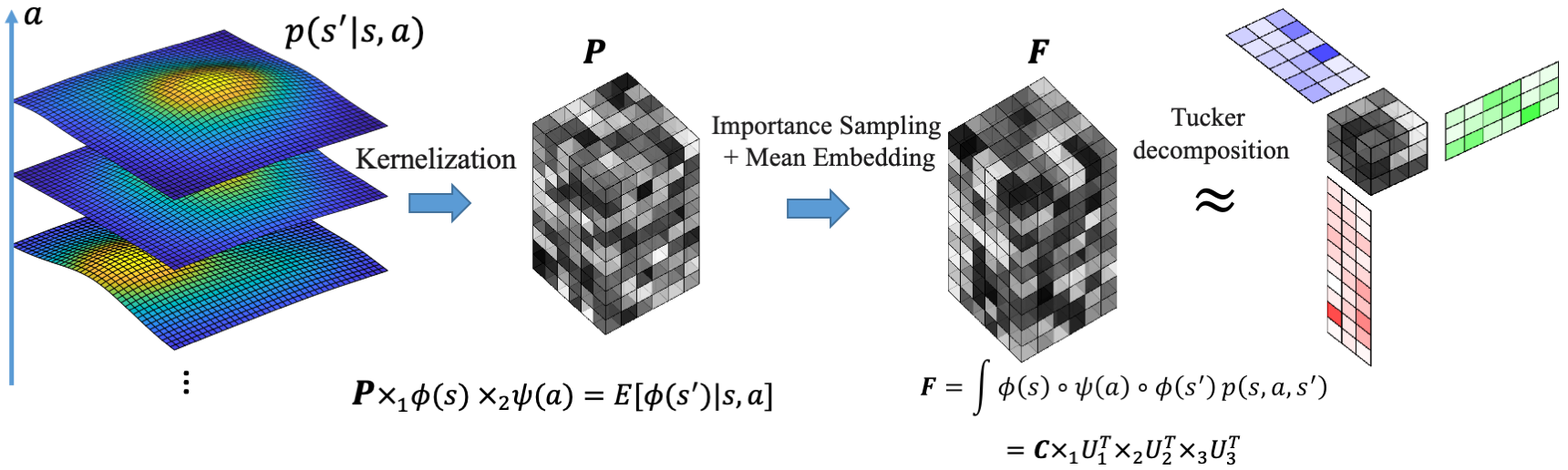}
	\caption{An illustration of our tensor-inspired state and action embedding method}
	\label{fig:illustration-plot}
\end{figure*}
\subsection{Our Approach} 

In this paper, we study the state and action abstraction of Markov decision processes (MDP) from  a tensor decomposition view. We focus on the batch data setting. The Tucker decomposition structure of a transition kernel $p$ provides natural abstractions of the state and action spaces. We illustrate the low-Tucker-rank property in a number of reduced-order MDP models, including the block MDP (i.e., hard aggregation), latent-state model (i.e., soft aggregation). 

Suppose we are given state-action-state transition samples $\mathcal{D}=\{(s,a,s')\}$ from a long sample path generated by a behavior policy. Our objective is to identify a state embedding map and an action embedding map, which map the original state and action spaces (maybe continuous and high-dimensional) into low-dimensional representations, respectively. The embedding maps are desired to be maximally ``predicative", by preserving a notion of kernelized diffusion distance that measures similarity between states in terms of their future dynamics. 

To handle continuous state and action spaces, we use nonparametric function approximation with known kernel functions over the state and action spaces. By approximately decomposing the kernel into finitely many features, we are able to handle the continuous problem by estimating a transition tensor of finite dimensions. Next, we leverage importance sampling and low-rank tensor approximation to identify the desired state and action embedding maps. They yield ``good" representations of states and actions that are useful for linear function approximation in RL. Further, these representations can be used to find the best discrete approximation to the MDP, and in particular, recover the latent structures of block MDP with high accuracy. To the best of knowledge, this paper makes the first attempt to learn low-rank representations for high-dimensional continuous Markov decision, with statistical guarantee. Figure \ref{fig:illustration-plot} illustrates the main idea of our approach. 
\textbf{Contributions} of this paper include:
\begin{itemize}
	\item A tensor-inspired kernelized embedding method to learn low-dimensional state and action representations from empirical trajectories. The method exploits the MDP's tensor structure by importance sampling, mean embedding and low-rank approximation.
	\item Theoretical guarantee that the embedding maps largely preserve the ``predictability" of states and actions in terms of a kernelized diffusion distance, which is proved using a novel tensor concentration analysis.
	\item A discrete state/action abstraction method that provably recovers latent block structures of aggregable MDP.  Theoretical guarantee that the learned abstractions are ``good" representations for approximating transition/value functions within a small error tolerance. 
	\item The numerical studies to corroborate our theoretical findings. The simulation results show the advantage of the proposed method over the baselines of vanilla and top $r$ kernel PCA methods.
\end{itemize}

\subsection{Related Literature}

Spectral and low-rank methods for dimension reduction have a long history. Our approach traces back to the diffusion map approach for manifold learning and graph analysis \citep{lafon2006diffusion}, which comes with a notion of diffusion distance that quantifies similarity between two nodes in a random walk. \cite{coifman2008diffusion} extended the idea to systems driven by stochastic differential equations. \cite{schutte2011coreset} and \cite{klus2016operator,klus2020eigendecompositions} studied how to infer dynamics of a system from leading spectrum of transition operator and find coresets of the state space. 

The statistical theory of low-rank Markov model estimation received attention in recent years. \cite{zhang2019spectral,zhu2022learning} studied the low-rank estimation of finite-state Markov chains. \cite{loffler2021spectral} studied the nonparametric estimation of transition kernel for continuous-state reversible Markov processes with exponentially decaying eigenvalues. \cite{sun2019learning} studied kernelized state embedding and statistical estimation of metastable clusters. These results only apply to Markov processes.

In control theory and RL, state aggregation is a long known approach for reducing the complexity of the state space; see e.g., \cite{moore1991variable,bertsekas1995neuro, singh1995reinforcement, tsitsiklis1996feature, ren2002state}. Representation learning methods were proposed that uses diagonalization or dilation of some Laplacian operator as a surrogate of the transition operator; see e.g. \cite{johns2007constructing, mahadevan2005proto, parr2007analyzing, petrik2007analysis}. See \cite{mahadevan2009learning} for a review. For online RL problems, representation learning approaches have been proposed to find good state-action representations while maintaining a sub-linear regret \citep{modi2021model, agarwal2020flambe, uehara2021representation, zhang2022efficient}.  \citep{ni2022representation} recently applied the representation learning approach to the multi-agent setting. These methods typically require prior knowledge about structures of the problem such as the transition function, or assume access to a finite feature class that covers the ground-truth feature. For tensor-based methods, \cite{mahajan2021tesseract} uses low-rank tensor approximations to model agent interactions in the multi-agent setting. The approach views the Q-function as a tensor whose modes correspond to the action spaces of different agents. \cite{van2021model} considers model-based multi-agent RL and applies low-rank tensor approximation to estimate the transition probabilities and rewards. These approaches only apply to finite state-action MDPs with a low CP rank.

General methods for tensor decomposition and low-rank approximation have been studied in the applied math, statistics, and computer science literature, including the high-order singular value decomposition (HOSVD) \citep{de2000multilinear}, high-order orthogonal iteration (HOOI) \citep{de2000best}, best low-rank approximation \citep{richard2014statistical,zhang2018tensor}, sketched-based algorithms \citep{song2016sublinear}, power iteration, $k$-means power iteration \citep{anandkumar2014guaranteed,sun2017provable}, sparse high-order SVD \citep{zhang2019optimal-statsvd}, generalized tensor decomposition \citep{hong2020generalized,han2022optimal}, etc. The readers are also referred to surveys on tensor decomposition \citep{kolda2009tensor,cichocki2015tensor} and their applications in machine learning \citep{sidiropoulos2017tensor,janzamin2019spectral,panagakis2021tensor}.

\subsection{Markov Decision Process}
An instance of a Markov decision process can be specified by a tuple $\mathcal{M} = (\mathcal{S}, \mathcal{A}, p, r)$, where $\mathcal{S}$ and $\mathcal{A}$ are state and action spaces, $p$ is the transition probability kernel, $r: \mathcal{S}\times\mathcal{A}\rightarrow \mathbb{R}$ is the one-step reward function. At each step $t$, suppose the current state is $s_t$. If the agent chooses an action $a_t$, she will receive an instant reward $r(s_t, a_t)\in [0,1]$ and system's state will transit to $s_{t+1}$ according to the probability distribution $p(\cdot\vert s_t, a_t)$. A policy $\pi$ is a rule for choosing actions based on states, where $\pi(\cdot|s)$ is a probability distribution over $\mathcal{A}$ conditioned on $s\in\mathcal{S}$. Under a given policy, the transition of the MDP will reduce to a Markov chain, whose transition kernel is denoted by $p^\pi$ where $p^\pi(s'|s) = p^{1, \pi}(s'|s) = \int_{\mathcal{A}} \pi(a| s) p(s'|s,a) da$. Based on that, we define the $t$-step transition kernel $p^{t, \pi}(\cdot\vert s)$ inductively by $p^{t, \pi}(\cdot\vert s) = \int p^{t-1, \pi}(s'\vert s)p^{\pi}(\cdot\vert s')ds'$. And we further use $\nu^\pi$ to denote the invariant distribution of that Markov chain. Define the worst-case mixing time \cite[Page 55]{levin2009markov} as 
\begin{align*}
t_{mix} =\max_\pi \min \big\{ t \big| \| p^{t', \pi}(\cdot\vert s_0) - \nu^\pi &\|_{TV} \leq 1/4, \forall s_0 \in \mathcal{S}, t'\geq t  \big\},
\end{align*}
where $\|\cdot\|_{TV}$ denotes the total variation distance.  Throughout the paper, we use $C$ to denote generic constants, while the actual values of $C$ may vary from line to line.

\subsection{Tensor and Tucker Decomposition}

For a general tensor $\bm{X}\in\mathbb{R}^{p_1\times p_2\times\cdots p_N}$, we denote $\bm{X}\times_n \bm{U}$ as the product between $\bm{X}$ and a matrix $\bm{U} \in \mathbb{R}^{q\times p_n}$ on the $n$th mode, which is of size $p_1 \times\ldots\times p_{n-1} \times q \times p_{n+1} \times\ldots\times p_N$. Each element of $\bm{X}\times_n \bm{U}$ is defined as 
$(\bm{X}\times_n \bm{U})_{i_1 \ldots i_{n-1}j i_{n+1} \ldots i_N} = \sum_{i=1}^{p_n}\bm{X}_{i_1 \ldots i_{n-1}i i_{n+1} \ldots i_N}\bm{U}_{ji}.$ We denote by $\mathcal{M}_k(\bm{X})\in\mathbb{R}^{p_k\times \prod_{i\neq k}p_i}$ the factor-$k$ matricization (or flattening) of $\bm{X}$. The Tucker decomposition of $\bm{X}$ is of the form $\bm{X} = \bm{G}\times_1\bm{U}_1\times_2\ldots\times_N\bm{U}_N$, where $\bm{G}\in\mathbb{R}^{q_1\times\ldots \times q_N}$ is a smaller core tensor. In particular, we call the smallest size of $\bm{G}$ the Tucker-rank of $\bm{X}$. Rigorously, we define
$\textrm{Tucker-Rank}(\bm{X}) = (R_1, R_2, \ldots, R_N)$, where 
$R_k = \textrm{Rank}(\mathcal{M}_k(\bm{X}))$. The inner product between two tensors $\bm{X}, \bm{Y}\in\mathbb{R}^{p_1\times p_2\times\cdots p_N}$ is defined as 
\begin{align*}
\langle \bm{X}, \bm{Y}\rangle = \sum_{i_1=1}^{p_1}\sum_{i_2=1}^{p_2}\cdots \sum_{i_N=1}^{p_N} \bm{X}_{i_1i_2\cdots i_N}\bm{Y}_{i_1i_2\cdots i_N}.
\end{align*}
The spectral norm and Frobenius norm of a tensor $\bm{X}\in\mathbb{R}^{p_1\times p_2\times\ldots\times p_N}$ are defined as
\begin{align*}
    \Vert \bm{X}\Vert_{\sigma} &= \sup_{\Vert u_i\Vert = 1, 1\leq i\leq N}\langle \bm{X}, u_1\circ u_2\circ\ldots\circ u_N\rangle, \ \Vert \bm{X}\Vert_F = \sqrt{\langle \bm{X}, \bm{X}\rangle}.
\end{align*}
Suppose $S, A$ are reproducing kernel Hilbert space. We define the Tucker-rank of an operator $\mathbb{P}: S\times A \to S$ as an analogue of Tucker decomposition of tabular tensors: suppose there exist $c_{ijk}\in\mathbb{R}$ and functions $u_i, w_k\in\mathcal{H}_S, v_j\in\mathcal{H}_A, i\in[r], j\in[l], k\in[m]$, such that $(\mathbb{P}f)(s, a) = \sum_{i=1}^r\sum_{j= 1}^l\sum_{k=1}^mc_{ijk}u_i(s)v_j(a)\langle f, w_k\rangle_{\mathcal{H}_S}$. Then write $\textrm{Tucker-Rank}(\mathbb{P})$ as the minimum $(r, l, m)$ that ensure this equation holds.

\section{A Tensor View of Markov Decision Process}

Consider a continuous-state MDP with the transition kernel $p$, where each $p(\cdot|s,a)$ is a conditional transition density function. We adopt a tensor view to exploit structures of $p$ for abstractions of state and action spaces. The Tucker rank of $p$ turns out related to commonly used reduced-order models such as state aggregation and latent models. We handle the continuous state and action spaces using kernel function approximation. Suppose we have a Reproducing Kernel Hilbert Space (RKHS) $\cH_S$ for functions over states and a RKHS $\cH_A$ for functions over actions. We make the assumption that the MDP's transition kernel $p$ can be represented in these function spaces.

\begin{assump}\label{assump1} Let $\mathbb{P}$  be the transition operator of $p$, i.e., $(\mathbb{P}f)(s,a) = \int p(s'\vert s,a) f(s') ds' $. Assume that $\textrm{Tucker-Rank}(\mathbb{P})\leq(r, l, m)$ $^1$, and $\mathbb{P} f \in \cH_S\times \cH_A, \forall f\in\cH_S.$
\end{assump}
Here, the low-Tucker rankness assumption captures the structure that state/action space can be compressed into a lower-dimensional space while preserving the dynamics. This assumption naturally holds in many well-known reinforcement learning models, such as soft state aggregation \citep{singh1995reinforcement,bertsekas1995dynamic,sutton1998reinforcement}, rich-observation MDP \citep{azizzadenesheli2016reinforcement,du2019provably}, contextual MDP \citep{jiang2017contextual}, linear/factor MDP \citep{jin2019provably}, kernel MDP \citep{ormoneit2002kernel,chowdhury2019online}.


In the remainder of the paper, we assume without loss of generality that the state and action kernel spaces admit finitely many known basis functions, which we refer to as state features $\phi(s)\in\mathbb{R}^{d_S}$ and action features $\psi(a)\in\mathbb{R}^{d_A}$. This is a rather mild assumption: Even if we do not know the basis function but are only given kernel functions $K_S$ and $K_A$ for $\mathcal{H}_S$ and $\mathcal{H}_A$. According to \cite{rahimi2008random}, one can generate finitely many random features to approximately span these kernel spaces such that $K_S(s,s') \approx\sum^{d_S}_{i=1}\phi_i(s)^{\top}\phi_i(s')$ and $K_A(a,a') \approx\sum^{d_A}_{i=1}\psi_i(a)^{\top}\psi_i(a')$. Also note that our approach applies to {\it arbitrary} state and action spaces, as long as they come with appropriate kernel functions. Although $p$ is infinitely dimensional, we use the given kernel spaces and represent $p$ with a finite-dimensional tensor. In particular, Assumption \ref{assump1} implies the following tensor linear model:
\begin{lemma}[Conditional transition tensor and linear model]\label{lm1} Suppose Assumption \ref{assump1}  holds. There exists a tensor $\bP\in\mathbb{R}^{d_S\times d_A\times d_S}$ such that $\textrm{Tucker-Rank}(\bm{P})\leq(r, l, m)$ and
$$\bP \times_1\phi(s)^\top \times_2 \psi(a)^\top = \mathbb{E}[ \phi(s')|s,a], \qquad \forall s \in S, a\in A. $$
\end{lemma}
Tucker decomposition is one of the most general low-rank structure for tensors. 
Remarkably, the low-Tucker-rank property (Assumption \ref{assump1}) turns out to be universal in a number of reduced-order MDP models. Typical examples include block MDP \citep{du2019provably} and soft MDP aggregation \citep{singh1995reinforcement}, whose detailed descriptions are placed in Appendix B. The low-Tucker-rank property also holds in MDPs with rich observations \citep{azizzadenesheli2016reinforcement}, and is related to the Bellman rank \citep{jiang2017contextual}. We remark that the tensor rank is determined solely by the transition model $p$ (i.e., the environment), regardless of the reward $r$.

\section{Tensor-Inspired State and Action Embedding Learning}

In this section, we develop a tensor-inspired representation learning method, which embeds states and actions into decoupled low-dimensional spaces. Next, we will develop the method step by step, and provide theoretical guarantees. 

\subsection{Tensor MDP Mean Embedding by Importance Sampling}
Suppose we have a batch dataset of state-action samples. 
\begin{assump}\label{assump-data}
The data $\mathcal{D}=\{(s,a,s')\}$ consists of state-action-state transitions from a single sample path generated by a known behavior policy $\bar \pi$.
\end{assump}
Let $\xi$ be the stationary state distribution of the sample path under policy $\bar\pi$. Let $\eta$ be a positive probability measure over the action space. Consider the tensor mean embedding 
$$\bF = \int \phi(s) \circ \psi(a)\circ \phi(s') p(s,a,s') ds dads' \in\mathbb{R}^{d_S\times d_A\times d_S},$$
where $p(s, a, s') = p(s'\vert s, a)\xi(s)\eta(a)$. 

\begin{lemma}\label{lm2}
Assumption 1 implies $\textrm{Tucker-Rank}(\bm{F})\leq (r, l, m).$
\end{lemma}

We estimate the mean embedding tensor $\bm{F}$ by importance sampling: 
\begin{equation}\label{eq-is}
\bar{\bm{F}} = n^{-1}\textstyle\sum_{i=1}^n\frac{\eta(a_i)}{\bar\pi(a_i\vert s_i)}\cdot \phi(s_i)\circ \psi(a_i)\circ\phi(s_i').
\end{equation} 
The mean embedding tensor $\bm{F}$ is related to the transition tensor $\bm{P}$ through a simple relation.
\begin{lemma}[Relation between $\bm{P}$ and $\bm{F}$]\label{relation}
When $\{\psi_i(\cdot)\}_{i=1}^{d_A}$ forms a set of orthogonal basis with respect to $L^2(\eta)$, we have $\bm{P} = \bm{F} \times_1 \bm{\Sigma}^{-1},$ where $\bm{\Sigma} =\int\xi(s) \phi(s)\phi(s)^{\top} ds$.
\end{lemma}

\paragraph{Necessity of importance sampling.}
The importance sampling step \eqref{eq-is} is necessary to decouple states from actions. Without importance sampling, the naive mean embedding tensor 
$$\bm{W}:= \int \phi(s) \circ \psi(a)\circ \phi(s') \xi(s)\bar{\pi}(a|s)p(s'|a,s) ds dads'$$ 
may have large ranks on the first two dimensions. This is due to that the behavior policy $\bar\pi$ couples the state and action spaces together, therefore their independent low-dimensional structures are lost in the mean embedding tensor $\bm{W}$. Without using importance sampling, if we replace $\bm{F}$ with plain mean $\bm{W}$, Lemma \ref{lm2} and Lemma \ref{relation} no longer hold. As a result, one cannot learn the best low-dimensional structure of $p$ from $\bm{W}$.

\subsection{Low-Rank Estimation of Transition Tensor}

We estimate a low-rank approximation to $\bm{F} $ by solving:
\begin{equation}\label{eq:spectral-norm-min}
\hat{\bm{F}}= \textrm{argmin } \Vert \bm{Q} - \bar{\bm{F}}\Vert_\sigma, \textrm{ subject to } \textrm{Tucker-Rank}(\bm{Q}) \leq (r, l, m)
\end{equation}
and estimate the transition operator ${\bm{P}}$ by $\hat{\bm{P}} = \hat{\bm{F}}\times_1 \hat{\bm{\Sigma}}^{-1}$, where $\hat{\bm{\Sigma}} = \frac{1}{n}\sum_{i=1}^n\phi(s_i)\phi^\top(s_i).$
Define
\begin{align*}
&K_{max} = \max\left\{\sup_{s}K_S(s, s), \sup_{a}K_A(a, a)\right\},\\
&\bar{\mu} = \Vert \mathbb{E}[K_1(S, S)\phi(S)\phi(S)^\top]\Vert_\sigma,\\
&\kappa = \sup_{s\in\mathcal{S}, a\in\mathcal{A}}\frac{\eta(a)}{\pi(a\vert s)}, \\
&\bar{\lambda} = \sup_{\bm{u}, \bm{v}, \bm{w}} \mathbb{E}_{\xi\circ \eta \circ  p(\cdot\vert \cdot)}[[(\bm{u}^\top\phi(S) )(\bm{v}^\top\psi(A))(\bm{w}^\top\phi(S'))]^2],\textrm{where } \bm{u},\bm{w}\in S^{d_S-1}, \bm{v}\in S^{d_A-1}.
\end{align*}
Here, $K_S, K_A$ are the kernels associated with the state RKHS space $\cH_S$ and action RKHS space $\cH_A$, respectively.
\begin{theorem}[Low-rank estimation of the transition tensor $\bm{P}$]\label{main_thm2}
Suppose Assumptions \ref{assump1}-\ref{assump-data} hold.
Suppose $\psi$ is orthonormal with respect to $L^2(\eta)$, and
\begin{align*}
\frac{n/ t_{mix}}{(\log(n/t_{mix}))^2}&\geq 1024\bigg(\Vert\bm{\Sigma^{-1}}\Vert^2_\sigma \bar{\mu} + \frac{K_{max}^2}{\bar{\mu}} + \frac{\kappa K_{max}^3}{\bar{\lambda}}\bigg)\bigg(\log\frac{2t_{mix}}{\delta} + 8(d_S + d_A)\bigg), 
\end{align*}
then with probability $1-\delta$, we have
\begin{align*}
&\Vert \bm{P}-\hat{\bm{P}}\Vert_\sigma\leq 256 \Vert\bm{\Sigma}^{-1}\Vert_\sigma\sqrt{\frac{\bar{\lambda}(\log\frac{2t_{mix}}{\delta} + d_S + d_A)(\kappa + \bar{\mu}\Vert\bm{\Sigma}^{-1}\Vert_\sigma^2)}{(n/t_{mix})\log^{-2}(n/t_{mix})}}. 
\end{align*}
\end{theorem}

The derivation of $\hat{\bm{P}}$ also provides a tractable way to estimate $\mathbb{E}[\phi(s')\vert s, a]$ by $$\hat{\mathbb{E}}[\phi(s')\vert s, a] := \hat{\bm{P}}\times_1 \phi(s)^\top\times_2\psi(a)^\top.$$ And we have the following guarantee on the estimation error:
$$
\Vert\hat{\mathbb{E}}[\phi(s')\vert s, a] - \mathbb{E}[\phi(s')\vert s, a] \Vert \leq K_{max}\Vert \bm{\hat{P}} - \bm{P}\Vert_\sigma. 
$$

	\paragraph{Rank selection} Our theory assumes the prior knowledge of tensor rank.
		In practice, it is common to tune the rank parameters by checking the elbow in the scree plot and using cross validation (see discussions in the classical literature on PCA, e.g., \cite{jolliffe1986principal}). In theory, rank estimation is hard unless one makes additional strong assumptions, like that the eigengap is bounded from below. 
\paragraph{Computation} Finding the exact optimum of \eqref{eq:spectral-norm-min} can be computationally intense in general \cite{de2008tensor}. In practice, we can apply classic tensor decomposition algorithms, such as higher-order orthogonal iteration (HOOI) \citep{de2000best}, high-order SVD \citep{de2000multilinear}, sequential-HOSVD \citep{vannieuwenhoven2012new}, gradient descent \citep{han2022optimal}, to find an approximate solution to \eqref{eq:spectral-norm-min}. In particular, the statistical optimality of tensor power iterations, e.g., HOOI and HOSVD (Appendix A), have been justified in some special cases \cite{zhang2018tensor}. We expect these approximations also work for our problems, which is later validated in our experiment. 

\subsection{Learning State and Action Embeddings}

Next, we show how to embed states and actions to low-dimensional representations to be maximally ``predictive." Consider a kernelized diffusion distance of the MDP, which measures similarity in terms of future dynamics restricted to a function class:
\begin{align*}
\dist[(s_1, a_1), (s_2, a_2)] 
= \sup_{\|f\|_{\cH_S}\leq 1 } | &\mathbb{E}[f(s') |s_1,a_1] -  \mathbb{E}[f(s') |s_2,a_2] | .
\end{align*}
This distance quantifies how well one can generalize the predicted value at a seen state-action pair $(s,a)$ to a new $(s',a')$. Under the low-tensor-rank assumption, we have $\bm{P} = \bm{C} \times_1 \bm{U}_1\times_2 \bm{U}_2 \times_3 \bm{U}_3$, where $\bm{U}_1, \bm{U}_2, \bm{U}_3$ are columnwisely orthonormal matrices. Then we can define the {\it kernelized state diffusion map}, {\it kernelized action diffusion map}  and their joint map as
\begin{align*}
&f(\cdot) := \bm{U}_1^{\top} \phi(\cdot),\ \   g(\cdot) := \bm{U}_2^{\top}\psi(\cdot),\ \Phi(s,a) := \bm{C}\times_1 f(s)^\top\times_2 g(a)^\top,
\end{align*}
respectively. It follows that $\dist[(s, a), (s', a')] = \Vert \Phi(s,a) - \Phi(s',a')\Vert$, if $\phi$ is a collection of orthonormal basis functions of $\cH_S$. Motivated by the preceding analysis, we propose to estimate state and action embedding maps based on the tensor estimator. 
After we obtain $\hat{\bm{P}}$, we can simply find the corresponding state and action embedding maps from factors of its Tucker decomposition
\begin{align*}
\hat{\bm{P}} = \hat{\bm{C}}\times_1 \hat{\bm{U}}_1 \times_2 \hat{\bm{U}}_2 \times_3 \hat{\bm{U}}_3,
\end{align*}
where we require that $\hat{\bm{U}}_k, k=1,2,3$ are column-wisely orthonormal. The full procedure is given in Algorithm \ref{alg1}.
\begin{algorithm}[htb!]
\begin{algorithmic}[1]
\caption{Learning State and Action Embedding Maps \label{alg1}}
\STATE \textbf{Input}: $\{(s_i, a_i, s_i')\}_{i=1}^n, (r, l, m)$
\STATE Calculate $$\bar{\bm{F}} = \frac{1}{n}\sum_{i=1}^n\frac{\eta(a_i)}{\pi(a_i\vert s_i)}\phi(s_i)\circ \psi(a_i)\circ\phi(s_i'),$$ and get $\hat{\bm{F}}$ as the low-rank approximation of $\bar{\bm{F}}$ using \eqref{eq:spectral-norm-min}
\STATE Calculate $\hat{\bm{\Sigma}} = \frac{1}{n}\sum_{i=1}^n\phi(s_i)\phi^\top(s_i)$, $\hat{\bm{P}} = \hat{\bm{F}}\times_1 \hat{\bm{\Sigma}}^{-1}$
\STATE Let $\bm{P}_1 = \hat{\bm{P}}$. For $k=1,2,3$, derive $\hat{\bm{U}}_k$ from the SVD $$\mathcal{M}_k(\bm{P}_k) = \hat{\bm{U}}_k \bm{\Lambda}_k \bm{V}_k^\top,$$ and let $\bm{P}_{k+1} = \bm{P}_k\times_k \hat{\bm{U}}_k$. 
\STATE \textbf{Output}: \\
State and action embedding maps $\hat{f}: s\mapsto \hat{\bm{U}}_1^\top \phi(s), \hat{g}: a \mapsto \hat{\bm{U}}_2^\top \psi(a)$; \\ Core transition tensor $\hat{\bm{C}} = \bm{P}_4$. 
\end{algorithmic}
\end{algorithm}
Now we have obtained the state embedding map $\hat f$ and the action embedding map $\hat{g}$. Accordingly, we define the joint state-action embedding and the empirical embedding distance as
$$\hat \Phi(s,a) =\hat{\bm{C}}\times_1\hat f(s)^\top\times_2 \hat g(a)^\top, \ \widehat{\dist}[(s, a), (s', a')] = \Vert\hat \Phi(s,a)  - \hat \Phi(s',a')\Vert. $$ 

\begin{theorem}[Embedding error bound]\label{dist} 
Let Assumptions \ref{assump1}-\ref{assump-data} hold.
Suppose $\phi$ is an orthogonal basis of $\mathcal{H}_S$, and $\psi$ is orthogonal w.r.t $L^2(\eta)$, then we can find an orthogonal matrix $\bm{O}$, such that 
\begin{align*}
&\|\hat \Phi(s,a) - \bm{O}\Phi(s,a)\| \leq \epsilon,\\
&\vert\widehat{\dist}[(s, a), (s', a')] - \dist[(s, a), (s', a')] \vert
\leq 2\epsilon, \forall s, a, s', a'
\end{align*}
where $\epsilon$ is controlled by the low-rank estimation error,
$\epsilon:=K_{max}\left(1 + \frac{\sqrt{2}\Vert\bm{P}\Vert_\sigma}{\sigma}\right)\Vert\hat{\bm{P}}-\bm{P}\Vert_\sigma,$
and
$\sigma: = \sup_{\Vert w\Vert \leq 1}\sigma_m(\bm{P}\times_1 w^\top),$
where $\sigma_m$ denotes the $m$-th singular value of a matrix.
\end{theorem}

\paragraph{Advantage of tensor method.} As an alternative, one could ignore the tensor structure and treat the state and action jointly, yielding a low-dimensional representation for the pair $(s,a)$ directly. 
This approach may be favorable if the $(s,a)$ has a very simple joint structure.  However, the tensor approach may be significantly more sample efficient if $s$ and $a$ admit {\it separate} low-dimensional structures. 
To see this, suppose the state and action features have dimensions $d_S$ and $d_A$ before embedding. Also assume the Tucker rank is $r=l=m$ for simplicity. By treating $(s,a)$ jointly and ignoring the tensor structure, one would need $\tilde{\Omega}(d_Ad_Sr)$ samples to reliably recover the low-dimensional structure. In comparison, our tensor-based approach requires only $\tilde{\Omega}((d_X+d_A)r)$ samples. 

\section{Estimating the Optimal Discrete MDP Abstraction}
Next, we study how to provably reduce a continuous-state continuous-action MDP into a discrete one, by an application of the learned kernelized diffusion distance to partition the state and action spaces. 

\subsection{Optimal Partition of State and Action Spaces}
Our goal is to learn an optimal discretization of a continuous MDP. Specifically, we want to find a partition of $\mathcal{S}$ and $\mathcal{A}$, denoted as blocks $A_i, B_j, i\in [n_s],  j\in [n_a]$ and a collection of probability transition distributions $\{q_{ij}(\cdot)\}$ on the blocks. For each state-action pair $(s, a)\in A_i\times B_j$, and some function $f\in \mathcal{H}_S$, we want to approximate the one-step expected value $(\mathbb{P}f)(s, a) = \int p(s'\vert s, a)f(s')ds'$ by 
$$(\mathbb{P}f)(s, a) = \int p(s'\vert s, a)f(s')ds'\approx \int q_{ij}(s')f(s')ds'.$$

We formalize the {\it optimal state-action partition problem} as: 
\begin{equation}
\label{eq-partition}
\min_{\{A_i, B_j, q_{ij}\}} L(\{A_i, B_j, q_{ij}\}):= \sum_{i, j}\int_{A_i\times B_j} \xi(s)\eta(a)\Vert  p(\cdot\vert s, a) - q_{ij}(\cdot)\|_{\mathcal{H}_S}^2 dsda,
\end{equation}
whose solution is denoted as $\{A_i^*, B_j^*, q_{ij}^*\}$, and the corresponding optimal value is denoted by $L^*$. 

In particular, if $|\mathcal{A}|=1$, the MDP reduces to a Markov process 
and the optimization problem reduces to $\min_{A_i}\min_{\{q_{i}\}} \sum_{i=1}^{n_s}\int_{A_i} \xi(s)\Vert  p(\cdot\vert s) - q_{i}(\cdot)\|_{\mathcal{H}_S}^2 dsda$, which becomes equivalent to the metastable state partition problem for random walk and dynamic systems \cite{e2008optimal}. 



\subsection{Decoupled State and Action Clustering}
Next, consider the RL setting where we wish to learn $\{A_i^*,B_j^*\}$ when $p$ is unknown. Observe that the optimal partition is determined solely by the kernelized diffusion distance equipped by the state-action space. This allows the approximation of \eqref{eq-partition} by the {\it empirical state-action clustering problem}:
\begin{equation}\label{prob}
\begin{split}
\min_{\{A_i, B_j,\bm{z}_{ij}\}} \sum_{i, j}
& \int_{A_i\times B_j} \xi(s)\eta(a)
 \cdot \Vert \hat\Phi(s,a) - \bm{z}_{ij}\Vert^2 dsda,
\end{split}
\end{equation}
whose solution is denoted by $\{\hat{A}_i, \hat{B}_j, \hat{\bm{z}}_{ij}\}$. Then the corresponding discrete transition distribution from state abstraction $i$ and action abstraction $j$ takes the form $\hat{q}_{ij}(\cdot) = \hat{\bm{z}}_{ij}^\top\hat{\bm{U}}_3^\top\phi(\cdot)$. 

To facilitate computation, we provide a relaxation of problem \eqref{prob} that can be solved using $k$-means-type algorithms.
By taking
${\bm{z}}_{ij} = \hat{\bm{C}}\times_1 f_i^\top\times_2 g_j^\top$ for some $f_i,g_j$, the partition problem becomes 
\begin{align*}
\min_{\{A_i, B_j\}}\min_{\{f_i, g_j\}} &\sum_{i, j}\int_{A_i\times B_j}  \xi(s)\eta(a)  \cdot \Vert \hat{\bm{C}}\times_1 \hat{f}(s)^\top \times_2\hat{g}(a)^\top - \hat{\bm{C}}\times_1 f_i^\top \times_2 g_j^\top \Vert^2 dsda.
\end{align*}
Using the relation
$$
\Vert \hat{\bm{C}}\times_1 \hat{f}(s)^\top \times_2\hat{g}(a)^\top - \hat{\bm{C}}\times_1 f_i^\top \times_2 g_j^\top \Vert^2 \leq 2\Vert \hat{\bm{C}}\Vert_\sigma^2 K_{max}(\Vert \hat{f}(s) - f_i\Vert^2 + \Vert \hat{g}(a) - g_j\Vert^2),
$$
we can relax the problem \eqref{prob} into two simpler subproblems:
\begin{align*}
\min_{A_i}\min_{f_i}\sum_i \int_{A_i}\xi(s)\Vert \hat{f}(s) - f_i\Vert^2 ds; \qquad
\min_{B_j}\min_{g_j}\sum_j \int_{B_j}\eta(a)\Vert \hat{g}(a) - g_j\Vert^2 da.
\end{align*}
In short, one can efficiently compute the decoupled state and action clusters using the learned representations from the tensor method. The full procedure is given in Alg.\ \ref{alg-cluster}.

\begin{algorithm}[htb!]
\begin{algorithmic}[1]
\caption{Learning Optimal State-Action Abstractions\label{alg2}}
\label{alg-cluster}
\STATE \textbf{Input}: $\{(s_i, a_i, s_i')\}_{i=1}^n, (r, l, m)$
\STATE Estimate the state embedding and action embedding maps $\hat{f}, \hat{g}, \hat{\bm{U}}_3$ using Algorithm \ref{alg1}.
\STATE Apply k-means to solve the following two problems, respectively:
\begin{align*}
&\min_{\{A_i\}}\min_{\{f_i\}}\sum_i \int_{A_i}\xi(s)\Vert \hat{f}(s) - f_i\Vert^2 ds,\\ &\min_{\{B_j\}}\min_{\{g_j\}}\sum_j \int_{B_j}\eta(a)\Vert \hat{g}(a) - g_j\Vert^2 da.
\end{align*}
\STATE \textbf{Output}: The state partition $\{\hat A_j\}$ and action partition $\{\hat B_j\}$
\end{algorithmic}
\end{algorithm}

\subsection{Theoretical Guarantee}
The following theorem guarantees that the empirical discretion is not far from the groundtruth. 

\begin{theorem}[Mean squared clustering error]\label{discrete}
Let Assumptions \ref{assump1}-\ref{assump-data} hold.
Suppose $\psi$ is a orthonormal basis with respect to $L^2(\eta)$ and $n$ sufficiently large, 
then with probability at least $1-\delta$, we have
\begin{align*}
	\begin{split}
L(\{\hat{A}_i, \hat{B}_j, q_{ij}\}) \leq &  C\frac{\Vert\bm{\Sigma}^{-1}\Vert_\sigma rlm(1 + \frac{2\bar{\lambda}\Vert \bm{\Sigma}^{-1}\Vert_\sigma^2}{\sigma^2})}{\max\{r, l, m\}} \cdot\frac{\bar{\lambda}(\log(2t_{mix}/\delta) + d_S + d_A)(\kappa + \bar{\mu}\Vert\bm{\Sigma}^{-1}\Vert_\sigma^2)}{(n/t_{mix})\log^{-2}(n/t_{mix})} \\
& + 4L^*,
\end{split}
\end{align*}
where $L^*$ is the optimal value of problem \eqref{eq-partition}, $\sigma$ is defined as in Theorem \ref{dist}, $C$ is an absolute constant.
\end{theorem}

Next, we focus on the case where the true MDP has latent block structures. 
\begin{assump}\label{block} Let there be blocks on the state and action spaces $A_i, B_j,  i\in [n_s],  j\in [n_a]$, i.e., 
$$p(\cdot\vert s, a) = \sum_{i=1}^{n_s}\sum_{j=1}^{n_a}q^*_{ij}(\cdot)\bm{1}_{s\in A_i}\bm{1}_{a\in B_j}$$ for some probability density functions $q_{ij}$. 
\end{assump}
Suppose we have applied Algorithm \ref{alg-cluster} to recover the latent blocks.
Let $\{\hat{A}_i\}_{i=1}^{n_s}, \{\hat{B}_j\}_{j=1}^{n_a}$ be the estimated state and action clusters.  Define the misclassification error as
\begin{align*}
M(\{\hat{A}_i\}, \{\hat{B}_j\}) = \min_{\sigma_1,\sigma_2}
\sum_{i=1}^{n_s}\sum_{j=1}^{n_a}\frac{(\xi\times\eta)((A_i\times B_j)\setminus (\hat{A}_{\sigma_1(i)}\times \hat{B}_{\sigma_2(j)}))}{(\xi\times\eta)(A_i\times B_j)},
\end{align*}
where $\sigma_1$ and $\sigma_2$ are permutations over the state and action blocks, respectively.
We prove the following clustering error bound:
\begin{theorem}[Misclassification error for block MDP]\label{partition}
Let Assumptions \ref{assump1}, \ref{assump-data}, \ref{block} hold. 
For $n$ sufficiently large, with probability $1-\delta$, we have
\begin{align*}
M(\{\hat{A}_i\}, \{\hat{B}_j\}&) \leq  C\Vert\bm{\Sigma}^{-1}\Vert_\sigma\frac{rlm(1 + \frac{2\bar{\lambda}\Vert \bm{\Sigma}^{-1}\Vert_\sigma^2}{\sigma^2})}{\max\{r, l, m\}} \cdot \frac{\bar{\lambda}(\log(2t_{mix}/\delta) + d_S + d_A)(\kappa + \bar{\mu}\Vert\bm{\Sigma}^{-1}\Vert_\sigma^2)}{\Delta_1^2(n/t_{mix})(\log\frac{n}{t_{mix}})^{-2}},
\end{align*}
where $\Delta_1^2 := \min_{i,j} \min_{(k, l)\neq (i, j)}\xi(A_i)\eta(B_j)\Vert q_{ij}^*(\cdot) - q^*_{kl}(\cdot)\Vert_{\mathcal{H}_S}^2$ and $C$ is an absolute constant. 
\end{theorem}
\begin{rmk}
	
	The bounds in Theorems \ref{discrete} and \ref{partition} grow proportionally to $rlm$ (i.e., the product of Tucker ranks of $\bP$), which can be large even when $r,l,m$ are individually small. However, the term is essential as it represents the degree of freedom of a Tucker rank $(r, l, m)$ tensor, which is given by $p(r+l+m) +rlm$ \cite[Proposition 1]{zhang2019cross}.
\end{rmk}

Next we investigate how the statistical inaccuracy of state abstraction would affect downstream RL tasks. We consider block-structured MDP whose transition kernel $p$, reward $r$ and policy $\pi$ are defined on state and action blocks $\{A_i\}, \{B_j\}$.
We use $M =  (p,\{r_h\}_{h=1}^H)$ to denote such an MDP instance, and use $\mathcal{M}$ to denote the collection of all such $M$. One may wonder if the state abstraction error would blow up, particularly if we want to evaluate a multi-step cumulative return. 
Define the {\it $H$-step state abstraction error} as the worst-case policy evaluation error over horizon $H$, given by
\begin{align*}
&D(\{\hat{A}_i\}, \{\hat{B}_j\}) = \sup_{M\in\mathcal{M}, \pi}\inf_{\hat{p}} \left\vert\mathbb{E}^\pi_p\left[ \sum_{h=1}^H r_h(S_h, A_h)\right] - {\mathbb{E}}^\pi_{\hat p}\left[\sum_{h=1}^H r_h(S_h, A_h)\right] \right \vert,
\end{align*}
where the supremum is taken over all block-structured MDP instances and policies on $\{A_i,B_j\}$, and the infimum is to find a best-fit transition model on the estimated clusters $\{\hat A_i,\hat B_j\}$, of the form
\begin{align*}
\hat{p}(s'\vert s, a) = \textstyle\sum_{i, k=1}^{n_s}\textstyle\sum_{j=1}^{n_a} \frac{\hat{q}(k\vert i, j)}{\xi(\hat{A}_k)}\bm{1}_{s\in \hat{A}_i}\bm{1}_{a\in \hat{B}_j}\bm{1}_{s'\in \hat{A}_k},
\end{align*}
where $\hat{q}$ is a set of discrete transition probabilities. 

\begin{theorem}[Policy evaluation error due to inaccurate state abstraction]\label{partition2}
Let Assumptions \ref{assump1}, \ref{assump-data}, \ref{block} hold. 
Then for $n$ sufficiently large, with probability $1-\delta$, we have
\begin{align*}
D(\{\hat{A}_i\}, \{\hat{B}_j\})\leq 4 \bar{c}\underline{c}^{-1}H^2 M(\{\hat{A}_i\}, \{\hat{B}_j\}) 
\end{align*} 
where
$\underline{c} = \min_{i,j}(\xi\times\eta)(A_i\times B_j), \bar c = \max_{i,j}(\xi\times\eta)(A_i\times B_j)$.
\end{theorem}
Theorem \ref{partition2} shows that the $H$-step state abstraction error grows at most quadratically with $H$, not exponentially. In other words, inaccuracy in state abstraction does not suffer from the curse of horizon. Thus the learned state and action abstractions are useful for approximate policy evaluation. 

\section{Numerical Experiment}

\begin{figure}[htb!]
\center
\includegraphics[width=.3\linewidth]{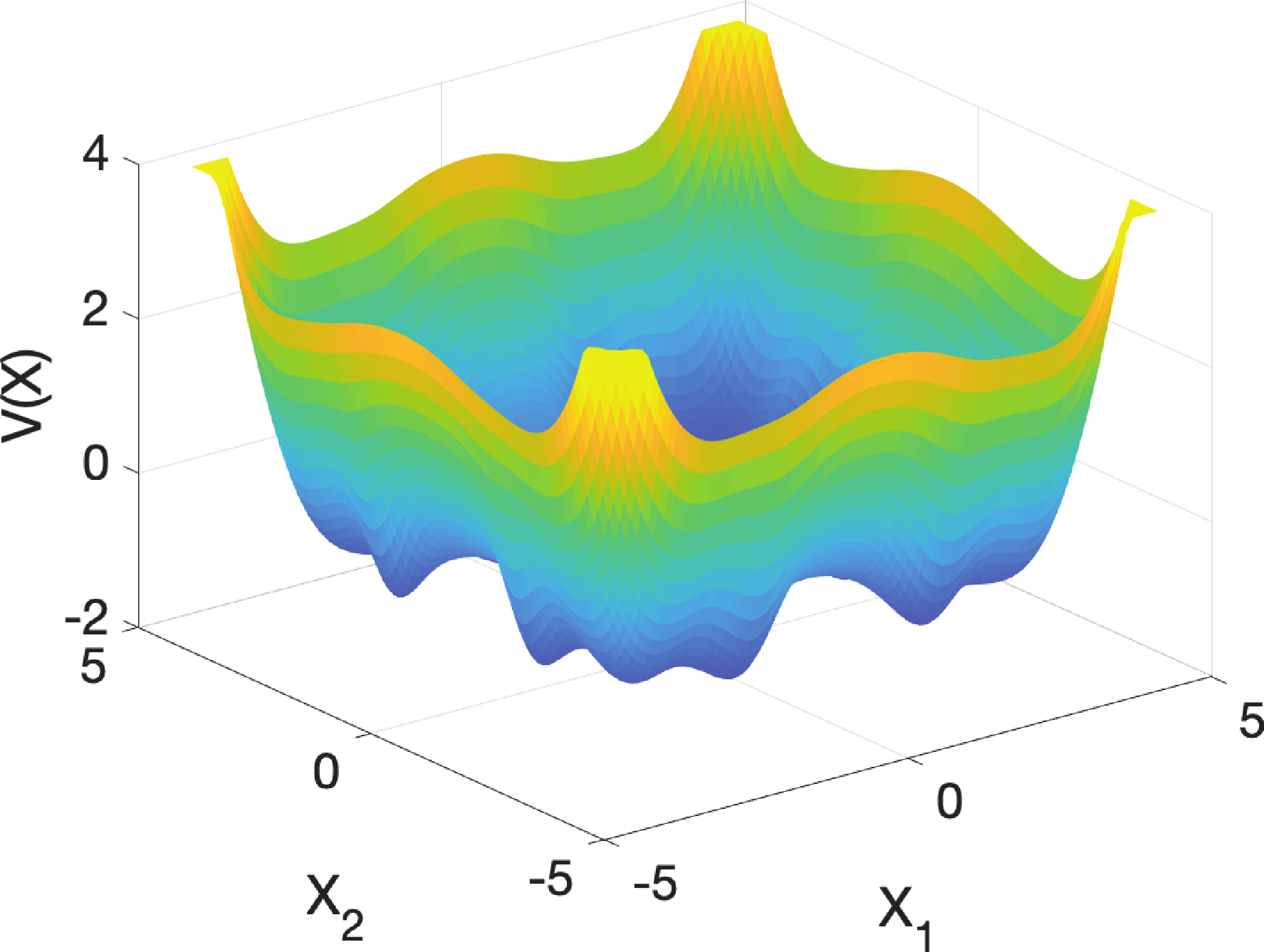}\quad \includegraphics[width=.3\linewidth]{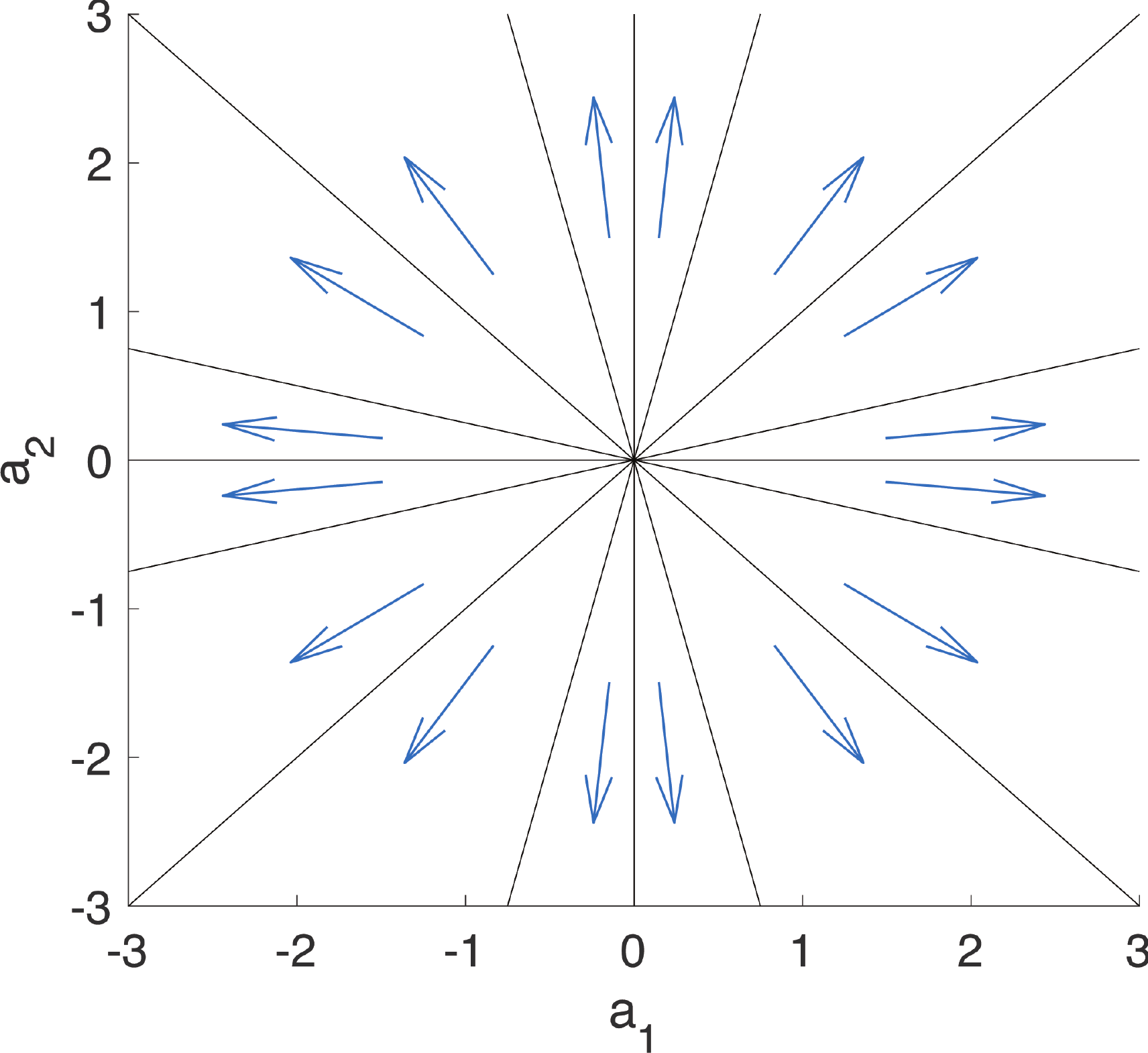}\\
\includegraphics[width=.24\linewidth]{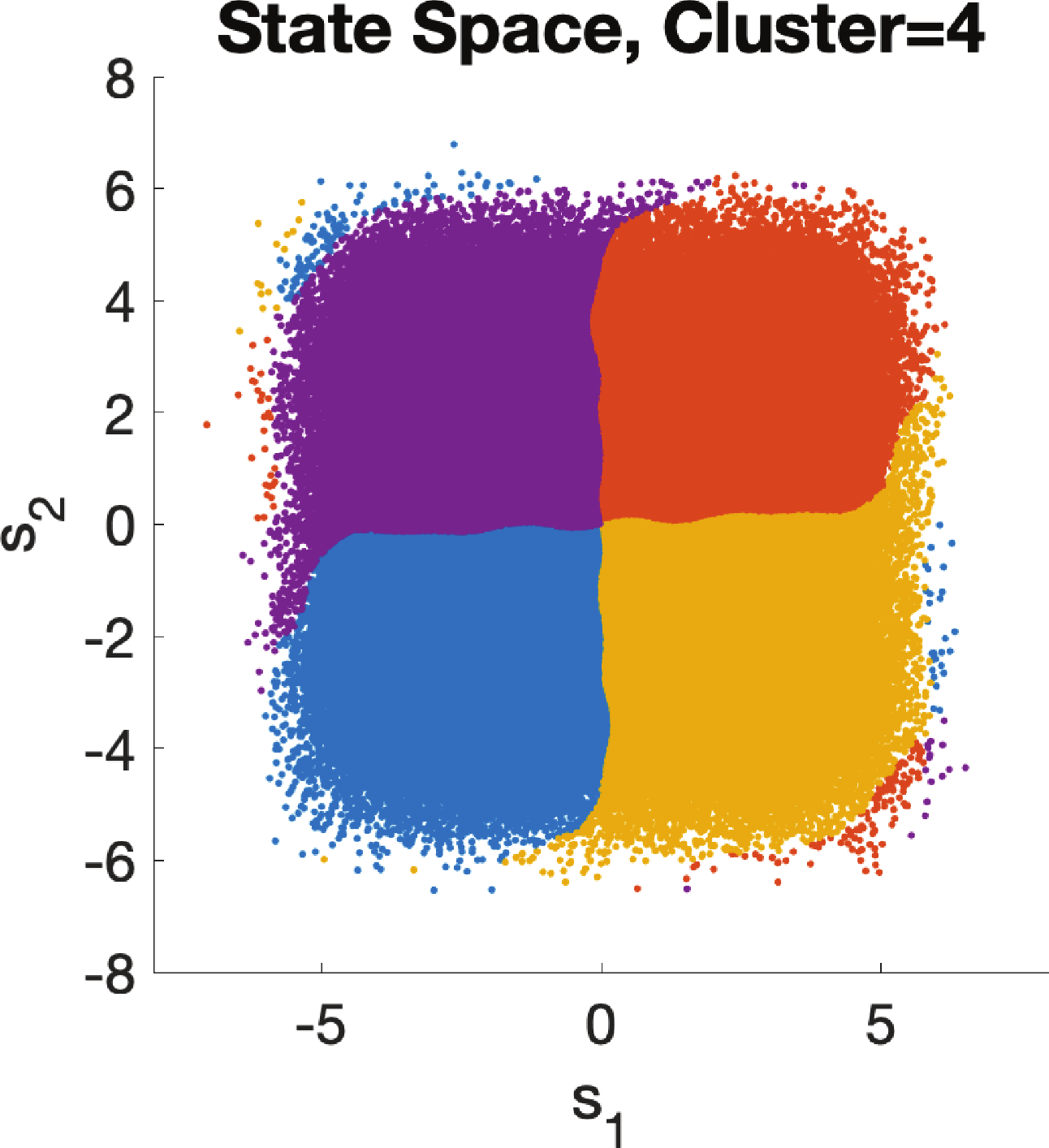}
\includegraphics[width=.24\linewidth]{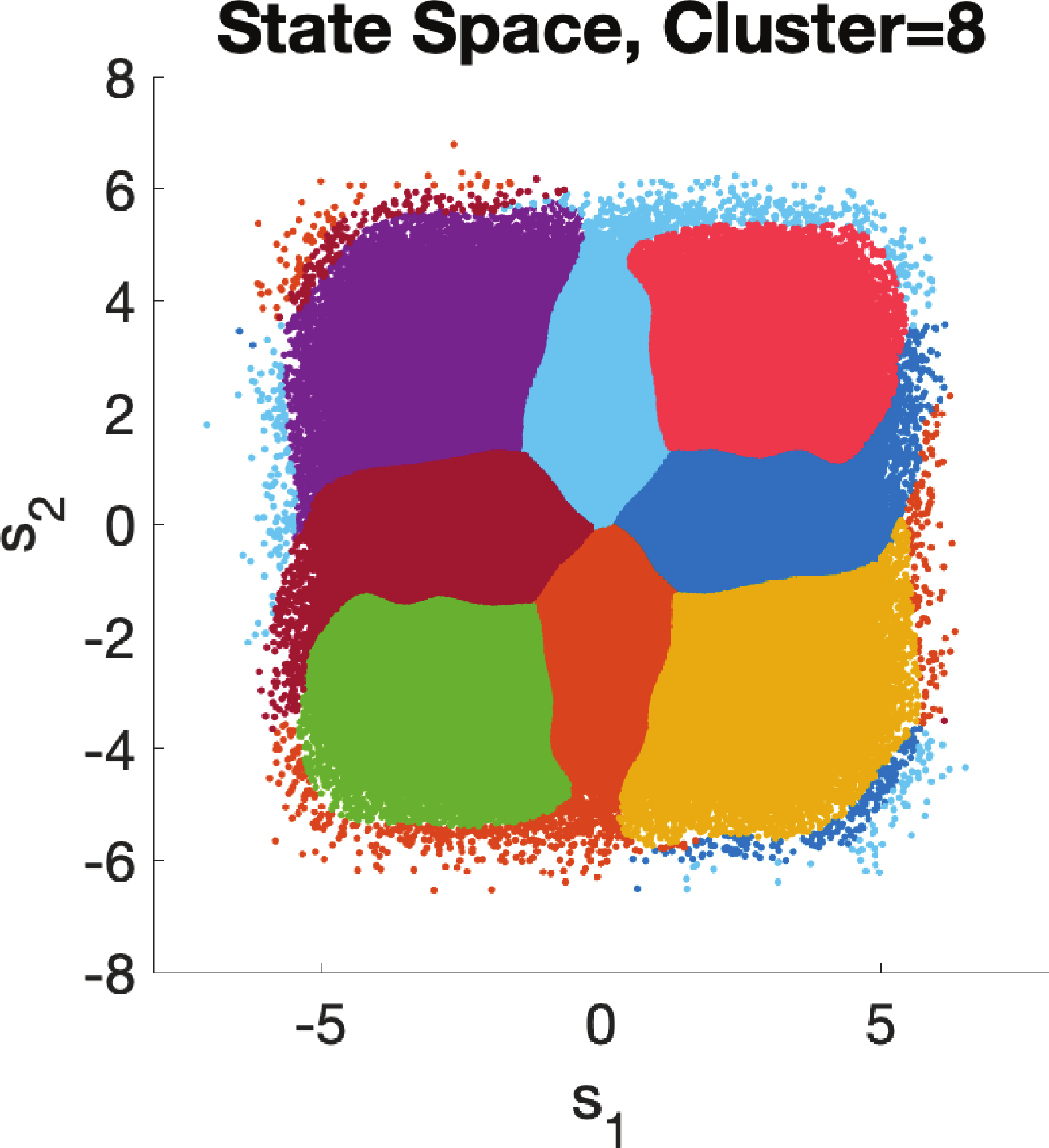}
\includegraphics[width=.24\linewidth]{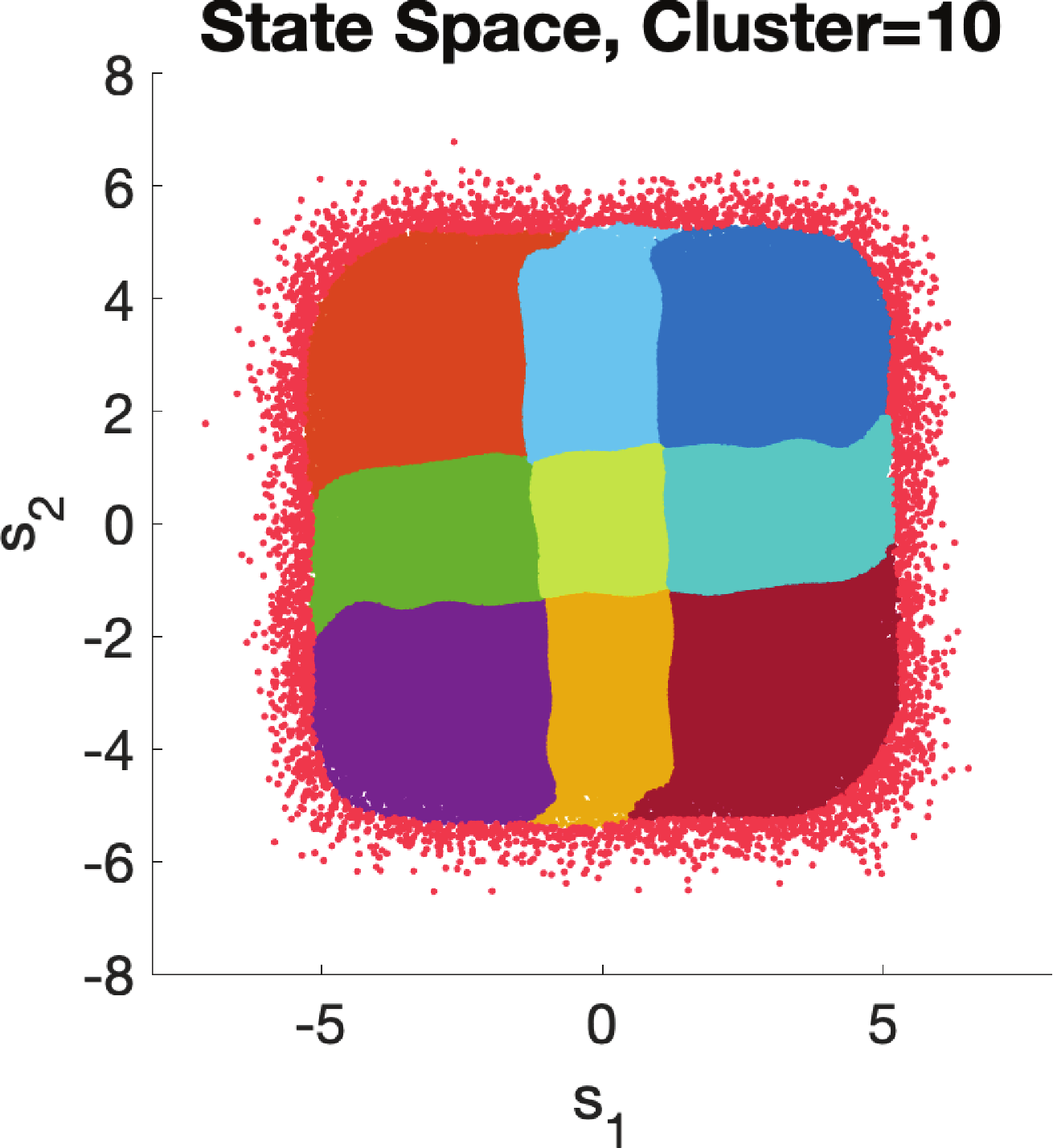}
\includegraphics[width=.24\linewidth]{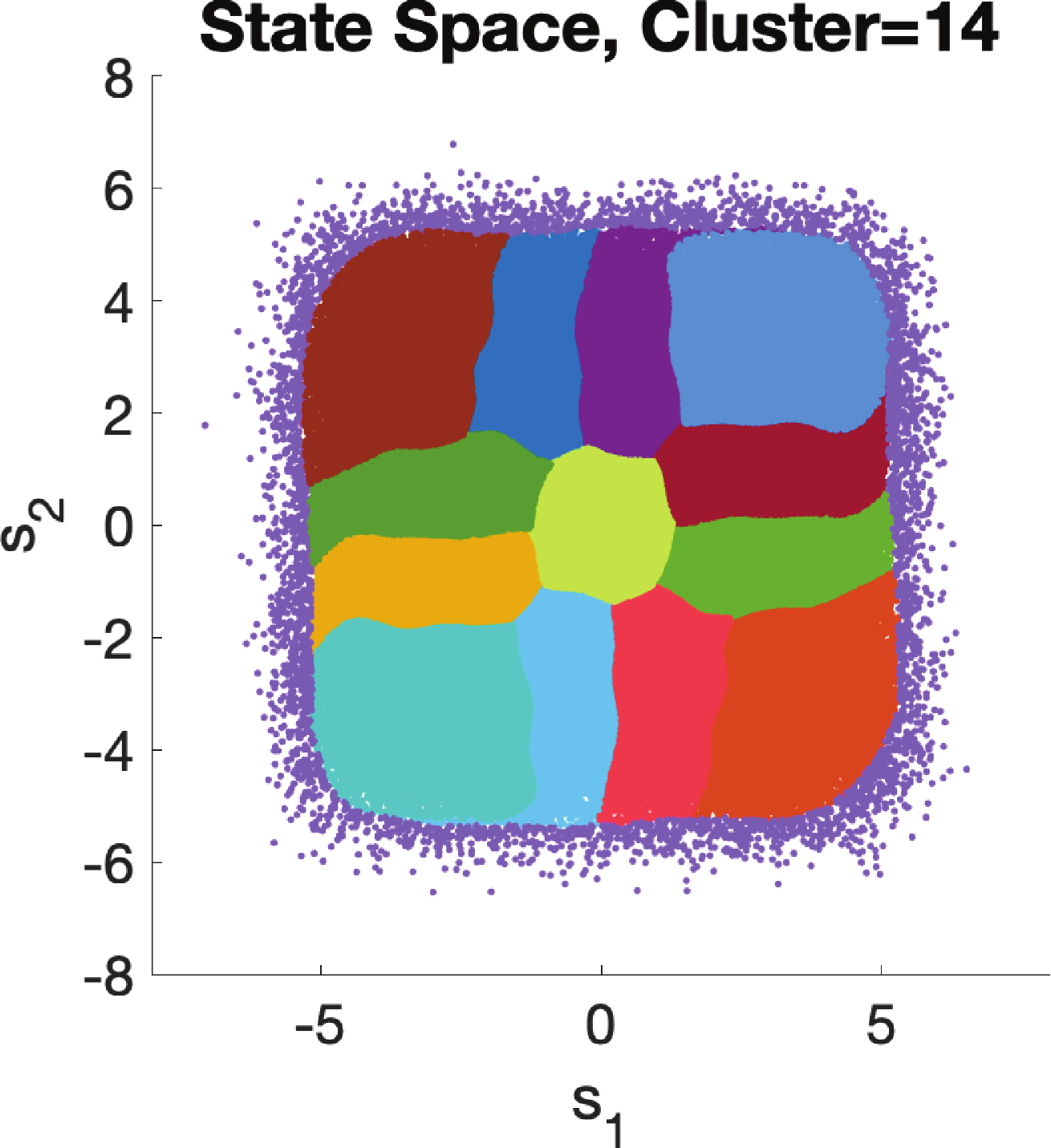}\\

\includegraphics[width=.24\linewidth]{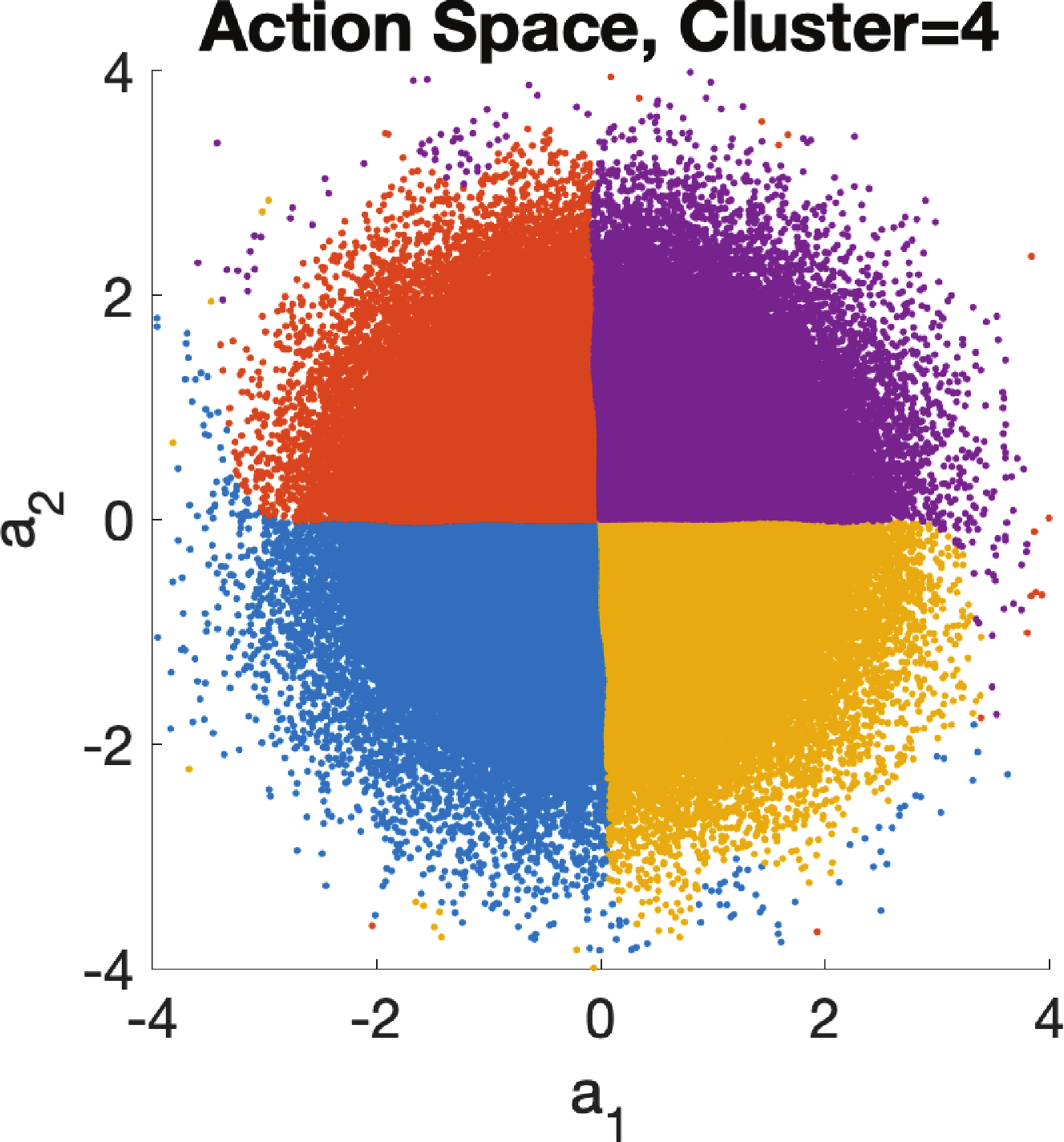}
\includegraphics[width=.24\linewidth]{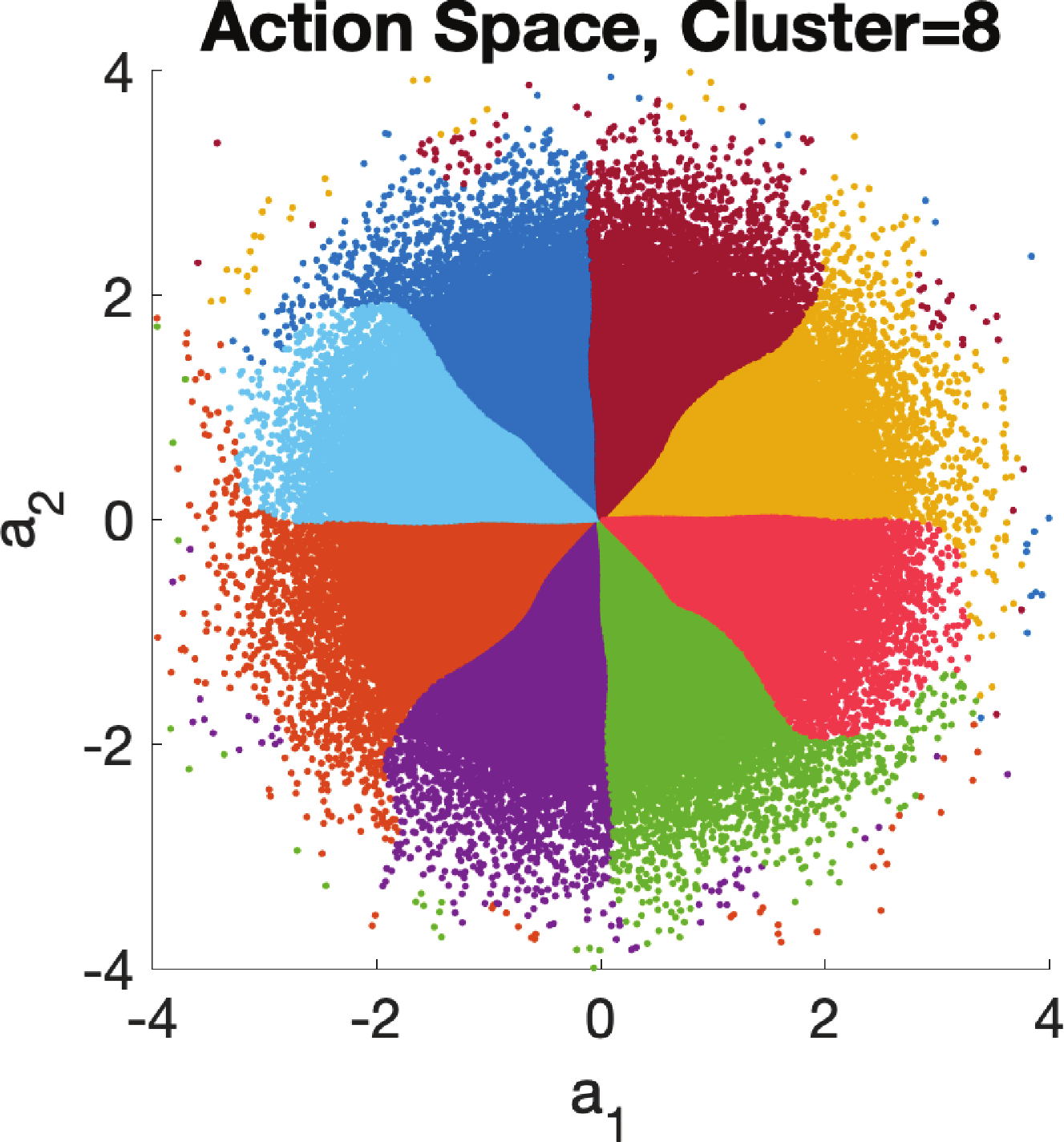}
\includegraphics[width=.24\linewidth]{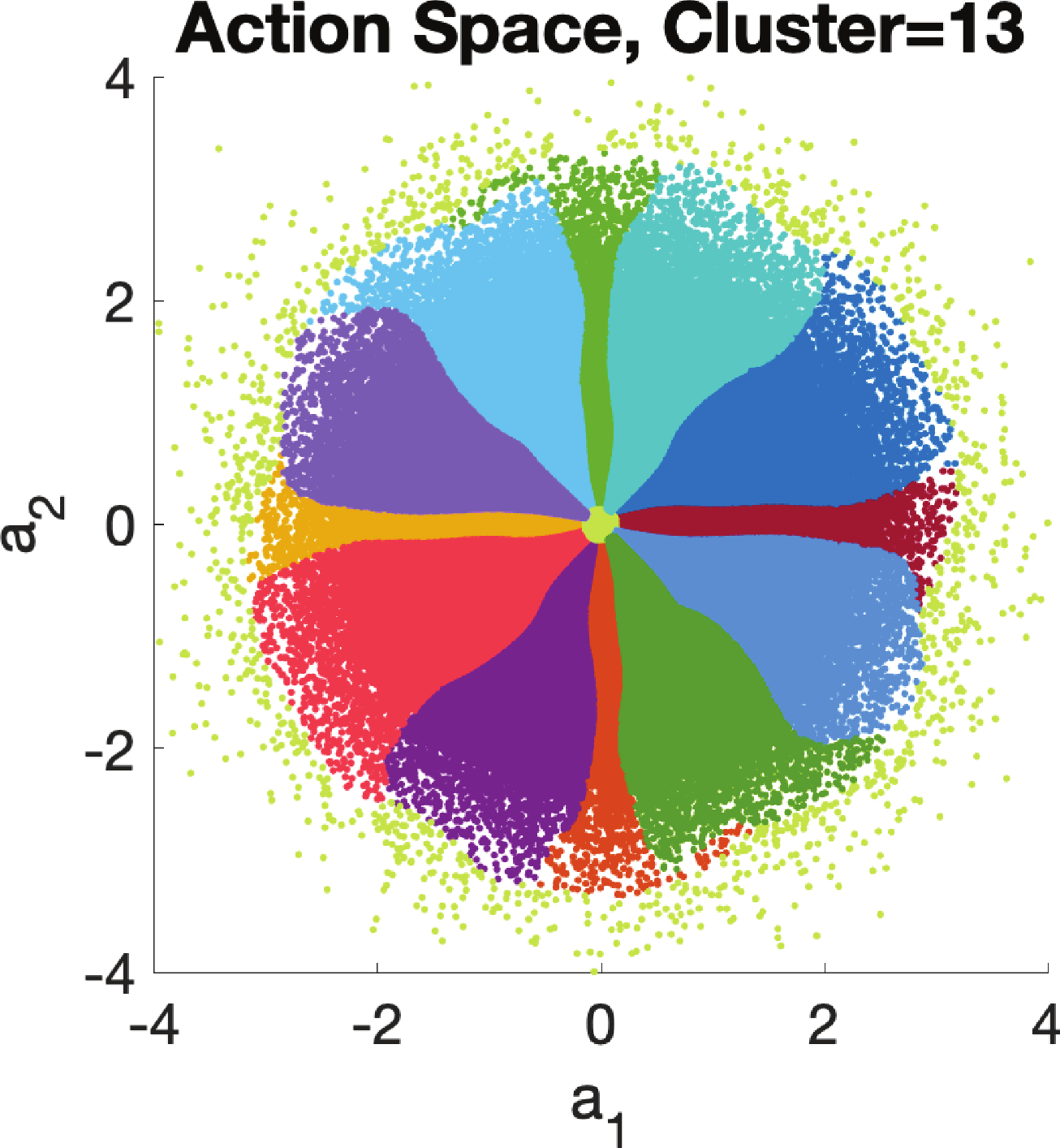}
\includegraphics[width=.24\linewidth]{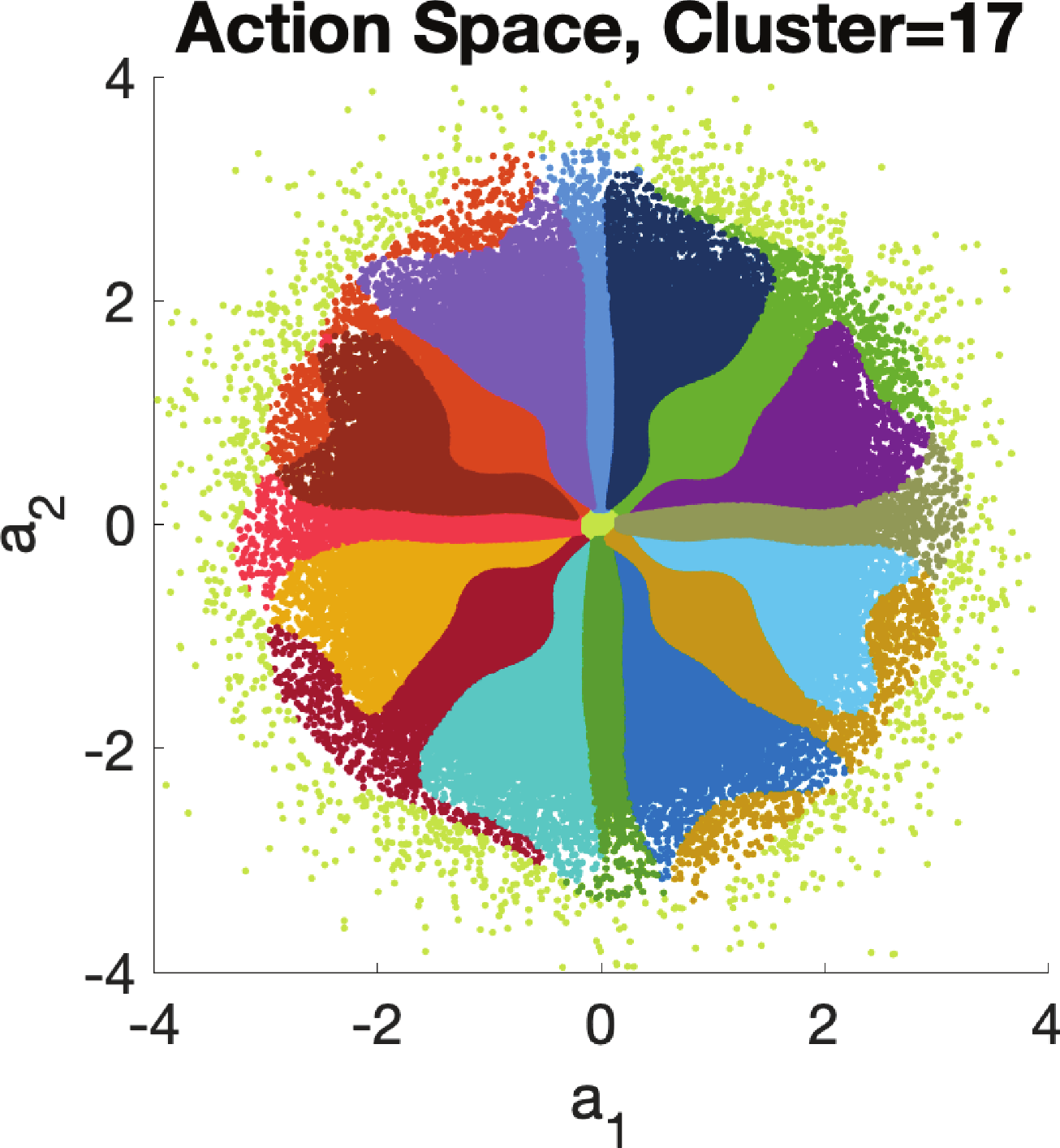}
\caption{Row 1: Left: Potential function $V(\cdot)$; Right: Block-wise control function $F(\cdot)$. The action space has 16 blocks, and in each block $F(\cdot)$ is a constant drift vector (see the arrows); Row 2: Learned state abstractions with varying clustering sizes; Row 3: Learned action abstractions with varying clustering sizes.}
\label{illu}
\end{figure}

We test our approach on a particular MDP derived from a controlled stochastic process. Let the state and action spaces be both $\mathbb{R}^2$. Suppose the state-action pair at step $k$ is $(s_k, a_k)$. Then the next state $s_{k+1}$ is set to be $X_{\tau(k+1)}$ for some $\tau > 0$, where $X_t$ is the solution of the SDE:
$$d X_t = -[\nabla V(X_t) + F(a_k)]dt + \sqrt{2}dB_t, k\tau \leq t \leq (k+1)\tau, $$
where $V(\cdot)$ is a wavy potential function, $F(\cdot)$ is a block-wise constant function (Figure \ref{illu}), $B_t$ is the standard Brownian motion. Let the behavior policy be always choosing $a$ from a standard normal distribution. 
We use the Gaussian kernels and obtain state/action features by generating $N$ random Fourier features $h = [h_1, h_2, \cdots, h_N]$ such that $K(x, y) \approx \sum_{i=1}^N h_i(x)h_i(y)$. 
\paragraph{State-action Clustering}
We first apply Algorithm \ref{alg2} to estimate state and action clusters. The results are shown at the right side of Figure \ref{illu}. Comparing them with the ground truth, we can validate that our method indeed reveal the latent state and action blocks.  

\begin{figure}[htb!]
\center
\includegraphics[width=.45\linewidth]{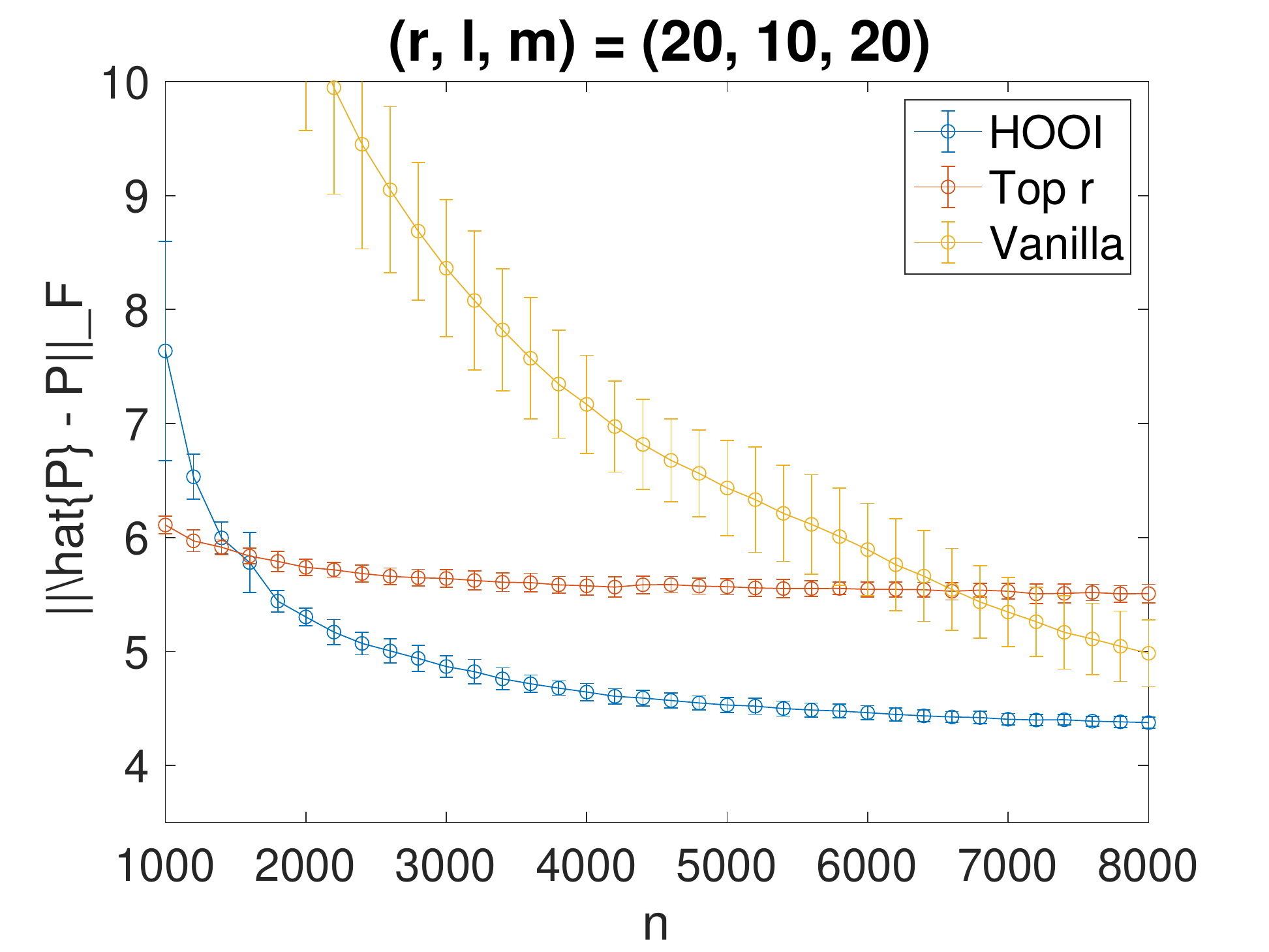}\hspace{0.1cm}\quad 
\includegraphics[width=.45\linewidth]{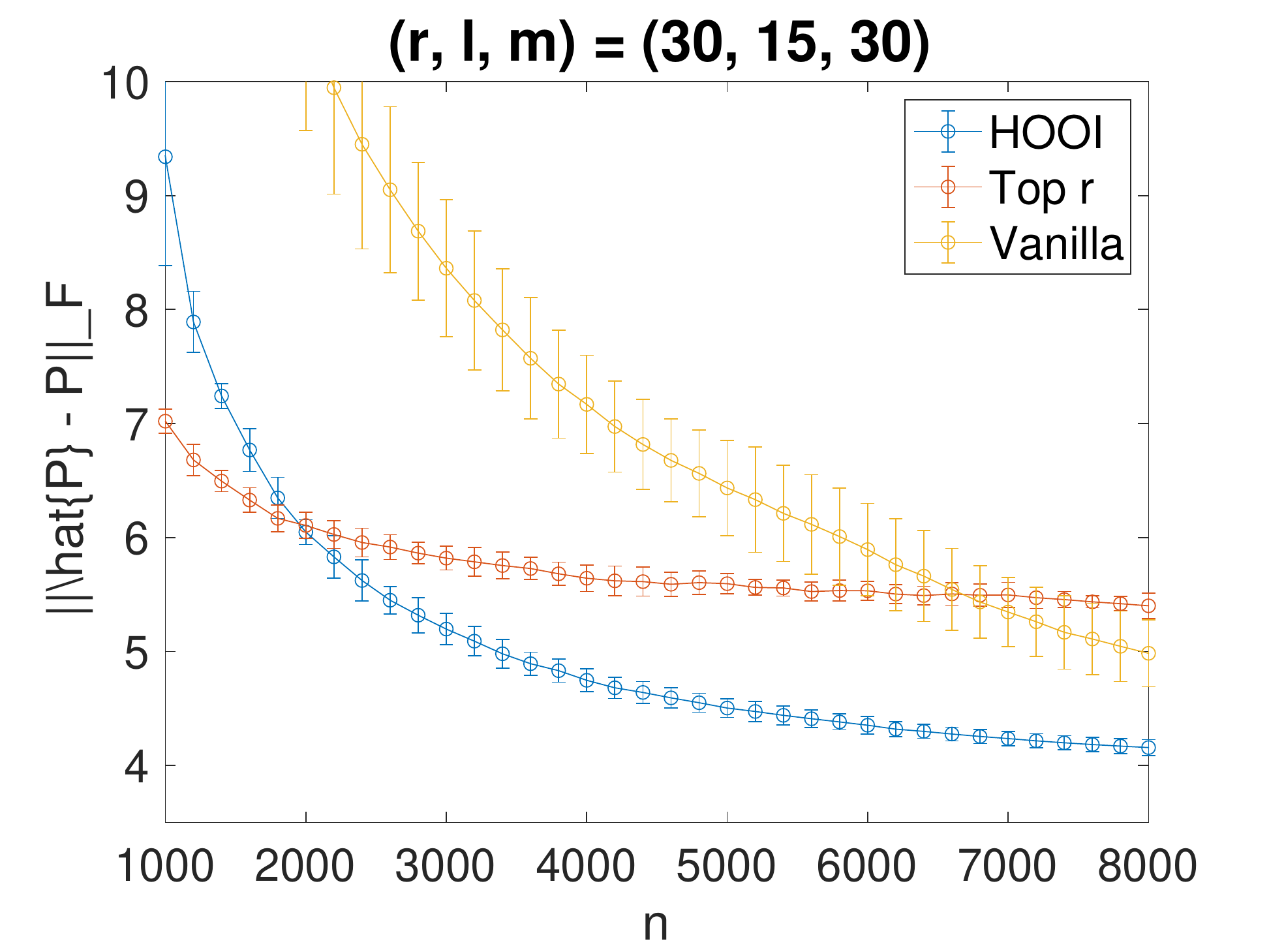}\hspace{0.1cm}\\
\includegraphics[width=.45\linewidth]{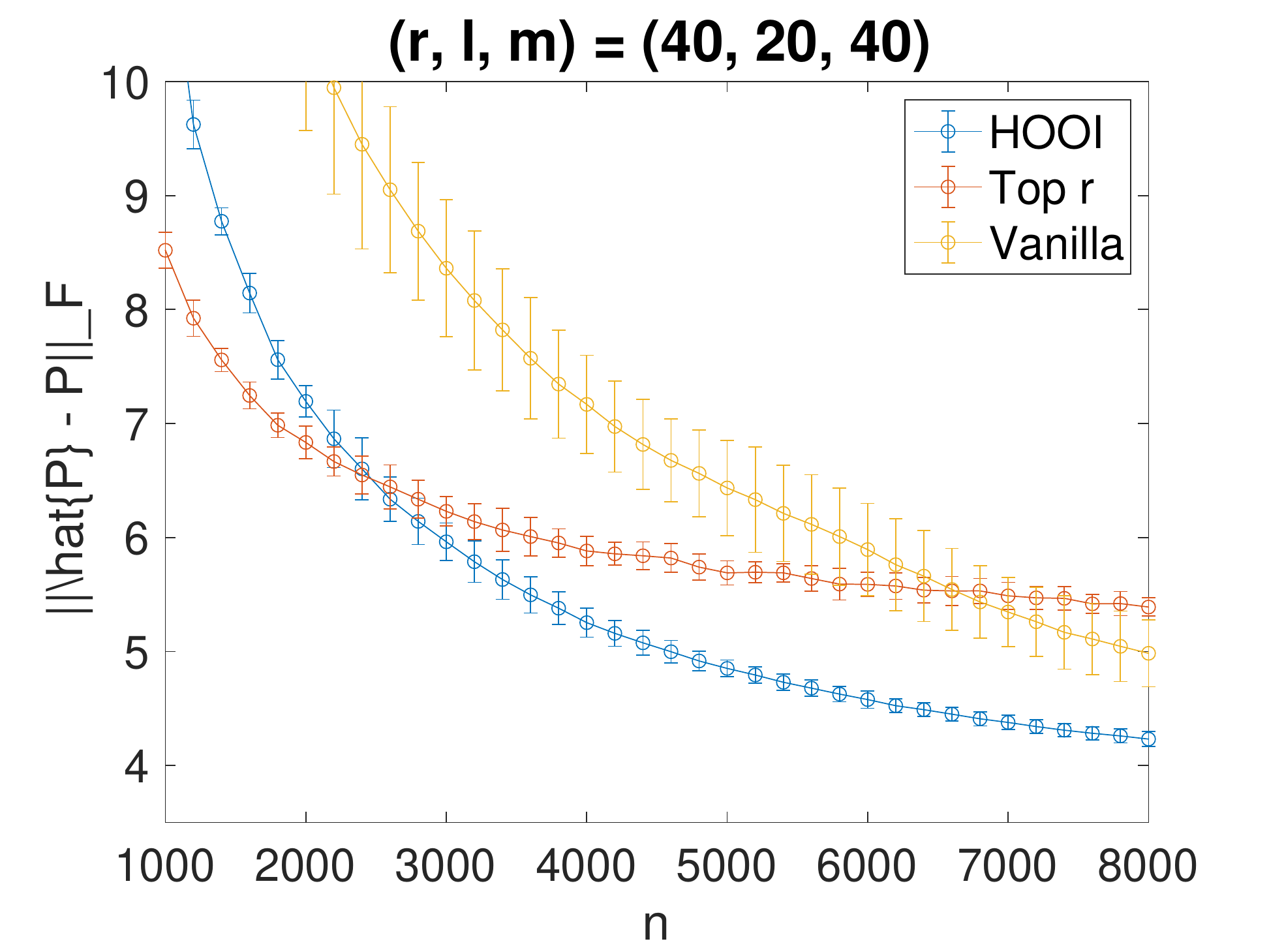}\hspace{0.1cm}\quad
\includegraphics[width=.45\linewidth]{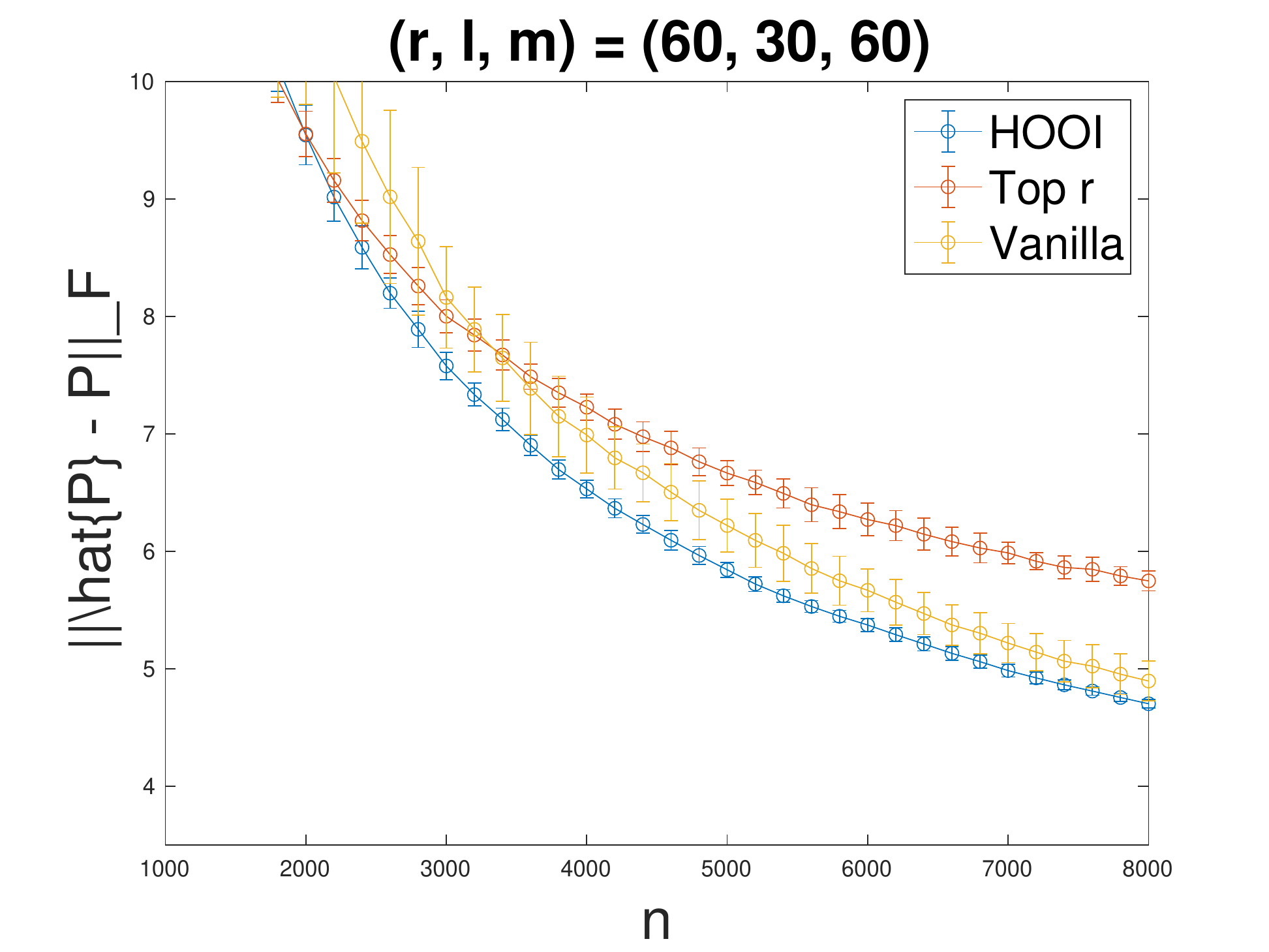}\hspace{0.1cm}
\caption{Low-tensor-rank estimation of $\bm{P}$, compared with baseline methods. }
\label{res2}
\end{figure}

\paragraph{Low-Rank Estimation of the Transition Tensor}
We then investigate the efficiency of estimating $\bm{P}$ via our tensor method. We compare our method with two baselines: (1) The vanilla method, which directly estimates the transition tensor by $\hat{\bm P} = \bar{\bm F}\times_1 \bar{\bm\Sigma}^{-1}$ without any low-rank approximation; (2) The ``top $r$'' method, whose the procedure is: i) calculate the top $r$ (or $l,m$) principle components of the sample covariance per mode; ii) project features onto the subspace spanned by the top principle components; iii) estimate the transition tensor via the vanilla method (discussed above) in the space of projected features. Fig.\ \ref{res2} visualizes the estimation errors of these methods with different choices of $(r, l, m)$, where errors are averaged over five independent runs. We observe that, for most of the time, our method consistently outperforms the baselines. Note that the top $r$ method performs slightly better when $n$ is very small, because in this case data is too small to get meaningful estimate of $\bm{P}$. The three approaches have similar performance when the rank constraint is set to be $(60, 30, 60)$ or higher. This is because the rank constraint is already close to the dimensions of the original state-action features, which reduces the impact of the rank-constrained estimator and introduces additional noise due to computational limitations. In practice, small rank constraints are preferred for both statistical and computational reasons.
\newpage

\bibliography{reference,reference-ys}

\begin{thebibliography}{66}
\providecommand{\natexlab}[1]{#1}
\providecommand{\url}[1]{\texttt{#1}}
\expandafter\ifx\csname urlstyle\endcsname\relax
  \providecommand{\doi}[1]{doi: #1}\else
  \providecommand{\doi}{doi: \begingroup \urlstyle{rm}\Url}\fi

\bibitem[Agarwal et~al.(2020)Agarwal, Kakade, Krishnamurthy, and
  Sun]{agarwal2020flambe}
Alekh Agarwal, Sham Kakade, Akshay Krishnamurthy, and Wen Sun.
\newblock Flambe: Structural complexity and representation learning of low rank
  mdps.
\newblock \emph{Advances in neural information processing systems},
  33:\penalty0 20095--20107, 2020.

\bibitem[Anandkumar et~al.(2014)Anandkumar, Ge, and
  Janzamin]{anandkumar2014guaranteed}
Animashree Anandkumar, Rong Ge, and Majid Janzamin.
\newblock Guaranteed non-orthogonal tensor decomposition via alternating
  rank-$1 $ updates.
\newblock \emph{arXiv preprint arXiv:1402.5180}, 2014.

\bibitem[Azizzadenesheli et~al.(2016)Azizzadenesheli, Lazaric, and
  Anandkumar]{azizzadenesheli2016reinforcement}
Kamyar Azizzadenesheli, Alessandro Lazaric, and Animashree Anandkumar.
\newblock Reinforcement learning in rich-observation mdps using spectral
  methods.
\newblock \emph{arXiv preprint arXiv:1611.03907}, 2016.

\bibitem[Bertsekas(2007)]{bertsekas1995dynamic}
Dimitri~P Bertsekas.
\newblock \emph{Dynamic programming and optimal control}.
\newblock Athena scientific Belmont, MA, 2007.

\bibitem[Bertsekas and Tsitsiklis(1996)]{bertsekas1995neuro}
Dimitri~P Bertsekas and John~N Tsitsiklis.
\newblock \emph{Neuro-dynamic programming}.
\newblock Athena Scientific, Belmont, MA, 1996.

\bibitem[Cai and Zhang(2018)]{cai2018rate}
T~Tony Cai and Anru Zhang.
\newblock Rate-optimal perturbation bounds for singular subspaces with
  applications to high-dimensional statistics.
\newblock \emph{The Annals of Statistics}, 46\penalty0 (1):\penalty0 60--89,
  2018.

\bibitem[Chowdhury and Gopalan(2019)]{chowdhury2019online}
Sayak~Ray Chowdhury and Aditya Gopalan.
\newblock Online learning in kernelized markov decision processes.
\newblock In \emph{The 22nd International Conference on Artificial Intelligence
  and Statistics}, pages 3197--3205. PMLR, 2019.

\bibitem[Cichocki et~al.(2015)Cichocki, Mandic, De~Lathauwer, Zhou, Zhao,
  Caiafa, and Phan]{cichocki2015tensor}
Andrzej Cichocki, Danilo Mandic, Lieven De~Lathauwer, Guoxu Zhou, Qibin Zhao,
  Cesar Caiafa, and Huy~Anh Phan.
\newblock Tensor decompositions for signal processing applications: From
  two-way to multiway component analysis.
\newblock \emph{IEEE Signal Processing Magazine}, 32\penalty0 (2):\penalty0
  145--163, 2015.

\bibitem[Coifman et~al.(2008)Coifman, Kevrekidis, Lafon, Maggioni, and
  Nadler]{coifman2008diffusion}
Ronald~R. Coifman, Ioannis~G. Kevrekidis, St\'ephane Lafon, Mauro Maggioni, and
  Boaz Nadler.
\newblock Diffusion maps, reduction coordinates, and low dimensional
  representation of stochastic systems.
\newblock \emph{SIAM Journal on Multiscale Modeling and Simulation}, 7\penalty0
  (2):\penalty0 852--864, 2008.

\bibitem[De~Lathauwer et~al.(2000{\natexlab{a}})De~Lathauwer, De~Moor, and
  Vandewalle]{de2000best}
Lieven De~Lathauwer, Bart De~Moor, and Joos Vandewalle.
\newblock On the best rank-1 and rank-(r 1, r 2,..., rn) approximation of
  higher-order tensors.
\newblock \emph{SIAM journal on Matrix Analysis and Applications}, 21\penalty0
  (4):\penalty0 1324--1342, 2000{\natexlab{a}}.

\bibitem[De~Lathauwer et~al.(2000{\natexlab{b}})De~Lathauwer, De~Moor, and
  Vandewalle]{de2000multilinear}
Lieven De~Lathauwer, Bart De~Moor, and Joos Vandewalle.
\newblock A multilinear singular value decomposition.
\newblock \emph{SIAM journal on Matrix Analysis and Applications}, 21\penalty0
  (4):\penalty0 1253--1278, 2000{\natexlab{b}}.

\bibitem[De~Silva and Lim(2008)]{de2008tensor}
Vin De~Silva and Lek-Heng Lim.
\newblock Tensor rank and the ill-posedness of the best low-rank approximation
  problem.
\newblock \emph{SIAM Journal on Matrix Analysis and Applications}, 30\penalty0
  (3):\penalty0 1084--1127, 2008.

\bibitem[Du et~al.(2019{\natexlab{a}})Du, Kakade, Wang, and Yang]{du2019good}
Simon~S Du, Sham~M Kakade, Ruosong Wang, and Lin~F Yang.
\newblock Is a good representation sufficient for sample efficient
  reinforcement learning?
\newblock \emph{arXiv preprint arXiv:1910.03016}, 2019{\natexlab{a}}.

\bibitem[Du et~al.(2019{\natexlab{b}})Du, Krishnamurthy, Jiang, Agarwal,
  Dud{\'\i}k, and Langford]{du2019provably}
Simon~S Du, Akshay Krishnamurthy, Nan Jiang, Alekh Agarwal, Miroslav
  Dud{\'\i}k, and John Langford.
\newblock Provably efficient rl with rich observations via latent state
  decoding.
\newblock \emph{arXiv preprint arXiv:1901.09018}, 2019{\natexlab{b}}.

\bibitem[E et~al.(2008)E, Li, and Vanden-Eijnden]{e2008optimal}
Weinan E, Tiejun Li, and Eric Vanden-Eijnden.
\newblock Optimal partition and effective dynamics of complex networks.
\newblock \emph{Proceedings of the National Academy of Sciences}, 105\penalty0
  (23):\penalty0 7907--7912, 2008.

\bibitem[Han et~al.(2022)Han, Willett, and Zhang]{han2022optimal}
Rungang Han, Rebecca Willett, and Anru~R Zhang.
\newblock An optimal statistical and computational framework for generalized
  tensor estimation.
\newblock \emph{The Annals of Statistics}, 50\penalty0 (1):\penalty0 1--29,
  2022.

\bibitem[Hong et~al.(2020)Hong, Kolda, and Duersch]{hong2020generalized}
David Hong, Tamara~G Kolda, and Jed~A Duersch.
\newblock Generalized canonical polyadic tensor decomposition.
\newblock \emph{SIAM Review}, 62\penalty0 (1):\penalty0 133--163, 2020.

\bibitem[Janzamin et~al.(2019)Janzamin, Ge, Kossaifi, and
  Anandkumar]{janzamin2019spectral}
Majid Janzamin, Rong Ge, Jean Kossaifi, and Anima Anandkumar.
\newblock Spectral learning on matrices and tensors.
\newblock \emph{Foundations and Trends{\textregistered} in Machine Learning},
  12\penalty0 (5-6):\penalty0 393--536, 2019.

\bibitem[Jiang et~al.(2017)Jiang, Krishnamurthy, Agarwal, Langford, and
  Schapire]{jiang2017contextual}
Nan Jiang, Akshay Krishnamurthy, Alekh Agarwal, John Langford, and Robert~E
  Schapire.
\newblock Contextual decision processes with low bellman rank are
  pac-learnable.
\newblock In \emph{Proceedings of the 34th International Conference on Machine
  Learning-Volume 70}, pages 1704--1713. JMLR. org, 2017.

\bibitem[Jin et~al.(2019)Jin, Yang, Wang, and Jordan]{jin2019provably}
Chi Jin, Zhuoran Yang, Zhaoran Wang, and Michael~I Jordan.
\newblock Provably efficient reinforcement learning with linear function
  approximation.
\newblock \emph{arXiv preprint arXiv:1907.05388}, 2019.

\bibitem[Johns and Mahadevan(2007)]{johns2007constructing}
Jeff Johns and Sridhar Mahadevan.
\newblock Constructing basis functions from directed graphs for value function
  approximation.
\newblock In \emph{Proceedings of the 24th international conference on Machine
  learning}, pages 385--392. ACM, 2007.

\bibitem[Jolliffe(1986)]{jolliffe1986principal}
Ian~T Jolliffe.
\newblock Principal components in regression analysis.
\newblock In \emph{Principal component analysis}, pages 129--155. Springer,
  1986.

\bibitem[Klus et~al.(2016)Klus, Koltai, and Sch{\"u}tte]{klus2016operator}
Stefan Klus, P{\'e}ter Koltai, and Christof Sch{\"u}tte.
\newblock On the numerical approximation of the perron–frobenius and koopman
  operator.
\newblock \emph{Journal of Computational Dynamics}, 3\penalty0 (1):\penalty0
  51--79, 2016.

\bibitem[Klus et~al.(2020)Klus, Schuster, and
  Muandet]{klus2020eigendecompositions}
Stefan Klus, Ingmar Schuster, and Krikamol Muandet.
\newblock Eigendecompositions of transfer operators in reproducing kernel
  hilbert spaces.
\newblock \emph{Journal of Nonlinear Science}, 30\penalty0 (1):\penalty0
  283--315, 2020.

\bibitem[Kolda and Bader(2009)]{kolda2009tensor}
Tamara~G Kolda and Brett~W Bader.
\newblock Tensor decompositions and applications.
\newblock \emph{SIAM review}, 51\penalty0 (3):\penalty0 455--500, 2009.

\bibitem[Lafon and Lee(2006)]{lafon2006diffusion}
St\'ephane Lafon and Ann Lee.
\newblock Diffusion maps and coarse-graining: A unied framework for
  dimensionality reduction, graph partitioning, and data set parameterization.
\newblock \emph{IEEE Trans. on Pattern Analysis and Machine Intelligence},
  29\penalty0 (9):\penalty0 1393--1403, 2006.

\bibitem[Lagoudakis and Parr(2003)]{lagoudakis2003least}
Michail~G Lagoudakis and Ronald Parr.
\newblock Least-squares policy iteration.
\newblock \emph{Journal of machine learning research}, 4\penalty0
  (Dec):\penalty0 1107--1149, 2003.

\bibitem[Lattimore and Szepesvari(2019)]{lattimore2019learning}
Tor Lattimore and Csaba Szepesvari.
\newblock Learning with good feature representations in bandits and in rl with
  a generative model.
\newblock \emph{arXiv preprint arXiv:1911.07676}, 2019.

\bibitem[Levin et~al.(2009)Levin, Peres, and Wilmer]{levin2009markov}
David~Asher Levin, Yuval Peres, and Elizabeth~Lee Wilmer.
\newblock \emph{Markov chains and mixing times}.
\newblock American Mathematical Soc., 2009.

\bibitem[L{\"o}ffler and Picard(2021)]{loffler2021spectral}
Matthias L{\"o}ffler and Antoine Picard.
\newblock Spectral thresholding for the estimation of markov chain transition
  operators.
\newblock \emph{Electronic Journal of Statistics}, 15\penalty0 (2):\penalty0
  6281--6310, 2021.

\bibitem[Mahadevan(2005)]{mahadevan2005proto}
Sridhar Mahadevan.
\newblock Proto-value functions: Developmental reinforcement learning.
\newblock In \emph{Proceedings of the 22nd international conference on Machine
  learning}, pages 553--560. ACM, 2005.

\bibitem[Mahadevan et~al.(2009)]{mahadevan2009learning}
Sridhar Mahadevan et~al.
\newblock Learning representation and control in markov decision processes: New
  frontiers.
\newblock \emph{Foundations and Trends{\textregistered} in Machine Learning},
  1\penalty0 (4):\penalty0 403--565, 2009.

\bibitem[Mahajan et~al.(2021)Mahajan, Samvelyan, Mao, Makoviychuk, Garg,
  Kossaifi, Whiteson, Zhu, and Anandkumar]{mahajan2021tesseract}
Anuj Mahajan, Mikayel Samvelyan, Lei Mao, Viktor Makoviychuk, Animesh Garg,
  Jean Kossaifi, Shimon Whiteson, Yuke Zhu, and Animashree Anandkumar.
\newblock Tesseract: Tensorised actors for multi-agent reinforcement learning.
\newblock In \emph{International Conference on Machine Learning}, pages
  7301--7312. PMLR, 2021.

\bibitem[Misra et~al.(2019)Misra, Henaff, Krishnamurthy, and
  Langford]{misra2019kinematic}
Dipendra Misra, Mikael Henaff, Akshay Krishnamurthy, and John Langford.
\newblock Kinematic state abstraction and provably efficient rich-observation
  reinforcement learning.
\newblock \emph{arXiv preprint arXiv:1911.05815}, 2019.

\bibitem[Modi et~al.(2021)Modi, Chen, Krishnamurthy, Jiang, and
  Agarwal]{modi2021model}
Aditya Modi, Jinglin Chen, Akshay Krishnamurthy, Nan Jiang, and Alekh Agarwal.
\newblock Model-free representation learning and exploration in low-rank mdps.
\newblock \emph{arXiv preprint arXiv:2102.07035}, 2021.

\bibitem[Moore(1991)]{moore1991variable}
Andrew~W Moore.
\newblock Variable resolution dynamic programming: Efficiently learning action
  maps in multivariate real-valued state-spaces.
\newblock In \emph{Machine Learning Proceedings 1991}, pages 333--337.
  Elsevier, 1991.

\bibitem[Ni et~al.(2023)Ni, Song, Zhang, Jin, and Wang]{ni2022representation}
Chengzhuo Ni, Yuda Song, Xuezhou Zhang, Chi Jin, and Mengdi Wang.
\newblock Representation learning for general-sum low-rank markov games.
\newblock \emph{International Conference on Learning Representations}, 2023.

\bibitem[Ormoneit and Glynn(2002)]{ormoneit2002kernel}
Dirk Ormoneit and Peter Glynn.
\newblock Kernel-based reinforcement learning in average-cost problems.
\newblock \emph{IEEE Transactions on Automatic Control}, 47\penalty0
  (10):\penalty0 1624--1636, 2002.

\bibitem[Panagakis et~al.(2021)Panagakis, Kossaifi, Chrysos, Oldfield,
  Nicolaou, Anandkumar, and Zafeiriou]{panagakis2021tensor}
Yannis Panagakis, Jean Kossaifi, Grigorios~G Chrysos, James Oldfield, Mihalis~A
  Nicolaou, Anima Anandkumar, and Stefanos Zafeiriou.
\newblock Tensor methods in computer vision and deep learning.
\newblock \emph{Proceedings of the IEEE}, 109\penalty0 (5):\penalty0 863--890,
  2021.

\bibitem[Parr et~al.(2007)Parr, Painter-Wakefield, Li, and
  Littman]{parr2007analyzing}
Ronald Parr, Christopher Painter-Wakefield, Lihong Li, and Michael Littman.
\newblock Analyzing feature generation for value-function approximation.
\newblock In \emph{Proceedings of the 24th international conference on Machine
  learning}, pages 737--744. ACM, 2007.

\bibitem[Petrik(2007)]{petrik2007analysis}
Marek Petrik.
\newblock An analysis of laplacian methods for value function approximation in
  mdps.
\newblock In \emph{IJCAI}, pages 2574--2579, 2007.

\bibitem[Rahimi and Recht(2008)]{rahimi2008random}
Ali Rahimi and Benjamin Recht.
\newblock Random features for large-scale kernel machines.
\newblock In \emph{Advances in neural information processing systems}, pages
  1177--1184, 2008.

\bibitem[Ren and Krogh(2002)]{ren2002state}
Zhiyuan Ren and Bruce~H Krogh.
\newblock State aggregation in markov decision processes.
\newblock In \emph{Decision and Control, 2002, Proceedings of the 41st IEEE
  Conference on}, volume~4, pages 3819--3824. IEEE, 2002.

\bibitem[Richard and Montanari(2014)]{richard2014statistical}
Emile Richard and Andrea Montanari.
\newblock A statistical model for tensor pca.
\newblock In \emph{Advances in Neural Information Processing Systems}, pages
  2897--2905, 2014.

\bibitem[Sch{\"u}tte et~al.(2011)Sch{\"u}tte, Noe, Lu, Sarich, and
  Vanden-Eijnden]{schutte2011coreset}
Christof Sch{\"u}tte, Frank Noe, Jianfeng Lu, Macro Sarich, and Eric
  Vanden-Eijnden.
\newblock Markov state models based on milestoning.
\newblock \emph{The Journal of Chemical Physics}, 134\penalty0 (20):\penalty0
  204105, 2011.

\bibitem[Sidiropoulos et~al.(2017)Sidiropoulos, De~Lathauwer, Fu, Huang,
  Papalexakis, and Faloutsos]{sidiropoulos2017tensor}
Nicholas~D Sidiropoulos, Lieven De~Lathauwer, Xiao Fu, Kejun Huang, Evangelos~E
  Papalexakis, and Christos Faloutsos.
\newblock Tensor decomposition for signal processing and machine learning.
\newblock \emph{IEEE Transactions on Signal Processing}, 65\penalty0
  (13):\penalty0 3551--3582, 2017.

\bibitem[Singh et~al.(1995)Singh, Jaakkola, and Jordan]{singh1995reinforcement}
Satinder~P Singh, Tommi Jaakkola, and Michael~I Jordan.
\newblock Reinforcement learning with soft state aggregation.
\newblock In \emph{Advances in neural information processing systems}, pages
  361--368, 1995.

\bibitem[Song et~al.(2016)Song, Woodruff, and Zhang]{song2016sublinear}
Zhao Song, David Woodruff, and Huan Zhang.
\newblock Sublinear time orthogonal tensor decomposition.
\newblock In \emph{Advances in Neural Information Processing Systems}, pages
  793--801, 2016.

\bibitem[Sun et~al.(2017)Sun, Lu, Liu, and Cheng]{sun2017provable}
Will~Wei Sun, Junwei Lu, Han Liu, and Guang Cheng.
\newblock Provable sparse tensor decomposition.
\newblock \emph{Journal of the Royal Statistical Society: Series B (Statistical
  Methodology)}, 79\penalty0 (3):\penalty0 899--916, 2017.

\bibitem[Sun et~al.(2019)Sun, Duan, Gong, and Wang]{sun2019learning}
Yifan Sun, Yaqi Duan, Hao Gong, and Mengdi Wang.
\newblock Learning low-dimensional state embeddings and metastable clusters
  from time series data.
\newblock In \emph{Advances in Neural Information Processing Systems}, pages
  4563--4572, 2019.

\bibitem[Sutton and Barto(1998)]{sutton1998reinforcement}
Richard~S Sutton and Andrew~G Barto.
\newblock \emph{Reinforcement learning: An introduction}, volume~1.
\newblock MIT press Cambridge, 1998.

\bibitem[Tropp(2011)]{tropp2011freedman}
Joel~A Tropp.
\newblock Freedman’s inequality for matrix martingales.
\newblock \emph{Electron. Commun. Probab}, 16:\penalty0 262--270, 2011.

\bibitem[Tsitsiklis and Van~Roy(1996)]{tsitsiklis1996feature}
John~N Tsitsiklis and Benjamin Van~Roy.
\newblock Feature-based methods for large scale dynamic programming.
\newblock \emph{Machine Learning}, 22\penalty0 (1-3):\penalty0 59--94, 1996.

\bibitem[Uehara et~al.(2021)Uehara, Zhang, and Sun]{uehara2021representation}
Masatoshi Uehara, Xuezhou Zhang, and Wen Sun.
\newblock Representation learning for online and offline rl in low-rank mdps.
\newblock \emph{arXiv preprint arXiv:2110.04652}, 2021.

\bibitem[Van Der~Vaart et~al.(2021)Van Der~Vaart, Mahajan, and
  Whiteson]{van2021model}
Pascal Van Der~Vaart, Anuj Mahajan, and Shimon Whiteson.
\newblock Model based multi-agent reinforcement learning with tensor
  decompositions.
\newblock \emph{arXiv preprint arXiv:2110.14524}, 2021.

\bibitem[Vannieuwenhoven et~al.(2012)Vannieuwenhoven, Vandebril, and
  Meerbergen]{vannieuwenhoven2012new}
Nick Vannieuwenhoven, Raf Vandebril, and Karl Meerbergen.
\newblock A new truncation strategy for the higher-order singular value
  decomposition.
\newblock \emph{SIAM Journal on Scientific Computing}, 34\penalty0
  (2):\penalty0 A1027--A1052, 2012.

\bibitem[Vershynin(2017)]{vershynin2017high}
Roman Vershynin.
\newblock \emph{High-Dimensional Probability}.
\newblock Cambridge University Press (to appear), 2017.

\bibitem[Wedin(1972)]{wedin1972perturbation}
Per-{\AA}ke Wedin.
\newblock Perturbation bounds in connection with singular value decomposition.
\newblock \emph{BIT Numerical Mathematics}, 12\penalty0 (1):\penalty0 99--111,
  1972.

\bibitem[Yang and Wang(2019)]{yang2019sample}
Lin Yang and Mengdi Wang.
\newblock Sample-optimal parametric q-learning using linearly additive
  features.
\newblock In \emph{International Conference on Machine Learning}, pages
  6995--7004, 2019.

\bibitem[Zanette et~al.(2019)Zanette, Lazaric, Kochenderfer, and
  Brunskill]{zanette2019limiting}
Andrea Zanette, Alessandro Lazaric, Mykel~J Kochenderfer, and Emma Brunskill.
\newblock Limiting extrapolation in linear approximate value iteration.
\newblock In \emph{Advances in Neural Information Processing Systems}, pages
  5616--5625, 2019.

\bibitem[Zhang(2019)]{zhang2019cross}
Anru Zhang.
\newblock Cross: Efficient low-rank tensor completion.
\newblock \emph{The Annals of Statistics}, 47\penalty0 (2):\penalty0 936--964,
  2019.

\bibitem[Zhang and Han(2019)]{zhang2019optimal-statsvd}
Anru Zhang and Rungang Han.
\newblock Optimal sparse singular value decomposition for high-dimensional
  high-order data.
\newblock \emph{Journal of the American Statistical Association}, pages
  1708--1725, 2019.

\bibitem[Zhang and Wang(2020)]{zhang2019spectral}
Anru Zhang and Mengdi Wang.
\newblock Spectral state compression of markov processes.
\newblock \emph{IEEE transactions on information theory}, 66\penalty0
  (5):\penalty0 3202--3231, 2020.

\bibitem[Zhang and Xia(2018)]{zhang2018tensor}
Anru Zhang and Dong Xia.
\newblock Tensor svd: Statistical and computational limits.
\newblock \emph{IEEE Transactions on Information Theory}, 64\penalty0
  (11):\penalty0 7311--7338, 2018.

\bibitem[Zhang et~al.(2022)Zhang, Song, Uehara, Wang, Agarwal, and
  Sun]{zhang2022efficient}
Xuezhou Zhang, Yuda Song, Masatoshi Uehara, Mengdi Wang, Alekh Agarwal, and Wen
  Sun.
\newblock Efficient reinforcement learning in block mdps: A model-free
  representation learning approach.
\newblock In \emph{International Conference on Machine Learning}, pages
  26517--26547. PMLR, 2022.

\bibitem[Zhu et~al.(2022)Zhu, Li, Wang, and Zhang]{zhu2022learning}
Ziwei Zhu, Xudong Li, Mengdi Wang, and Anru Zhang.
\newblock Learning markov models via low-rank optimization.
\newblock \emph{Operations Research}, 70\penalty0 (4):\penalty0 2384--2398,
  2022.

\end{thebibliography}

\appendix

\newpage
\onecolumn

\section*{Appendix}

\section*{A. The HOOI Algorithm}
\begin{algorithm}[htb!]
	\begin{algorithmic}[1]\label{alg-decomp}
		\caption{HOOI for MDP Tensor Decomposition}\label{alg:hooi}
		\STATE \textbf{Input}: tensor mean embedding $\bar{\bm{F}}$, $(r, l, m), t_{\max}$
		\STATE Initialization:
		$$\bar{\bm{U}}^{(0)}_1 = {\rm SVD}_{r}(\mathcal{M}_1(\bar{\bm{F}})), \bar{\bm{U}}^{(0)}_2 = {\rm SVD}_{l}(\mathcal{M}_2(\bar{\bm{F}}\times_1\bar{\bm{U}}_1^{(0)\top})),$$
		$$\bar{\bm{U}}^{(0)}_3 = {\rm SVD}_{m}(\mathcal{M}_3(\bar{\bm{F}}\times_1\bar{\bm{U}}_1^{(0)\top}\times_2\bar{\bm{U}}_2^{(0)\top})),$$
		where $\textrm{SVD}_{r}(\cdot)$ is the operation that returns the leading $r$ singular vector of matrix $\cdot$. 
		\FOR{$t=1,\ldots, t_{\max}$} 
		\STATE $\bar{\bm{U}}_1^{(t)} = \text{SVD}_r(\mathcal{M}_1(\bar{\bm{F}}\times_2 \bar{\bm{U}}_2^{(t-1)\top} \times_3 \bar{\bm{U}}_3^{(t-1)})\top)$,\\
		$\bar{\bm{U}}_2^{(t)} = \text{SVD}_l(\mathcal{M}_2(\bar{\bm{F}}\times_1 \bar{\bm{U}}_1^{(t)\top} \times_3 \bar{\bm{U}}_3^{(t-1)\top}))$,\\
		$\bar{\bm{U}}_3^{(t)} = \text{SVD}_m(\mathcal{M}_3(\bar{\bm{F}}\times_1 \bar{\bm{U}}_1^{(t)\top} \times_2 \bar{\bm{U}}_2^{(t)\top}))$.
		\ENDFOR
		\STATE \textbf{Output}: $\hat{\bm{F}} = \bar{\bm{F}}\times_1(\bar{\bm{U}}_1^{(t_{\max})\top}\bar{\bm{U}}_1^{(t_{\max})\top})\times_2(\bar{\bm{U}}_2^{(t_{\max})\top}\bar{\bm{U}}_2^{(t_{\max})\top})\times_3(\bar{\bm{U}}_3^{(t_{\max})\top}\bar{\bm{U}}_3^{(t_{\max})\top})$\\
	\end{algorithmic}
\end{algorithm}

\section*{B. Examples of Low Rank MDPs}
We give two basic examples.

\begin{figure}[h]
	\centering
	\begin{minipage}{\linewidth}
		\centering
		\includegraphics[width=.4\linewidth]{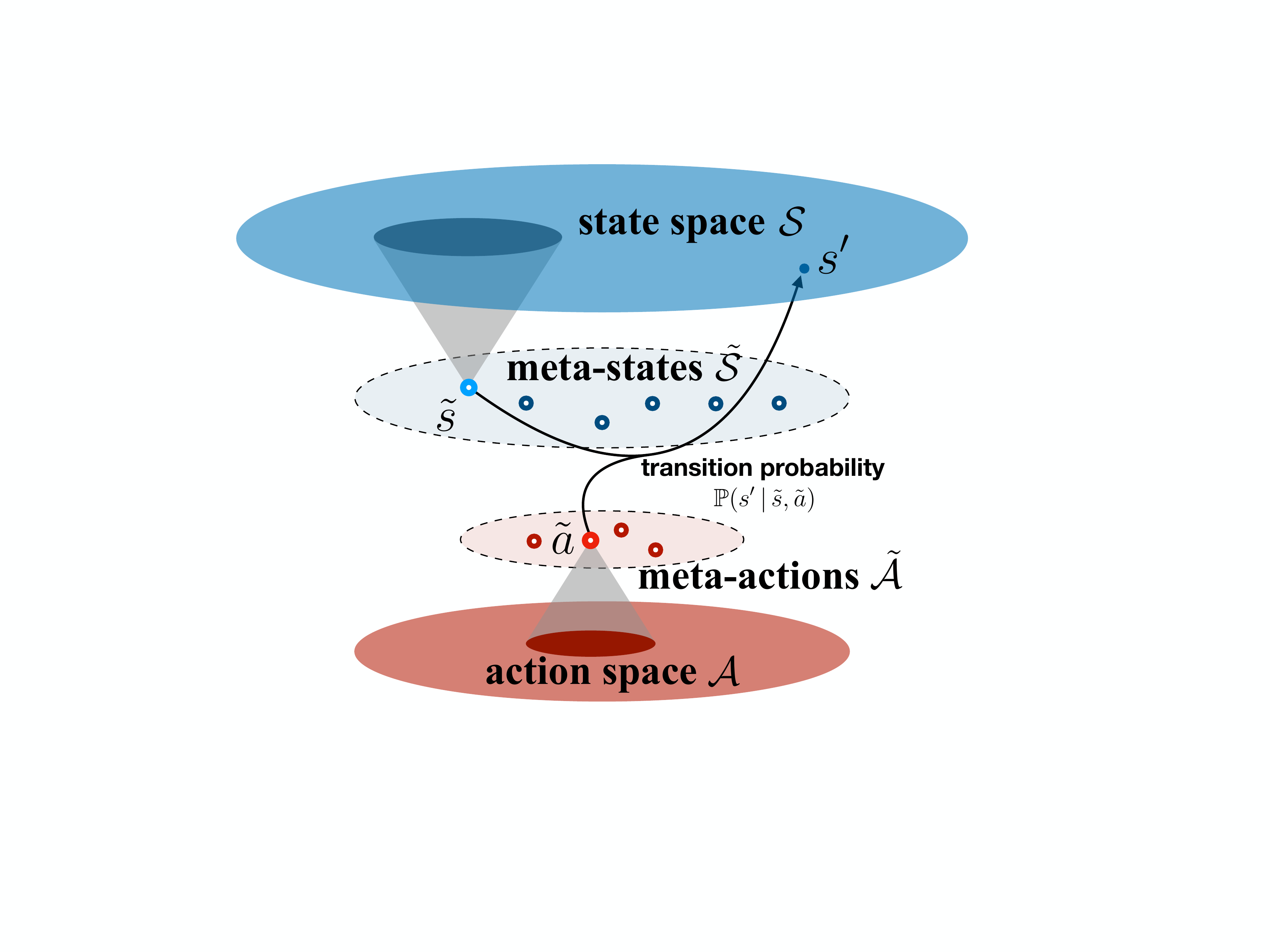}
		\hspace{0.5cm}
		\includegraphics[width=.4\linewidth]{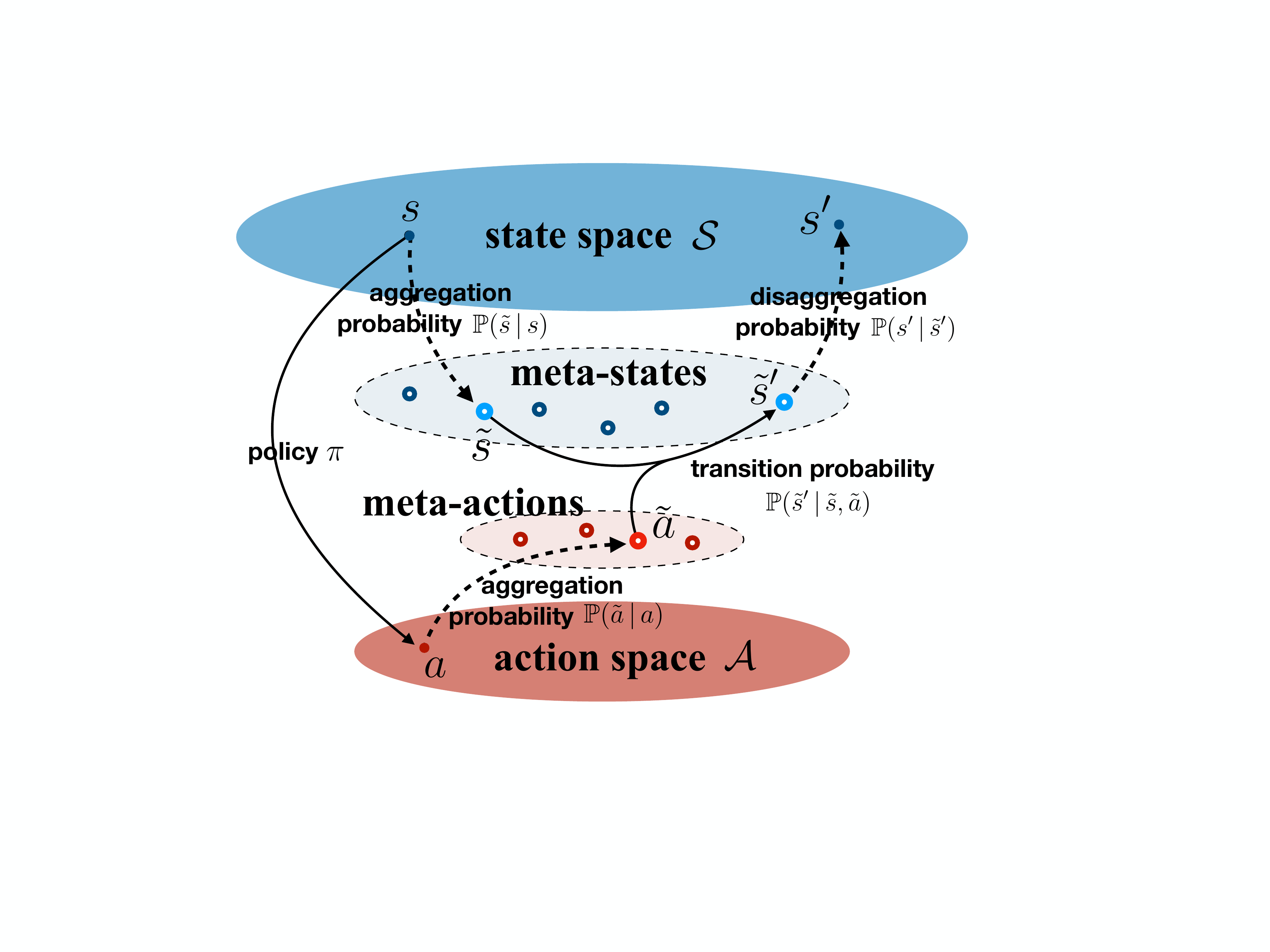}
	\end{minipage}
	\caption{Left: Block MDP (aka hard aggregation); Right: Latent-state-action MDP (aka soft aggregation).}
\end{figure}

\begin{example}[Block MDP (Hard Aggregation)] \rm
	Let $\tilde \cS$ and $\tilde \cA$ be finite sets. Suppose there exists state and action abstractions $f:\cS\mapsto \tilde \cS$ and $g:\cS\mapsto\tilde \cA$ such that
	$$p(\cdot| s,a)  = p(\cdot| s',a') \textrm{ if } f(s)=f(s'), g(a) = g(a')$$
	Then $p$ has Tucker rank at most $(|\tilde \cS|, |\tilde \cA|, \vert\cS\vert)$.
\end{example}

\begin{example}[Latent-State-Action MDP (Soft Aggregation)]\rm
	Given an MDP $\mathcal{M} = (\mathcal{S}, \mathcal{A}, p, r)$, we say $\mathcal{M}$ has an $(r, l, m)$-latent variable model if there exist a latent state-action-state stochastic process $\{\tilde s_t, \tilde a_t, \tilde s_t'\}\subseteq \tilde{\mathcal{S}}\times\tilde{\mathcal{A}}\times \tilde{\mathcal{S}}'$, with $\vert\tilde{\mathcal{S}}\vert = r, \vert\tilde{\mathcal{A}}\vert = l, \vert\tilde{\mathcal{S}}'\vert = m$, such that 
	\begin{align*}
	&\mathbb{P}(\tilde{s}_t, \tilde{a}_t \vert s_1, a_1, \ldots, s_t, a_t) = \mathbb{P}(\tilde{s}_{t} \vert s_t)\mathbb{P}(\tilde{a}_{t} \vert a_t), \mathbb{P}(\tilde{s}'_{t} \vert s_1, a_1, \ldots, s_t, a_t, \tilde{s}_t, \tilde{a}_t) = \mathbb{P}(\tilde{s}'_t \vert \tilde{s}_t, \tilde{a}_t),\\
	&\mathbb{P}(s_{t+1} \vert s_1, a_1, \ldots, s_t, a_t, \tilde{s}_t, \tilde{a}_t, \tilde{s}'_t) = \mathbb{P}(s_{t+1} \vert \tilde{s}'_t).
	\end{align*}
	In this case, one can verify that $p$ has Tucker rank $(r,l,m)$.
\end{example}

We give an illustrative example to show the advantage of utilizing tensor MDP formulation as opposed to the matrix ones. 
\begin{example}\rm
	Consider $\mathcal{A} = \{1, 2\}$, $\mathcal{S} = \{1,2,3,4\}$. Construct the MDP transition tensor $\bm{P}$ as
	$$\bm{P}_{\cdot 1\cdot}  =\begin{bmatrix}
	1/6 & 1/6 & 1/3 & 1/3\\
	1/6 & 1/6 & 1/3 & 1/3\\
	1/3 & 1/3 & 1/6 & 1/6\\
	1/3 & 1/3 & 1/6 & 1/6
	\end{bmatrix},$$  
	$$\bm{P}_{\cdot 2\cdot}  =\begin{bmatrix}
	1/3 & 1/3 & 1/6 & 1/6 \\
	1/3 & 1/3 & 1/6 & 1/6\\
	1/6 & 1/6 & 1/3 & 1/3\\
	1/6 & 1/6 & 1/3 & 1/3
	\end{bmatrix}.$$
	Then, $\bm{P} = \bm{C} \times_1 \bm{U} \times_3 \bm{U}$ for 
	$$\bm{U} = \begin{bmatrix}
	1 & 0\\
	1 & 0\\
	0 & 1\\
	0 & 1
	\end{bmatrix}, \bm{C}_{\cdot 1\cdot} = \begin{bmatrix}
	1/6 & 1/3\\
	1/3 & 1/6
	\end{bmatrix}, \bm{C}_{\cdot 2\cdot} = \begin{bmatrix}
	1/3 & 1/6\\
	1/6 & 1/3
	\end{bmatrix}$$ 
	and the state-space is aggregatable into two meta-states: $\{1,2\}$ and $\{3,4\}$. Consider a random policy: $\pi(a|s) = 1/2$ for $a=1,2$. Without taking into account the tensor structure induced by the policy, one can check that the state transitions $\{s_0, s_1, \ldots \}$ form a Markov process with the following transition matrix
	\begin{equation*}
	\tilde{\bm{P}} = \frac{1}{2} \bm{P}_{\cdot 1\cdot} + \frac{1}{2} \bm{P}_{\cdot 2\cdot} = \begin{bmatrix}
	1/4 & 1/4 & 1/4 & 1/4 \\
	1/4 & 1/4 & 1/4 & 1/4 \\
	1/4 & 1/4 & 1/4 & 1/4 \\
	1/4 & 1/4 & 1/4 & 1/4
	\end{bmatrix}.
	\end{equation*}
	Clearly, the meta-state partition is ``averaged out" using any matrix methods and there is no hope to extract the meta-state information merely from the state transitions $\{s_0, s_1, \ldots \}$. On the other hand, the tensor formulation, which preserves the original state-action-state, allows a reliable state aggregation efficiently.
	
\end{example}

\section*{C. Derivation of Optimization Problem (\ref{prob})}
The original optimization objective is 
\begin{align*}
&\min_{A_i, B_j}\min_{\{q_{ij}\}} \sum_{i, j}\int_{A_i\times B_j} \xi(s)\eta(a)\Vert  p(\cdot\vert s, a) - q_{ij}(\cdot)\|_{\mathcal{H}_S}^2 dsda\\
=&\min_{A_i, B_j}\min_{\{q_{ij}\}} \sum_{i, j}\int_{A_i\times B_j} \xi(s)\eta(a)\Vert\langle p(\cdot\vert s, a), \phi(\cdot)\rangle -  \langle q_{ij}(\cdot), \phi(\cdot)\rangle\Vert^2 dsda.
\end{align*}
Simple calculations show that given fixed $A_i, B_j$, the best choice of $\langle q_{ij}(\cdot), \phi(\cdot)\rangle$ is
\begin{align*}
\langle q_{ij}(\cdot), \phi(\cdot)\rangle &= \frac{1}{\xi(A_i)\eta(B_j)}\int_{A_i\times B_j} \xi(s)\eta(a)\langle p(\cdot\vert s, a), \phi(\cdot)\rangle dsda\\
&= \frac{1}{\xi(A_i)\eta(B_j)}\int_{A_i\times B_j} \xi(s)\eta(a)(\bm{P}\times_1 \phi(s)^\top \times_2\psi(a)^\top)dsda\\
&=\frac{1}{\xi(A_i)\eta(B_j)}\int_{A_i\times B_j} \xi(s)\eta(a)\big(\bm{C}\times_1 f(s)^\top\times_2 g(a)^\top \times_3 \bm{U}_3\big)dsda\\
&= \bm{U}_3\bm{z}_{ij},
\end{align*}
where $\bm{z}_{ij} = \frac{1}{\xi(A_i)\eta(B_j)}\int_{A_t\times B_j} \xi(s)\eta(a)(\bm{C}\times_1 f(s)^\top\times_2 g(a)^\top)dsda$. Note that
\begin{align*}
&\sum_{i, j}\int_{A_i\times B_j} \xi(s)\eta(a)\Vert\langle p(\cdot\vert s, a), \phi(\cdot)\rangle - \bm{U}_3\bm{z}_{ij}\Vert^2 dsda\\
=&\sum_{i, j}\int_{A_i\times B_j} \xi(s)\eta(a)\Vert\bm{C}\times_1 f(s)^\top \times_2 g(a)^\top \times_3\bm{U}_3 - \bm{U}_3\bm{z}_{ij}\Vert^2 dsda\\
=& \sum_{i, j}\int_{A_i\times B_j} \xi(s)\eta(a)\Vert\bm{C}\times_1 f(s)^\top \times_2 g(a)^\top - \bm{z}_{ij}\Vert^2 dsda.
\end{align*}
Therefore, our problem can be further formalized as
\begin{align*}
\min_{A_i, B_j}\min_{\bm{z}_{ij}}  \sum_{i, j}\int_{A_i\times B_j} \xi(s)\eta(a)\Vert\bm{C}\times_1 f(s)^\top \times_2 g(a)^\top - \bm{z}_{ij}\Vert^2 dsda.
\end{align*}
When we only have empirical data, the above problem can be approximated by
\begin{align*}
\min_{A_i, B_j}\min_{\bm{z}_{ij}}  \sum_{i, j}\int_{A_i\times B_j} \xi(s)\eta(a)\Vert\hat{\bm{C}}\times_1 \hat{f}(s)^\top \times_2 \hat{g}(a)^\top - \bm{z}_{ij}\Vert^2 dsda,
\end{align*}
which is exactly (\ref{prob}). 

\section*{D. Experiment Details}
In the experiment, we use the Gaussian kernels $K_1(x, y) = K_A(x, y) = \frac{1}{2\pi\sigma^2}\exp\{-\frac{\Vert x-y\Vert^2}{2\sigma^2}\}$. And the features are obtained by generating $N_s$ (or $N_a$) random Fourier features $h = [h_1, h_2, \cdots, h_{N_s}]$ such that $K(x, y) \approx \sum_{i=1}^{N_s(N_a)} h_i(x)h_i(y)$. And the action features are then orthogonalized with respect to $L^2(\eta)$. In the experiment, we choose $\tau = 0.1, \sigma = 0.5, N_s = 100, N_a=50$. 

For the clustering problem, we choose sample size $n=10^6$ and $(r, l, m) = (3, 3, 3)$, and the state features are further orthogonalized with respect to $L^2(\xi)$. For the estimation problem, the ground-truth is approximately obtained from the vanilla method with sample size $n=10^6$. The following figure shows the clustering result of the top-$r$ method, which does the clustering on the subspace spanned by the top-$r$ (or $l, m$) eigenvectors of the covariance matrix. From the figure we can see that the top-$r$ method does not capture the correct clustering information of the transition kernel compared with our method. 

\begin{figure}[htb!]
	\center
	\includegraphics[width=.22\linewidth]{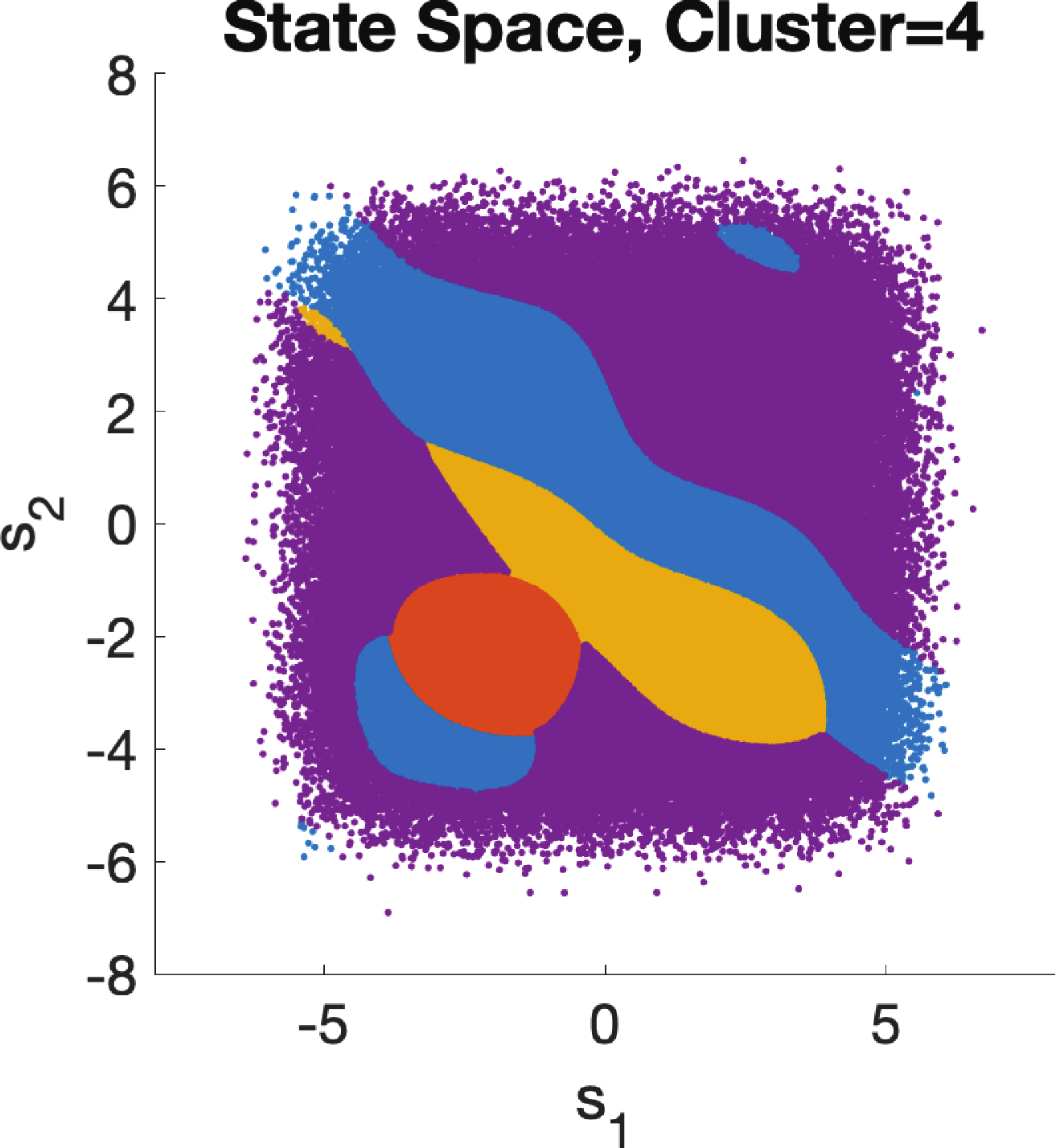}
	\includegraphics[width=.22\linewidth]{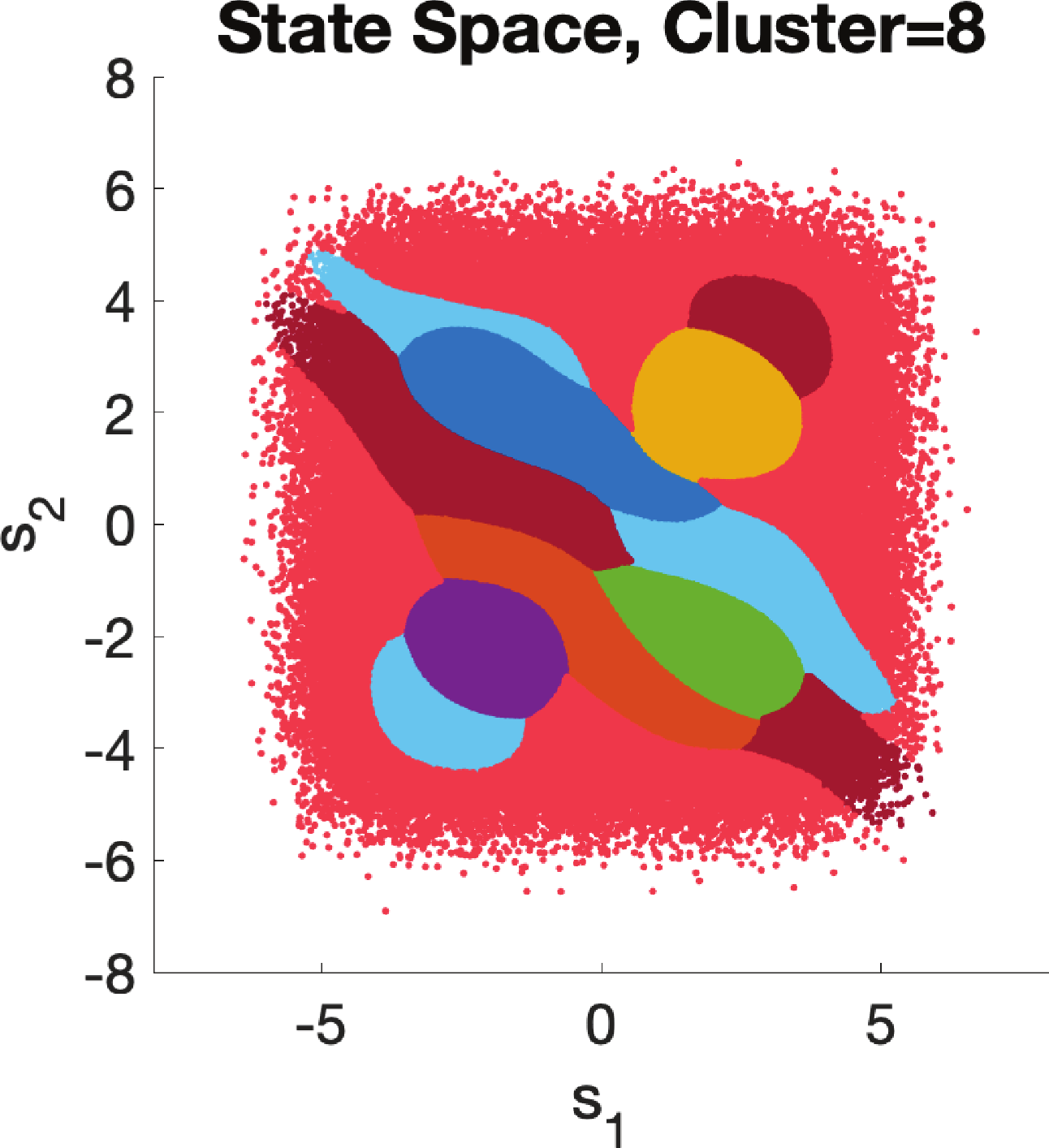}
	\includegraphics[width=.22\linewidth]{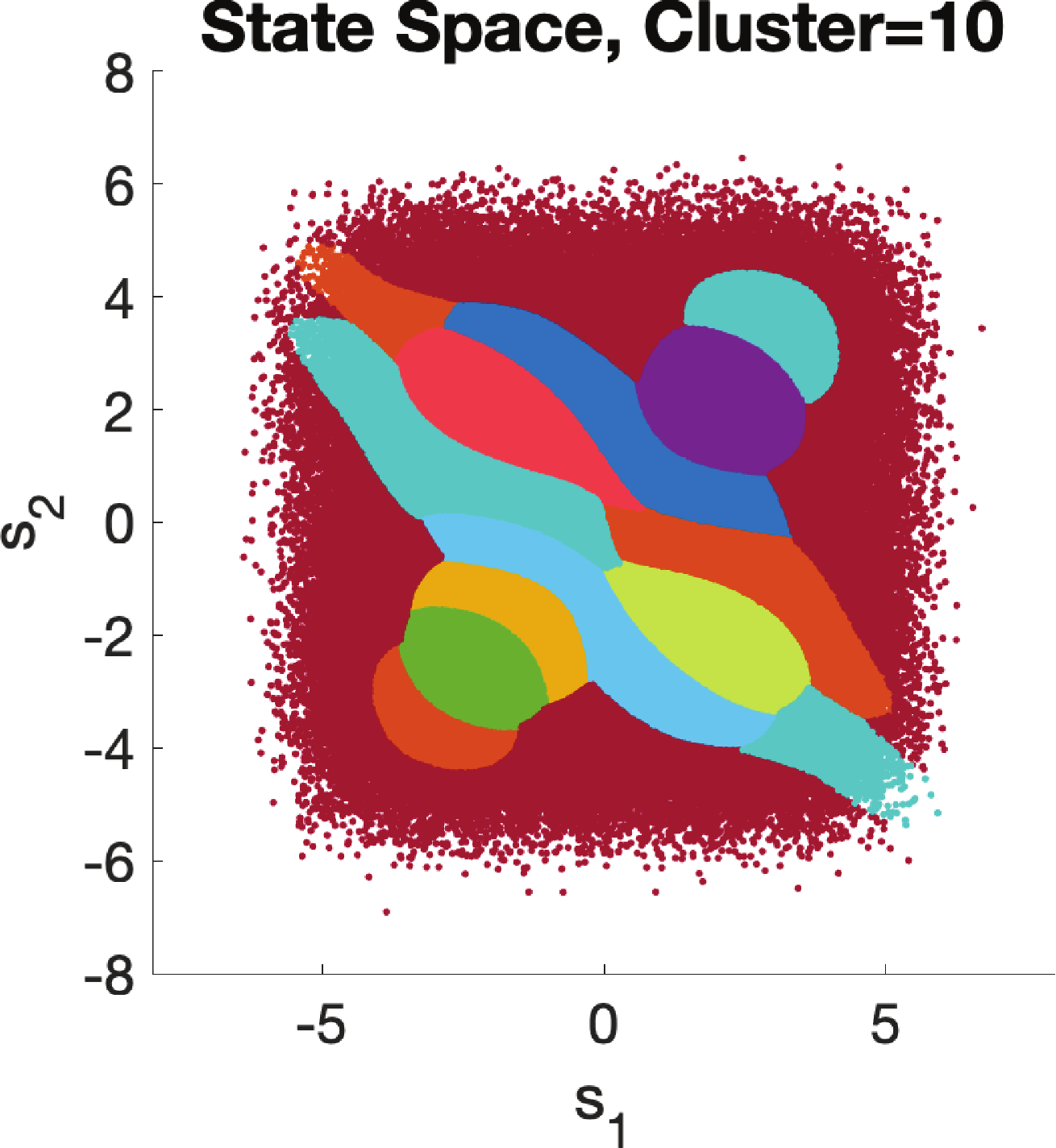}
	\includegraphics[width=.22\linewidth]{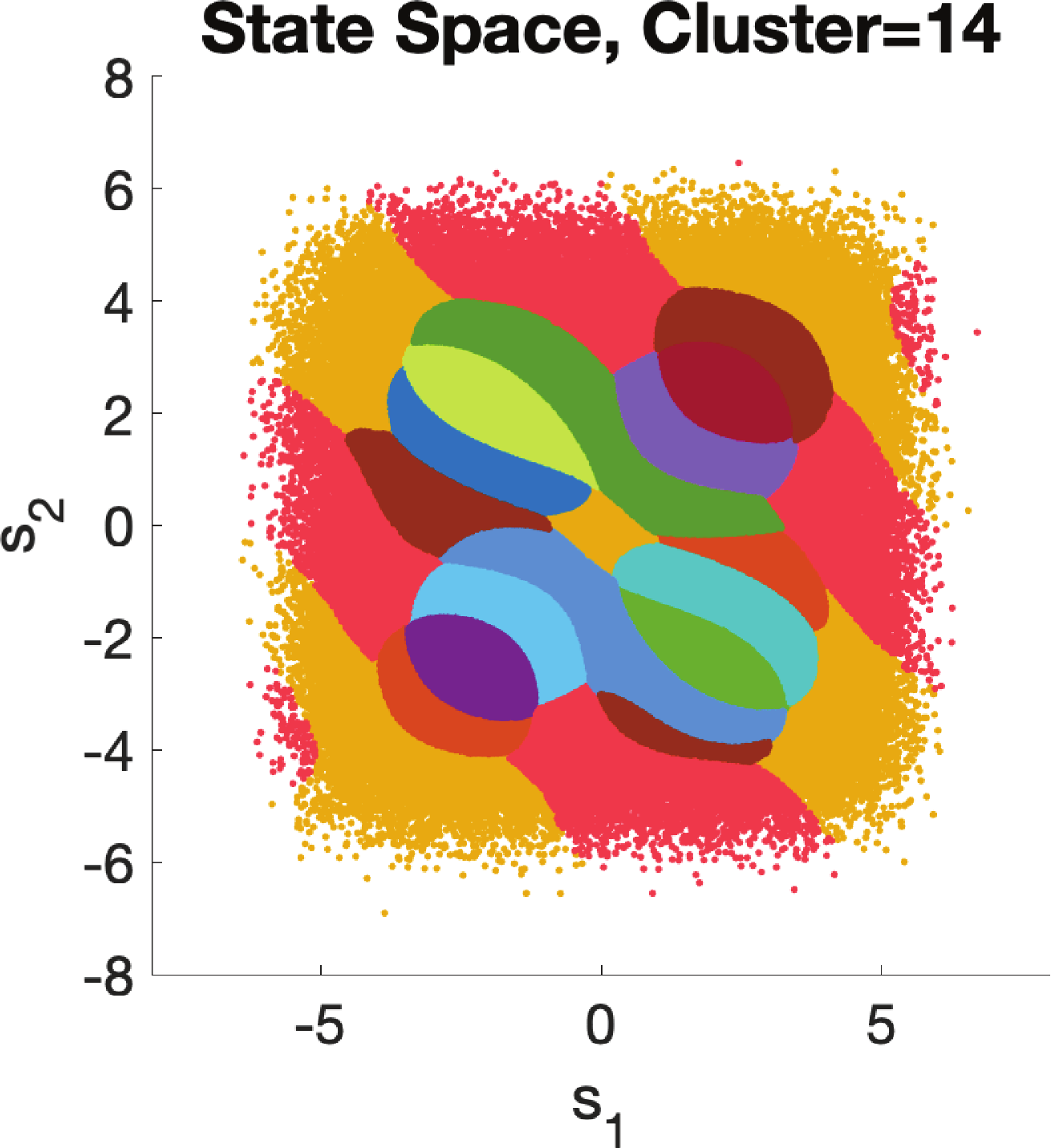}\\
	\includegraphics[width=.22\linewidth]{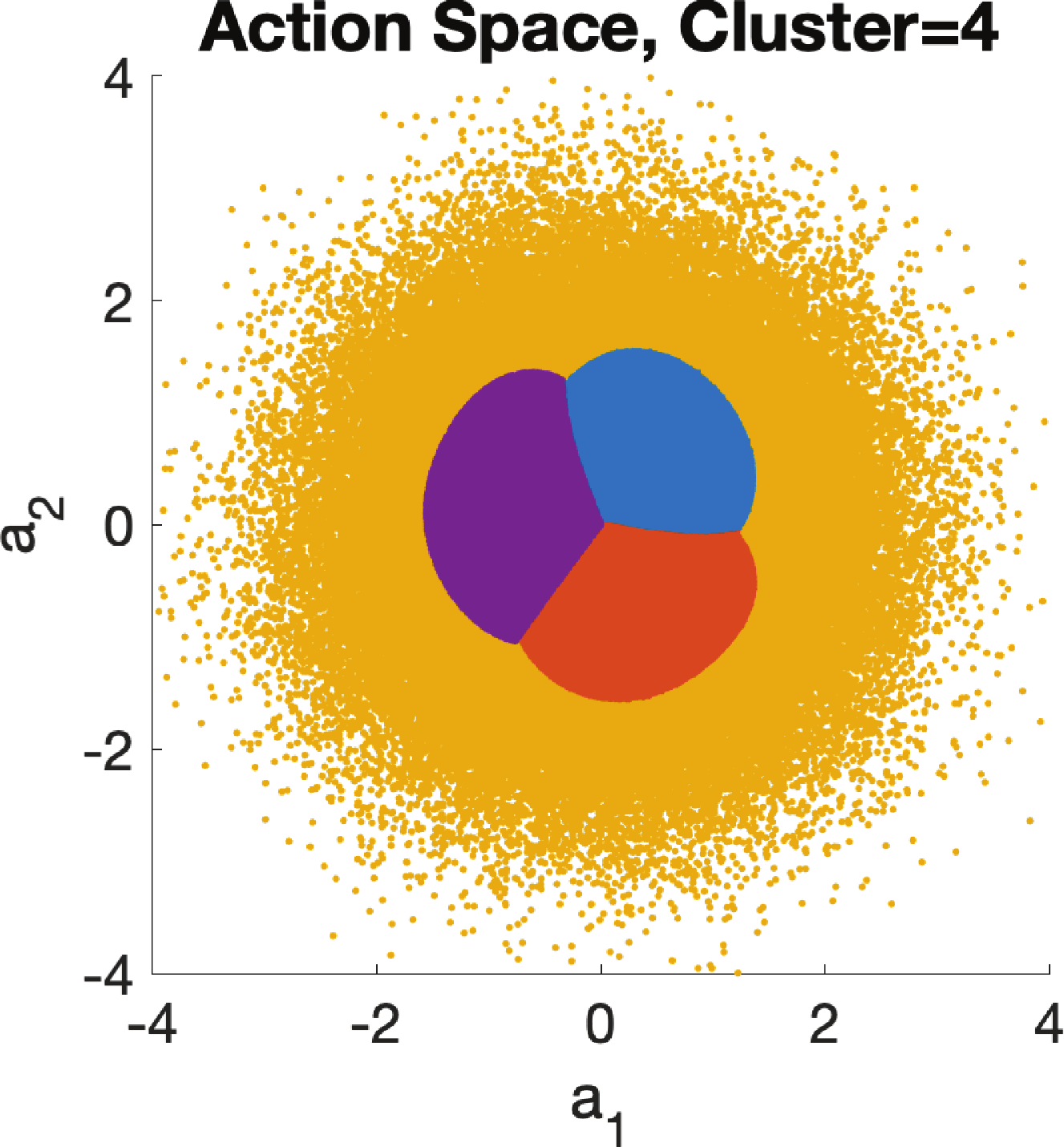}
	\includegraphics[width=.22\linewidth]{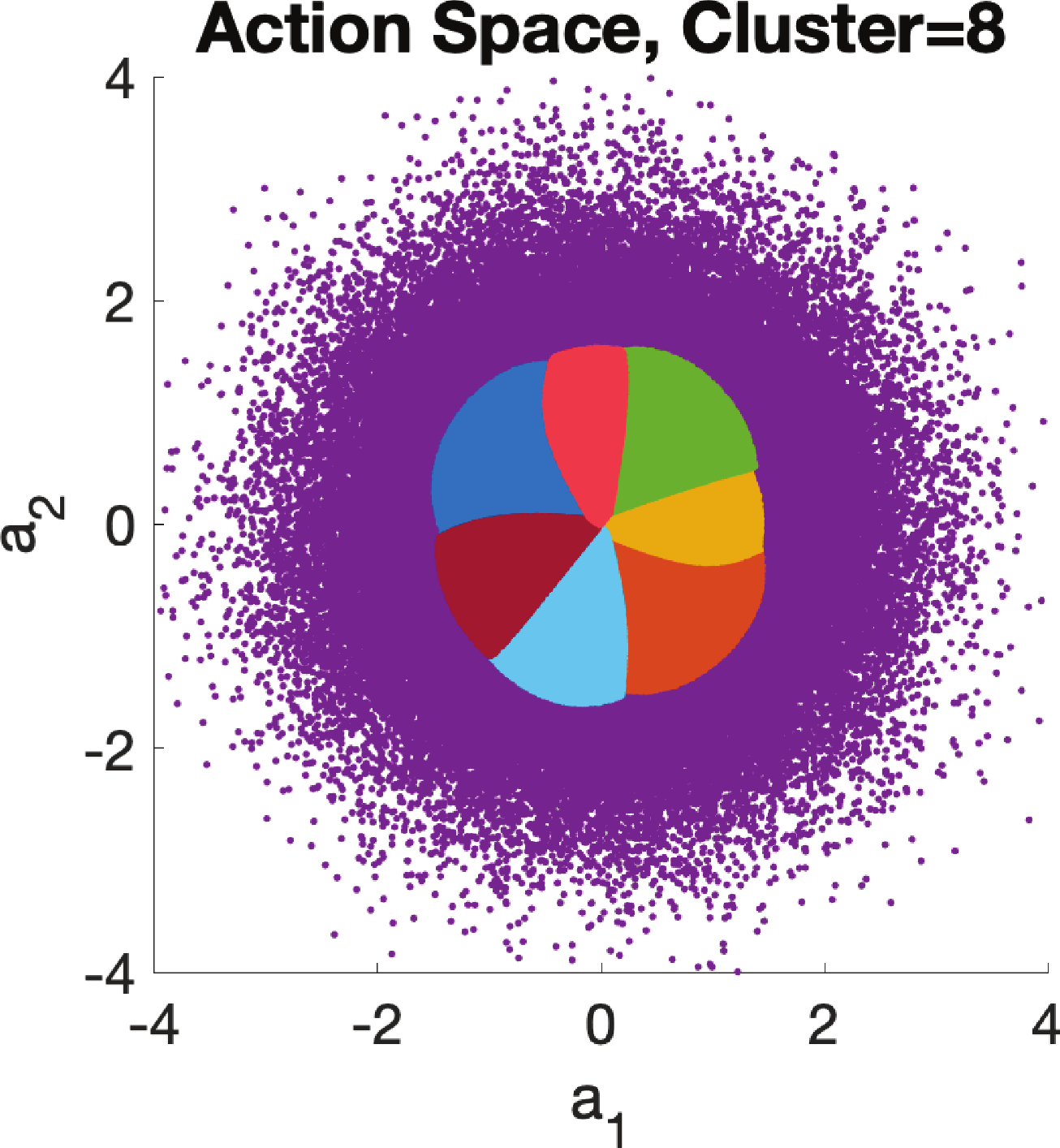}
	\includegraphics[width=.22\linewidth]{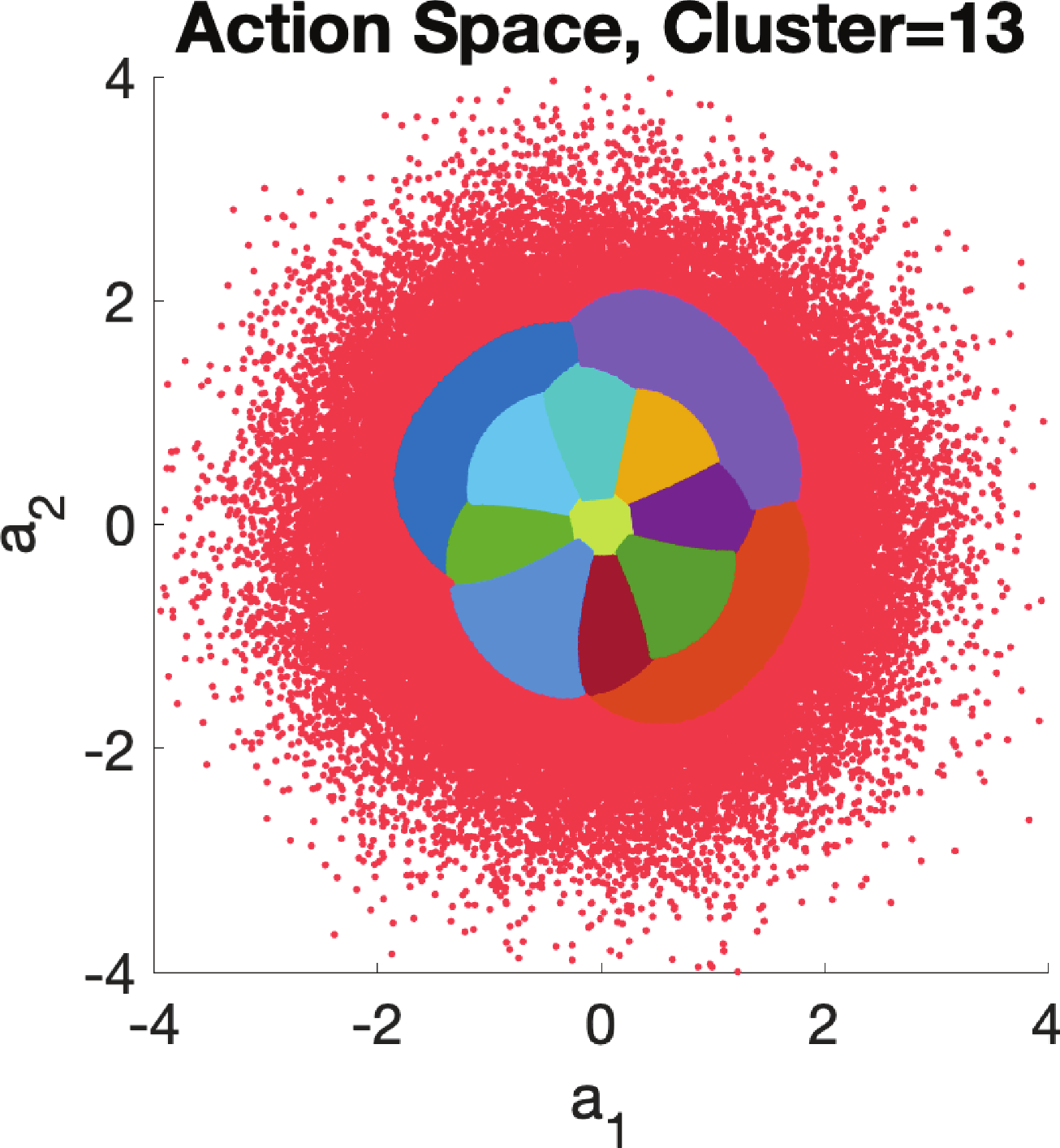}
	\includegraphics[width=.22\linewidth]{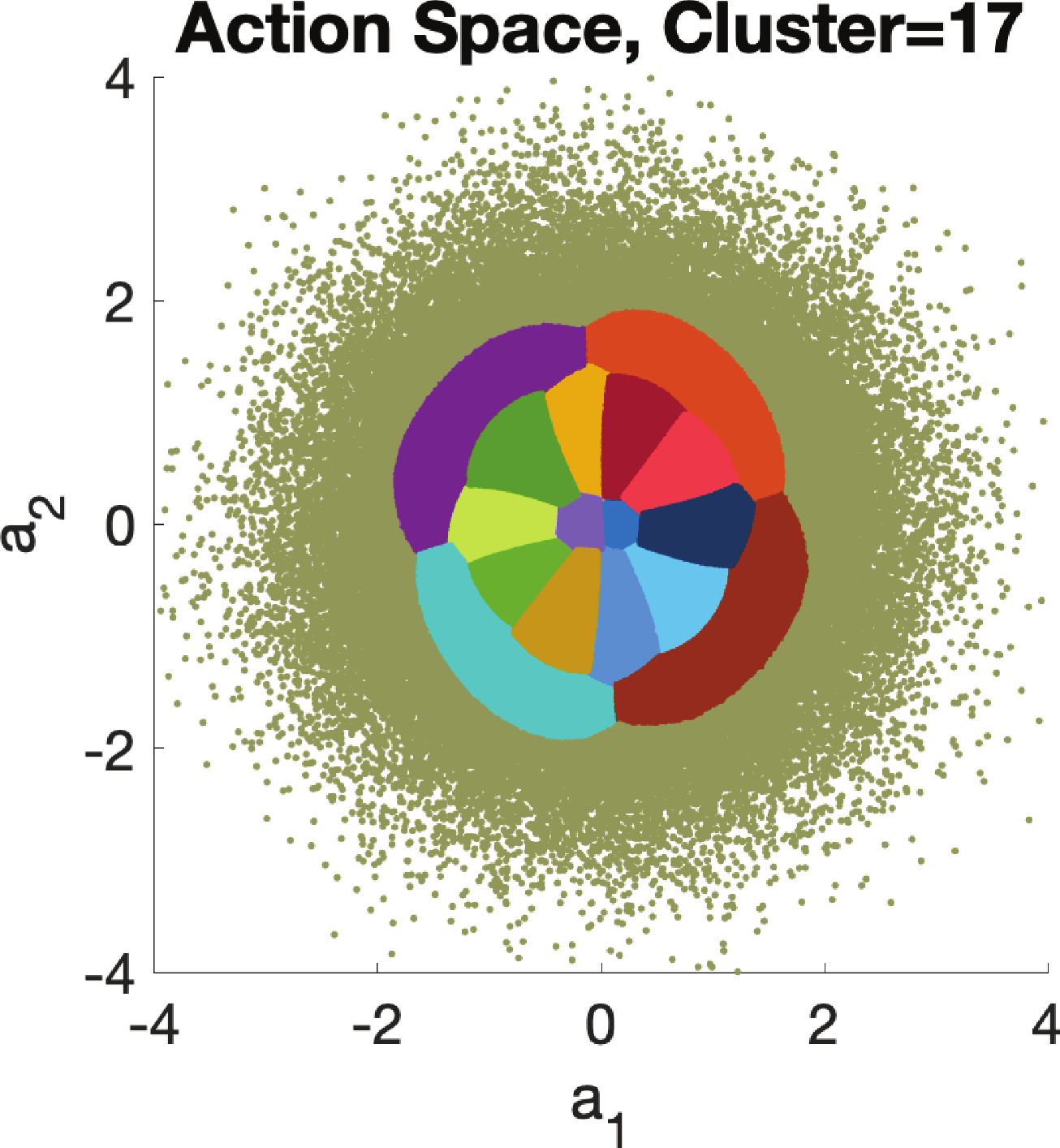}
	\caption{\footnotesize Row 1: Learned state abstractions with varying clustering sizes by top-$r$ method; Row 2: Learned action abstractions with varying clustering sizes by top-$r$ method.}
\end{figure}

\section*{E. Technical Lemmas}
\begin{lemma}
	\label{lm3}
	Suppose 
	\begin{align*}
	\frac{n/t_{mix}}{(\log(n/t_{mix}))^2}&\geq 1024\bigg(\bar{\mu}\Vert\bm{\Sigma}^{-1}\Vert_\sigma^2 + \frac{K_{max}^2}{\bar{\mu}}\bigg)\log\frac{d_St_{mix}}{\delta}.
	\end{align*}
	Then with probability $1-\delta$, we have
	\begin{align*}
	&\Vert \hat{\bm{\Sigma}}^{-1} - \bm{\Sigma}^{-1}\Vert_\sigma \leq 32\Vert \bm{\Sigma}^{-1}\Vert_\sigma^2\sqrt{\frac{\bar{\mu}\log\frac{d_St_{mix}}{\delta}(\log\frac{n}{t_{mix}})^2}{n/t_{mix}}}.
	\end{align*}
\end{lemma}

\begin{lemma}\label{decomp}
	For any tensor $\bm{X}\in\mathbb{R}^{p_1\times p_2\cdots p_N}$, such that $\textrm{Tucker-Rank}(\bm{X})\leq (r_1, r_2, \cdots, r_N)$, we can always find column-wise orthonormal matrices $\bm{U}_1\in\mathbb{R}^{p_1\times r_1},\ldots, \bm{U}_N\in\mathbb{R}^{p_N\times r_N}$ and a core tensor $\bm{C}\in\mathbb{R}^{r_1\times\cdots\times r_N}$, such that
	\begin{align*}
	\bm{X} = \bm{C} \times_1 \bm{U}_1 \times_2\cdots\times_N\bm{U}_N.
	\end{align*}
\end{lemma}

\begin{lemma}
	\label{mix}
	Suppose the worst-case mixing time of the MDP is $t_{mix}$, then for any $\varepsilon > 0$ and policy $\pi$, suppose $\nu^\pi$ is the invariant distribution of $\pi$, then for any initial distribution $\mu$, we have
	\begin{align*}
	\left\Vert \int p^{t, \pi}(\cdot\vert s_0)\mu(s_0)ds_0 - \nu^\pi(\cdot)\right\Vert_{TV} \leq \varepsilon, \forall t \geq 2t_{mix}\log\frac{1}{\varepsilon}.
	\end{align*}
\end{lemma}

\begin{lemma}
	\label{norm}
	For any given tensor $\bm{X}\in\mathbb{R}^{p_1\times p_2\times p_3}$ such that $\textrm{Tucker-Rank}(\bm{X})\leq (r_1, r_2, r_3)$, we have 
	\begin{align*}
	\Vert \bm{X}\Vert_F\leq \sqrt{\frac{r_1r_2r_3}{\max\{r_1, r_2, r_3\}}}\Vert \bm{X}\Vert_\sigma.
	\end{align*}
\end{lemma}

\begin{lemma}
	\label{net_num}
	Given $p\in\mathbb{N}, \varepsilon\in\mathbb{R}$, there always exists an $\varepsilon$-net of the sphere $S^{p-1}$ whose size is no more than $(1 + 2/\varepsilon)^p$. 
\end{lemma}

\begin{lemma}
	\label{net}
	Given a tensor $\bm{X}\in\mathbb{R}^{p_1\times p_2\times p_3}$ and three $\varepsilon$-nets of the unit sphere $\mathcal{N}_1\subset S^{p_1-1}, \mathcal{N}_2\subset S^{p_2-1}, \mathcal{N}_3\subset S^{p_3-1}$, we have
	\begin{align*}
	\Vert\bm{X}\Vert_\sigma\leq \frac{\max_{x\in\mathcal{N}_1, y\in\mathcal{N}_2, z\in\mathcal{N}_3}\vert \langle\bm{X}, x\circ y\circ z\rangle\vert}{1 - 3\varepsilon - 3\varepsilon^2 - \varepsilon^3}
	\end{align*}
\end{lemma}

\begin{lemma}\label{im_lm3}
	Suppose 
	\begin{align*}
	\frac{n/t_{mix}}{(\log(n/t_{mix}))^2}&\geq 1024\left(\bar{\mu}\Vert\bm{\Sigma}^{-1}\Vert_\sigma^2 + \frac{K_{max}^2}{\bar{\mu}}\right)\log\frac{d_St_{mix}}{\delta}.
	\end{align*}
	Then with probability $1-\delta$, we have
	\begin{align*}
	&\Vert (\hat{\bm{\Sigma}}^{-1} - \bm{\Sigma}^{-1})\bm{\Sigma}^{\frac{1}{2}}\Vert_\sigma \leq 32\Vert \bm{\Sigma}^{-1}\Vert_\sigma^{\frac{3}{2}}\sqrt{\frac{\bar{\mu}\log\frac{d_St_{mix}}{\delta}(\log\frac{n}{t_{mix}})^2}{n/t_{mix}}}.
	\end{align*}
\end{lemma}

\begin{lemma}[Concentration in tensor spectral norm] \label{th:theorem-hat_F}
	Let Assumptions \ref{assump1}-\ref{assump-data} hold.
	Suppose 
	\begin{align*}
	\frac{n/t_{mix}}{(\log (n/t_{mix}))^2} \geq 1024\frac{\kappa K_{max}^3}{\bar{\lambda}}(\log\frac{t_{mix}}{\delta} + 8(d_S + d_A)),
	\end{align*}
	then with probability $1-\delta$, we have
	\begin{align*}
	\Vert \hat{\bm{F}} -\bm{F}\Vert_\sigma \leq 64\sqrt{\frac{\kappa\bar{\lambda}(\log\frac{t_{mix}}{\delta} + 8(d_S + d_A))(\log\frac{n}{t_{mix}})^2}{n/t_{mix}}}.
	\end{align*}
\end{lemma}

\begin{lemma}\label{im_main_thm2}
	Suppose 
	\begin{align*}
	\frac{n/ t_{mix}}{(\log(n/t_{mix}))^2}&\geq 1024\left(\Vert\bm{\Sigma}^{-1}\Vert^2_\sigma \bar{\mu} + \frac{K_{max}^2}{\bar{\mu}} + \frac{\kappa K_{max}^3}{\bar{\lambda}}\right)\left(\log\frac{t_{mix}}{\delta} + 8(d_S + d_A)\right).
	\end{align*}
	Then with probability $1-\delta$, we have
	\begin{align*}
	& \Vert \hat{\bm{F}}\times_1 (\hat{\bm{\Sigma}}^{-1}\bm{\Sigma}^{1/2}) - \bm{F}\times_1(\bm{\Sigma}^{-1}\bm{\Sigma}^{1/2})\Vert_\sigma \\
	\leq & 256 \Vert\bm{\Sigma}^{-1}\Vert_\sigma^{\frac{1}{2}}\sqrt{\frac{\bar{\lambda}(\log\frac{2t_{mix}}{\delta} + d_S + d_A)(\kappa + \bar{\mu}\Vert\bm{\Sigma}^{-1}\Vert_\sigma^2)(\log\frac{n}{t_{mix}})^2}{n/t_{mix}}}.
	\end{align*}
\end{lemma}

\begin{lemma}\label{feature_sin}
	Suppose $\bm{P}, \hat{\bm{P}}$ are two order-3 tensors of the same dimension. Suppose $\bm{P} = \bm{C}\times_1 \bm{U}_1 \times_2 \bm{U}_2 \times_3 \bm{U}_3$ and $\hat{\bm{P}} = \hat{\bm{C}}\times_1 \hat{\bm{U}}_1 \times_2 \hat{\bm{U}}_2 \times_3 \hat{\bm{U}}_3$, where $\bm{C}, \hat{\bm{C}} \in \mathbb{R}^{r \times l \times m}$,
	$\bm{U}_3^\top \bm{U}_3 = \hat{\bm{U}}_3^{\top} \hat{\bm{U}}_3 = I$. We have
	\begin{align*}
	\Vert \sin\Theta(\bm{U}_3, \hat{\bm{U}}_3)\Vert_\sigma \leq   \frac{\Vert \bm{P} - \hat{\bm{P}}\Vert_\sigma}{\sigma},
	\end{align*}
	where
	\begin{align*}
	\sigma = \sup_{w\in\mathbb{R}^p, \Vert w\Vert = 1}\sigma_m(\bm{P}\times_1 w). 
	\end{align*}
\end{lemma}


\section*{F. Proofs}
\paragraph{Proof of Lemma \ref{lm1}} 

	Recall that under Assumption \ref{assump1}, there exist $c_{ijk} \in \mathbb{R}$, $u_i, w_k \in \mathcal{H}_S$, $v_j \in \mathcal{H}_A$, $i \in [r]$, $j \in [l]$, $k \in [m]$ such that \[ (\mathbb{P}f)(s,a) = \sum_{i=1}^r \sum_{j=1}^l \sum_{k=1}^m c_{ijk} u_i(s) v_j(a) \langle f, w_k \rangle_{\mathcal{H}_S}, \qquad \forall f \in \mathcal{H}_S. \]
	Let $\bm{C}\in \mathbb{R}^{r\times l\times m}$ be defined as $\bm{C}_{ijk} = c_{ijk}$. Then we can rewrite
	\begin{align*}
	&\sum_{i=1}^k\sum_{j=1}^l\sum_{k=1}^m c_{ijk}u(s)_iv(a)_j\langle f, w_k\rangle= \bm{C} \times_1 \bm{u}(s)^\top \times_2 \bm{v}(a)^\top \times_3 \langle f, \bm{w}\rangle^\top. 
	\end{align*}
	Now, because we have $u_i, w_k\in\mathcal{H}_S, v_j\in\mathcal{H}_A$, we can find three matrices $\bm{U}_1\in\mathbb{R}^{d_S\times r}, \bm{U}_2\in\mathbb{R}^{d_A\times l}, \bm{U}_3\in\mathbb{R}^{d_S\times m}$, such that
	\begin{align*}
	\bm{u} = \bm{U}_1^\top\phi, \bm{v} = \bm{U}_2^\top\psi, \bm{w} = \bm{U}_3^\top\phi. 
	\end{align*}
	Then we have
	\begin{align*}
	\bm{C} \times_1 \bm{u}(s)^\top\times_2 \bm{v}(a)^\top\times_3 \langle f, \bm{w}\rangle^\top &= \bm{C} \times_1(\bm{U}_1^\top\phi(s))^\top\times_2 (\bm{U}_2^\top\psi(a))^\top\times_3 (\bm{U}_3^\top\langle f, \phi\rangle)^\top\\
	&= (\bm{C}\times_1 \bm{U}_1 \times_2\bm{U}_2 \times_3\bm{U}_3) \times_1 \phi(s)^\top\times_2 \psi(a)^\top \times_3 \langle f, \phi\rangle^\top. 
	\end{align*}
	In particular, we take $f = \phi_i, i=1,2,\cdots, d_S$, and define $\bm{V}$ by
	\begin{align*}
	\bm{V}_{ij} = \langle \phi_i, \phi_j\rangle,
	\end{align*}
	then we have
	\begin{align*}
	\mathbb{E}[\phi(s')\vert s, a] &= (\bm{C}\times_1 \bm{U}_1 \times_2\bm{U}_2\times_3\bm{U}_3) \times_1 \phi(s)\times_2 \psi(a)^\top\times_3 \bm{V}^\top\\
	&= (\bm{C}\times_1 \bm{U}_1 \times_2\bm{U}_2\times_3(\bm{U}_3^\top \bm{V})^\top) \times_1 \phi(s)^\top\times_2 \psi(a)^\top. 
	\end{align*}
	Now, we define $\bm{P}\in\mathbb{R}^{d_S\times d_A\times d_S}$ by
	\begin{align*}
	\bm{P} = \bm{C}\times_1 \bm{U}_1\times_2\bm{U}_2\times_3 (\bm{U}_3^\top\bm{V})^\top.
	\end{align*}
	Then the Tucker-rank of $\bm{P}$ is no larger than the size of $\bm{C}$, i.e., 
	\begin{align*}
	\textrm{Tucker-Rank}(\bm{P}) \leq (r, l, m), 
	\end{align*}
	and we have
	\begin{align*}
	\mathbb{E}[\phi(s')\vert s, a] = \bm{P}\times_1 \phi(s)^\top\times_2\psi(a)^\top, 
	\end{align*}
	which finishes the proof. 

\paragraph{Proof of Lemma \ref{lm2}}
	To show this, simply notice
	\begin{align*}
	\bm{F} &= \int \phi(s)\circ \psi(a)\circ \phi(s')p(s'\vert s, a)\xi(s)\eta(a)dsdads' \\
	&= \int \phi(s)\circ \psi(a)\circ \bigg(\int\phi(s')p(s'\vert s, a)ds'\bigg)\xi(s)\eta(a) dsda\\
	&= \int \phi(s)\circ \psi(a)\circ \mathbb{E}[\phi(s')\vert s, a] \xi(s)\eta(a) dsda. 
	\end{align*}
	Now we use the notation in Lemma \ref{lm1} to write $\mathbb{E}[\phi(s')\vert s, a] = \bm{P}\times_1\phi(s)^\top\times_2\psi(a)^\top$ for some $\bm{P}\in \mathbb{R}^{d_S\times d_A\times d_S}$, and get 
	\begin{equation} \label{F=PS} \begin{aligned}
	\bm{F} &= \int \phi(s)\circ \psi(a)\circ (\bm{P}\times_1 \phi(s)^\top\times_2\psi(a)^\top)\xi(s)\eta(a) dsda\\
	&= \int \bm{P}\times_1 \left(\xi(s)\phi(s)\phi(s)^\top\right)^\top \times_2 \left(\eta(a)\psi(a)\psi(a)^\top\right)^\top dsda\\
	&=  \bm{P}\times_1 \left(\int \xi(s)\phi(s)\phi(s)^\top ds\right)^\top \times_2 \left(\int \eta(a)\psi(a)\psi(a)^\top da\right)^\top. 
	\end{aligned} \end{equation}
	The result of Lemma \ref{lm1} shows that $\textrm{Tucker-Rank}(\bm{P}) \leq (r, l, m)$, which implies that
	\begin{align*}
	\textrm{Tucker-Rank}(\bm{F}) \leq \textrm{Tucker-Rank}(\bm{P}) \leq (r, l, m). 
	\end{align*}

\paragraph{Proof of Lemma \ref{relation}}
	Notice that when $\psi$ is orthogonal with respect to $L^2(\eta)$, \eqref{F=PS} reduces to
	\begin{align*}
	\bm{F} &= \int \xi(s)\eta(a)p(s'\vert s, a)\phi(s)\circ \psi(a)\circ \phi(s')dsdads'\\
	& = \bm{P}\times_1 \left(\int \xi(s)\phi(s)\phi(s)^\top ds\right)^\top \times_2 \left(\int \eta(a)\psi(a)\psi(a)^\top da\right)^\top\\
	& = \bm{P}\times_1\bm{\Sigma},
	\end{align*}
	which implies 
	\begin{align*}
	\bm{P} = \bm{F}\times_1 \bm{\Sigma}^{-1}.
	\end{align*}

\paragraph{Proof of Lemma \ref{lm3}}
	According to the result of Lemma \ref{im_lm3}, we know that when 
	\begin{align*}
	\frac{n/t_{mix}}{(\log(n/t_{mix}))^2}&\geq 1024\left(\bar{\mu}\Vert\bm{\Sigma}^{-1}\Vert_\sigma^2 + \frac{K_{max}^2}{\bar{\mu}}\right)\log\frac{d_St_{mix}}{\delta}.
	\end{align*}
	Then with probability $1-\delta$, we have
	\begin{align*}
	&\left\Vert (\hat{\bm{\Sigma}}^{-1} - \bm{\Sigma}^{-1})\bm{\Sigma}^{1/2}\right\Vert_\sigma \leq 32\Vert \bm{\Sigma}^{-1}\Vert_\sigma^{\frac{3}{2}}\sqrt{\frac{\bar{\mu}\log\frac{d_St_{mix}}{\delta}(\log\frac{n}{t_{mix}})^2}{n/t_{mix}}}. 
	\end{align*}
	Therefore, we can directly get
	\begin{align*}
	\Vert \hat{\bm{\Sigma}}^{-1} - \bm{\Sigma}^{-1}\Vert_\sigma &= \Vert (\hat{\bm{\Sigma}}^{-1} - \bm{\Sigma}^{-1})\bm{\Sigma}^{\frac{1}{2}}\bm{\Sigma}^{-\frac{1}{2}}\Vert_\sigma \\
	&\leq \Vert (\hat{\bm{\Sigma}}^{-1} - \bm{\Sigma}^{-1})\bm{\Sigma}^{\frac{1}{2}}\Vert_\sigma\Vert\bm{\Sigma}^{-\frac{1}{2}}\Vert_\sigma \\
	&\leq 32\Vert \bm{\Sigma}^{-1}\Vert_\sigma^2\sqrt{\frac{\bar{\mu}\log\frac{d_St_{mix}}{\delta}(\log\frac{n}{t_{mix}})^2}{n/t_{mix}}}, 
	\end{align*}
	which finishes the proof. 

\paragraph{Proof of Lemma \ref{decomp}}
	Suppose the SVD of $\mathcal{M}_1(\bm{X})$ is 
	\begin{align*}
	\mathcal{M}_1(\bm{X}) = \bm{U}_1 \bm{\Sigma}\bm{V}^\top,
	\end{align*}
	where $\bm{U}_1\in\mathbb{R}^{p_1\times r_1}$ is a column-wise orthonormal matrix. Therefore, 
	\begin{align*}
	\mathcal{M}_1(\bm{X}) =  \bm{U}_1(\bm{U}_1^\top \mathcal{M}_1(\bm{X})), 
	\end{align*}
	which is equivalent to
	\begin{align*}
	\bm{X} = (\bm{X}\times_1 \bm{U}_1^\top)\times_1 \bm{U}_1. 
	\end{align*}
	Let $\bm{X}_1 = \bm{X}\times_1 \bm{U}_1^\top$, with a similar procedure we can find some column-wise orthogonal matrix $\bm{U}_2$, such that
	\begin{align*}
	\bm{X}_1 = (\bm{X}_1\times_2 \bm{U}_2^\top)\times_2 \bm{U}_2. 
	\end{align*}
	Repeating this process for $N$ times, we can find a series of column orthogonal matrix $\bm{U}_1, \bm{U}_2, \cdots, \bm{U}_N$ and a core tensor $\bm{C} = \bm{X}_N$, such that
	\begin{align*}
	\bm{X} = \bm{C}\times_1\bm{U}_1 \cdots \times_N\bm{U}_N,
	\end{align*}
	which has finished the proof. 


\paragraph{Proof of Lemma \ref{mix}}
	Let $\alpha = t_{mix}$, for any initial distribution $\mu$, we use the notation
	\begin{align*}
	p^{t, \pi}(\cdot\vert \mu) := \int p^{t, \pi}(\cdot\vert s)\mu(s)ds
	\end{align*}
	to denote the state distribution after $t$ steps starting from initial state distribution $\mu$. One direct fact is
	\begin{align*}
	\Vert p^{\alpha, \pi}(\cdot\vert \mu) - \nu^\pi(\cdot)\Vert_{TV} &= \frac{1}{2}\int \bigg\vert  \int p^{t, \pi}(s\vert s_0)\mu(s_0)ds_0 - \nu^\pi(s)\bigg\vert ds\\
	&= \frac{1}{2} \int \bigg\vert  \int (p^{t, \pi}(s\vert s_0) - \nu^\pi(s))\mu(s_0)ds_0 \bigg\vert ds\\
	& \leq \int \mu(s_0) \bigg(\frac{1}{2}\int \vert   p^{t, \pi}(s\vert s_0)- \nu^\pi(s)\vert ds \bigg)ds_0\\
	&\leq  \frac{1}{4}\int \mu(s_0)ds_0 = \frac{1}{4}. 
	\end{align*}
	Now,  for any initial distribution $\mu$ and any $n\geq 2\alpha$, we have 
	\begin{align*}
	& \Vert p^{n, \pi}(\cdot\vert \mu) - \nu^\pi(\cdot)\Vert_{TV} = \frac{1}{2}\int \left\vert \int p^{n, \pi}(s\vert s_0)\mu(s_0)ds_0- \nu^\pi(s)\right\vert ds\\
	= &\frac{1}{2}\int \left\vert \int p^{n-\alpha, \pi}(s\vert s_1)\left[\int p^{\alpha, \pi}(s_1\vert s_0)\mu(s_0)ds_0- \nu^\pi(s_1)\right]ds_1\right\vert ds\\
	= &\frac{1}{2}\int \left\vert \int p^{n-\alpha, \pi}(s\vert s_1)\left[(p^{\alpha, \pi}(s_1\vert \mu)- \nu^\pi(s_1))_+ - ( p^{\alpha, \pi}(s_1\vert \mu)- \nu^\pi(s_1))_-\right]ds_1\right\vert ds\\
	= &\frac{1}{2}\int \bigg\vert \int p^{n-\alpha, \pi}(s\vert s_1)\bigg[(p^{\alpha, \pi}(s_1\vert \mu)- \nu^\pi(s_1))_+ - \left(\int ( p^{\alpha, \pi}(\tilde{s}\vert \mu)- \nu^\pi(\tilde{s}))_+d\tilde{s}\right)\nu^\pi(s_1)\\
	& + \frac{1}{2} \left(\int (p^{\alpha, \pi}(\tilde{s}\vert \mu)- \nu^\pi(\tilde{s}))_-d\tilde{s}\right)\nu^\pi(s_1) - (p^{\alpha, \pi}(s_1\vert\mu)- \nu^\pi(s_1))_-\bigg]ds_1\bigg\vert ds, 
	\end{align*}
	where we use the relation
	\begin{align*}
	\int (p^{\alpha, \pi}(\tilde{s}\vert \mu)- \nu^\pi(\tilde{s}))_+d\tilde{s} =   \int (p^{\alpha, \pi}(\tilde{s}\vert \mu)- \nu^\pi(\tilde{s}))_-d\tilde{s}, 
	\end{align*}
	because we have
	\begin{align*}
	0 = \int (p^{\alpha, \pi}(\tilde{s}\vert \mu) - \nu^\pi(\tilde{s}))d\tilde{s} =   \int (p^{\alpha, \pi}(\tilde{s}\vert \mu)- \nu^\pi(\tilde{s}))_+d\tilde{s} - \int (p^{\alpha, \pi}(\tilde{s}\vert \mu) - \nu^\pi(\tilde{s}))_-d\tilde{s}. 
	\end{align*}
	Therefore, we get
	\begin{align*}
	&\Vert p^{n, \pi}(\cdot\vert \mu) - \nu^\pi(\cdot)\Vert_{TV} \\
	\leq&\frac{1}{2}\int \left\vert \int p^{n-\alpha, \pi}(s\vert s_1)\left[(p^{\alpha, \pi}(s_1\vert \mu)- \nu^\pi(s_1))_+ - \left(\int (p^{\alpha, \pi}(\tilde{s}\vert \mu)- \nu^\pi(\tilde{s}))_+d\tilde{s}\right)\nu^\pi(s_1)\right]ds_1\right\vert ds\\
	+&\frac{1}{2}\int\left\vert  \int p^{n-\alpha, \pi}(s\vert s_1)\left[\left(\int(p^{\alpha, \pi}(\tilde{s}\vert \mu)- \nu^\pi(\tilde{s}))_-d\tilde{s}\right)\nu^\pi(s_1) - (p^{\alpha, \pi}(s_1\vert \mu)- \nu^\pi(s_1))_-\right]ds_1\right\vert ds .
	\end{align*}
	For the first term, note that
	\begin{align*}
	&\int \left\vert \int p^{n-\alpha, \pi}(s\vert s_1)\left[(p^{\alpha, \pi}(s_1\vert \mu)- \nu^\pi(s_1))_+ - \left(\int (p^{\alpha, \pi}(\tilde{s}\vert \mu)- \nu^\pi(\tilde{s}))_+d\tilde{s}\right)\nu^\pi(s_1)\right]ds_1\right\vert ds \\
	=&\int (p^{\alpha, \pi}(\tilde{s}\vert \mu)- \nu^\pi(\tilde{s}))_+d\tilde{s}\cdot\int \left\vert \int p^{n-\alpha, \pi}(s\vert s_1)\left[\frac{(p^{\alpha, \pi}(s_1\vert\mu)- \nu^\pi(s_1))_+}{\int (p^{\alpha, \pi}(\tilde{s}\vert \mu)- \nu^\pi(\tilde{s}))_+d\tilde{s}} - \nu^\pi(s_1)\right]ds_1\right\vert ds \\
	=&\int (p^{\alpha, \pi}(\tilde{s}\vert \mu)- \nu^\pi(\tilde{s}))_+d\tilde{s}\cdot\int \left\vert \int p^{n-\alpha, \pi}(s\vert s_1)\frac{(p^{\alpha, \pi}(s_1\vert \mu)- \nu^\pi(s_1))_+}{\int (p^{\alpha, \pi}(\tilde{s}\vert \mu)- \nu^\pi(\tilde{s}))_+d\tilde{s}}ds_1 - \nu^\pi(s)\right\vert ds .
	\end{align*}
	Note that 
	\begin{align*}
	\frac{(p^{\alpha, \pi}(\cdot\vert \mu)- \nu^\pi(\cdot))_+}{\int (p^{\alpha, \pi}(\tilde{s}\vert \mu)- \nu^\pi(\tilde{s}))_+d\tilde{s}}
	\end{align*}
	is also a probability density of some initial distribution, and $n-\alpha \geq \alpha$, so we have
	\begin{align*}
	&\int \left\vert \int p^{\alpha, \pi}(s\vert s_1)\left[(p^{\alpha, \pi}(s_1\vert \mu)- \nu^\pi(s_1))_+ - \int (p^{\alpha, \pi}(\tilde{s}\vert \mu)- \nu^\pi(\tilde{s}))_+d\tilde{s})\nu^\pi(s_1)\right]ds_1\right\vert ds \\
	\leq&\frac{1}{2}\int (p^{\alpha, \pi}(\tilde{s}\vert \mu)- \nu^\pi(\tilde{s}))_+d\tilde{s}.
	\end{align*}
	Similarly, we also get
	\begin{align*}
	&\int \left\vert \int p^{\alpha, \pi}(s\vert s_1)\left[(p^{\alpha, \pi}(s_1\vert \mu)- \nu^\pi(s_1))_- - \int (p^{\alpha, \pi}(\tilde{s}\vert\mu)- \nu^\pi(\tilde{s}))_+d\tilde{s})\nu^\pi(s_1)\right]ds_1\right\vert ds \\
	\leq&\frac{1}{2}\int (p^{\alpha, \pi}(\tilde{s}\vert \mu)- \nu^\pi(\tilde{s}))_-d\tilde{s},
	\end{align*}
	which implies that 
	\begin{align*}
	\Vert p^{n, \pi}(\cdot\vert \mu) - \nu^\pi(\cdot)\Vert_{TV}  &\leq \frac{1}{4}\left(\int ( p^{\alpha, \pi}(\tilde{s}\vert \mu)- \nu^\pi(\tilde{s}))_+d\tilde{s} + \int (p^{\alpha, \pi}(\tilde{s}\vert \mu)- \nu^\pi(\tilde{s}))_-d\tilde{s}\right)\\
	&=\frac{1}{4}\left(\int \vert p^{\alpha, \pi}(\tilde{s}\vert \mu)- \nu^\pi(\tilde{s})\vert d\tilde{s}\right) \leq \frac{1}{4} \cdot \frac{1}{2}. 
	\end{align*}
	By induction, we can prove that for any $n\geq k\alpha$, we have
	\begin{align*}
	\Vert p^{n, \pi}(\cdot\vert \mu) - \nu^\pi(\cdot)\Vert_{TV} \leq \frac{1}{4}(\frac{1}{2})^{k-1}. 
	\end{align*}
	Therefore, for any $0 < \varepsilon < e^{-1}$, let $k = \lceil 2\log \frac{1}{\varepsilon}\rceil + 1$, then for any $n\geq 2\alpha\log\frac{1}{\varepsilon} \geq (k-1)\alpha$, we have
	\begin{align*}
	\Vert p^{n, \pi}(\cdot\vert \mu) - \nu^\pi(\cdot)\Vert_{TV} \leq \frac{1}{4}(\frac{1}{2})^{k-2} \leq \varepsilon,
	\end{align*}
	which has finished the proof. 

\paragraph{Proof of Lemma \ref{norm}}
	Without loss of generality, assume that $r_3 = \max\{r_1, r_2, r_3\}$. Since $\textrm{Tucker-Rank}(\bm{X})\leq (r_1, r_2, r_3)$, according to the result of Lemma \ref{decomp}, there exists a decomposition of $\bm{X}$
	\begin{align*}
	\bm{X} = \bm{G}\times_1\bm{U}, \bm{G}\in\mathbb{R}^{r_1\times p_2\times p_3}, \bm{U}\in\mathbb{R}^{p_1\times r_1},
	\end{align*}
	where $\bm{U}$ is a column-wise orthonormal matrix, i.e., $\bm{U}^\top \bm{U} = \bm{I}_{r_1}$. This formulation implies $\textrm{Tucker-Rank}(\bm{G})\leq (r_1, r_2, r_3)$, therefore for each $1\leq i\leq r_1$, we have $\textrm{rank}(\bm{G}_{i::})\leq \min\{r_2, r_3\} = r_2$. We then consider the SVD of $\bm{G}_{i::}$, 
	\begin{align*}
	\bm{G}_{i::} = \bm{V}_i\bm{\Lambda}_i\bm{W}_i^\top,
	\end{align*}
	where $\bm{\Lambda}_i=\textrm{diag}(\lambda_{i1}, \lambda_{i2},\ldots,\lambda_{ir_2}), \bm{V}_i\in\mathbb{R}^{p_2\times r_2}, \bm{W}_i\in\mathbb{R}^{p_3\times r_2}, \bm{V}_i^\top \bm{V}_i = \bm{W}_i^\top \bm{W}_i=\bm{I}_{r_2}$. The above formulation is equivalent to
	\begin{align*}
	\bm{X} = \sum_{i=1}^{r_1}\sum_{j=1}^{r_2}\lambda_{ij} \bm{u}_{i}\circ \bm{v}_{ij}\circ \bm{w}_{ij},
	\end{align*}
	where $\bm{v}_{ij}$ is the $j$th column of $\bm{V}_i$, $\bm{w}_{ij}$ is the $j$th column of $\bm{W}_i$, $\bm{u}_{i}$ is the $i$th column of $\bm{U}$. According to the definition of $\Vert\cdot\Vert_\sigma$, we have 
	\begin{align*}
	\lambda_{ij} = \bigg\langle \sum_{i=1}^{r_1}\sum_{j=1}^{r_2}\lambda_{ij} \bm{u}_{i}\circ \bm{v}_{ij}\circ \bm{w}_{ij}, \bm{u}_{i}\circ \bm{v}_{ij}\circ \bm{w}_{ij}\bigg\rangle \leq \Vert \bm{X}\Vert_\sigma\\
	-\lambda_{ij} = \bigg\langle \sum_{i=1}^{r_1}\sum_{j=1}^{r_2}\lambda_{ij} \bm{u}_{i}\circ \bm{v}_{ij}\circ \bm{w}_{ij}, (-\bm{u}_{i})\circ \bm{v}_{ij}\circ \bm{w}_{ij}\bigg\rangle \leq \Vert \bm{X}\Vert_\sigma.
	\end{align*}
	So we get
	\begin{align*}
	\vert\lambda_{ij}\vert^2 \leq \Vert \bm{X}\Vert_\sigma^2, \forall i, j.
	\end{align*}
	On the other hand, 
	\begin{align*}
	\Vert \bm{X}\Vert_F^2 &= \sum_{a=1}^{p_1}\sum_{b=1}^{p_2}\sum_{c=1}^{p_3}(\sum_{i=1}^{r_1}\sum_{j=1}^{r_2}\lambda_{ij}\bm{u}_{ia}\bm{v}_{ijb}\bm{w}_{ijc})^2 = \sum_{a=1}^{p_1}\sum_{b=1}^{p_2}\sum_{c=1}^{p_3}\sum_{i=1}^{r_1} \bm{u}_{ia}^2(\sum_{j=1}^{r_2}\lambda_{ij}\bm{v}_{ijb}\bm{w}_{ijc})^2\\
	&= \sum_{i=1}^{r_1}\sum_{b=1}^{p_2}\sum_{c=1}^{p_3}(\sum_{j=1}^{r_2}\lambda_{ij}\bm{v}_{ijb}\bm{w}_{ijc})^2 = \sum_{i=1}^{r_1}\sum_{j=1}^{r_2}\sum_{b=1}^{p_2}\sum_{c=1}^{p_3}\bm{v}_{ijb}^2\bm{w}_{ijc}^2\lambda_{ij}^2\\
	&= \sum_{i=1}^{r_1}\sum_{j=1}^{r_2}\lambda_{ij}^2\leq r_1r_2\Vert \bm{X}\Vert_\sigma^2,
	\end{align*}
	which implies
	\begin{align*}
	\Vert \bm{X}\Vert_F \leq \sqrt{r_1r_2}\Vert \bm{X}\Vert_\sigma = \sqrt{\frac{r_1r_2r_3}{\max\{r_1, r_2, r_3\}}}\Vert \bm{X}\Vert_\sigma. 
	\end{align*} 

\paragraph{Proof of Lemma \ref{net_num}}
	The conclusion can be directly derived from Corollary 4.2.13 in \cite{vershynin2017high}. 

\paragraph{Proof of Lemma \ref{net}}
	According to the definition of $\Vert\cdot\Vert_\sigma$, we can always find $x_0\in S^{p_1-1}, y_0\in S^{p_2-1}, z_0\in S^{p_3-1}$ such that
	\begin{align*}
	\langle \bm{X}, x_0\circ y_0\circ z_0\rangle = \Vert\bm{X}\Vert_\sigma
	\end{align*}
	Then according to the definition of $\varepsilon$-net, we can always find $x, y, z$ from these $\varepsilon$-nets such that $\Vert x - x_0\Vert_2\leq \varepsilon, \Vert y - y_0\Vert_2\leq \varepsilon, \Vert z - z_0\Vert_2\leq \varepsilon$, then
	\begin{align*}
	&\vert \langle \bm{X}, x_0\circ y_0\circ z_0\rangle - \langle \bm{X}, x\circ y\circ z\rangle \vert \\
	\leq &\vert \langle \bm{X}, (x_0 - x)\circ y_0\circ z_0\rangle\vert + \vert \langle \bm{X}, x\circ (y_0 - y)\circ z_0\rangle + \vert\vert \langle \bm{X}, x_0\circ y_0\circ (z_0-z)\rangle\vert\\
	+ &\vert \langle \bm{X}, (x_0 - x)\circ (y_0 - y)\circ z_0\rangle\vert + \vert \langle \bm{X}, (x_0 - x)\circ y_0\circ (z_0-z)\rangle\vert\\
	+ &\vert \langle \bm{X}, x_0\circ (y_0-y)\circ (z_0-z)\rangle\vert + \vert \langle \bm{X}, (x_0 - x)\circ (y_0-y)\circ (z_0-z)\rangle\vert\\
	\leq &\Vert\bm{X}\Vert_\sigma (\Vert x_0 - x\Vert_2 + \Vert y_0 - y\Vert_2 + \Vert z_0 - z\Vert_2\\
	+ &\Vert x_0 - x\Vert_2\Vert y_0 - y\Vert_2 + \Vert x_0 - x\Vert_2\Vert z_0 - z\Vert_2 + \Vert y_0 - y\Vert_2\Vert z_0 - z\Vert_2 \\
	+& \Vert x_0 - x\Vert_2\Vert y_0 - y\Vert_2\Vert z_0 - z\Vert_2)\\
	\leq & \Vert\bm{X}\Vert_\sigma(3\varepsilon + 3\varepsilon^2 + \varepsilon^3),
	\end{align*}
	which implies 
	\begin{align*}
	&\Vert\bm{X}\Vert_\sigma \leq \max_{x\in\mathcal{N}_1, y\in\mathcal{N}_2, z\in\mathcal{N}_3}\vert\langle \bm{X}, x\circ y\circ z\rangle\vert + \Vert\bm{X}\Vert_\sigma(3\varepsilon + 3\varepsilon^2 + \varepsilon^3)\\
	\Rightarrow&\Vert\bm{X}\Vert_\sigma \leq \frac{\max_{x\in\mathcal{N}_1, y\in\mathcal{N}_2, z\in\mathcal{N}_3}\vert\langle \bm{X}, x\circ y\circ z\rangle\vert}{1 - 3\varepsilon - 3\varepsilon^2 - \varepsilon^3},
	\end{align*}
	which has finished the proof.

\paragraph{Proof of Lemma \ref{im_lm3}}
	\textbf{Step 1:}
	
	Let $\bm{H}_i = \phi(s_i)\phi(s_i)^\top$. We introduce some sufficiently large integer $\alpha$ such that from any initial distribution $\mu$, one always has
	\begin{align*}
	\Vert p^{\alpha, \pi}(s\vert \mu) - \xi(s)\Vert_{TV} \leq \frac{\bar{\mu}}{2K_{max}^2}\wedge\frac{t}{2K_{max}}.
	\end{align*}
	By Lemma \ref{mix}, we can simply choose $\alpha = \lceil 2t_{mix}\log \frac{4K_{max}^3}{\bar{\mu}t}\rceil + 1$ to satisfy this condition. For each $0\leq l\leq \alpha-1$ and $1\leq k\leq n_l = \lceil\frac{n - l}{\alpha}\rceil$, we define $\bm{H}_{k}^l$ as $\bm{H}_{k\alpha + l}$. We also denote $\bm{H}$ as a random matrix independent with $\bm{H}_i, 1\leq i\leq n$, which is defined as
	\begin{align*}
	\bm{H} = \mathbb{E}_{S \sim \xi}\big[\phi(S)\phi(S)^\top\big]. 
	\end{align*}
	Denote $\mathcal{F}_i$ as the $\sigma$-algebra generated by the history up to step $i$. Then we have
	\begin{align*}
	\Vert \bm{H}_k^l - \mathbb{E}[\bm{H}_k^l\vert \mathcal{F}_{(k-1)\alpha + l}]\Vert_\sigma &\leq \Vert \bm{H}_k^l \Vert_\sigma +\Vert \mathbb{E}[\bm{H}_k^l\vert \mathcal{F}_{(k-1)\alpha + l}]\Vert_\sigma \\
	&\leq \Vert \phi(s_{k\alpha+l})\Vert^2 + \mathbb{E}[\Vert \phi(s_{k\alpha+l})\Vert^2\vert \mathcal{F}_{(k-1)\alpha + l}]\\
	&\leq 2K_{max}
	\end{align*}
	and
	\begin{align*}
	\Vert\mathbb{E}[\bm{H}_{k}^l\bm{H}_{k}^l\vert \mathcal{F}_{(k-1)\alpha + l}]\Vert_\sigma &\leq \Vert\mathbb{E}[\bm{H}_{k}^l\bm{H}_{k}^l - \mathbb{E}[\bm{H}\bm{H}]\vert \mathcal{F}_{(k-1)\alpha + l}]\Vert_\sigma + \Vert \mathbb{E}[\bm{H}\bm{H}]\Vert_\sigma\\
	&= \Vert\mathbb{E}[\bm{H}_{k}^l\bm{H}_{k}^l - \mathbb{E}[\bm{H}\bm{H}]\vert \mathcal{F}_{(k-1)\alpha + l}]\Vert_\sigma +\bar{\mu}\\
	&= \left\Vert \int \Vert\phi(s)\Vert^2\phi(s)\phi(s)^\top(p^{\alpha, \pi}(s\vert \mu_{(k-1)\alpha + l}) - \xi(s))ds\right\Vert_\sigma + \bar{\mu}\\
	&\leq 2K_{max}^2\Vert p^{\alpha, \pi}(s\vert \mu_{(k-1)\alpha + l}) - \xi(s)\Vert_{TV} + \bar{\mu} \leq 2\bar{\mu}, 
	\end{align*}
	where $\mu_{i}$ is the state distribution at step $i$. We then have
	\begin{align*}
	\Vert\mathbb{E}[(\bm{H}_{k}^l - \mathbb{E}[\bm{H}_k^l\vert \mathcal{F}_{(k-1)\alpha+l}])(\bm{H}_{k}^l - \mathbb{E}[\bm{H}_k^l\vert \mathcal{F}_{(k-1)\alpha+l}])\vert \mathcal{F}_{(k-1)\alpha + l}]\Vert_\sigma\leq \Vert\mathbb{E}[\bm{H}_{k}^l\bm{H}_{k}^l\vert \mathcal{F}_{(k-1)\alpha + l}]\Vert_\sigma\leq 2\bar{\mu}.
	\end{align*}
	So according to the martingale version of matrix Bernstein's inequality (See e.g. \cite{tropp2011freedman}), we have
	\begin{align*}
	\mathbb{P}\left(\bigg\Vert \frac{1}{n_l}\sum_{k=1}^{n_l}(\bm{H}_k^l - \mathbb{E}[\bm{H}_k^l\vert\mathcal{F}_{(k-1)\alpha + l}])\bigg\Vert_\sigma > t\right) \leq 2d_Se^{-\frac{1}{2}\frac{n_lt^2}{2\bar{\mu} + 2K_{max}t/3}}.
	\end{align*}
	
	\textbf{Step 2:}
	
	We have
	\begin{align*}
	\Vert \mathbb{E}[\bm{H}_k^l\vert\mathcal{F}_{(k -1)\alpha + l}] - \bm{H}\Vert_\sigma &= \bigg\Vert \int \phi(s)\phi(s)^\top(p^{\alpha, \pi}(s\vert\mu_{(k-1)\alpha + l}) - \xi(s))ds\bigg\Vert_\sigma\\
	&\leq 2K_{max}\Vert p^{\alpha, \pi}(s\vert\mu_{(k-1)\alpha + l}) - \xi(s)\Vert_{TV} \leq t,
	\end{align*}
	which implies
	\begin{align*}
	\mathbb{P}\left(\bigg\Vert \frac{1}{n_l}\sum_{k=1}^{n_l}(\bm{H}_k^l - \bm{H})\bigg\Vert_\sigma > 2t\right) \leq 2d_Se^{-\frac{1}{2}\frac{n_lt^2}{2\bar{\mu} + 2K_{max}t/3}}.
	\end{align*}
	We then use a union bound to get
	\begin{align*}
	\mathbb{P}\left(\exists l, \bigg\Vert \frac{1}{n_l}\sum_{k=1}^{n_l}(\bm{H}_k^l - \bm{H})\bigg\Vert_\sigma > 2t\right) &=\mathbb{P}\left(\bigcup_{l=0}^{\alpha-1}\bigg\{\bigg\Vert \frac{1}{n_l}\sum_{k=1}^{n_l}(\bm{H}_k^l - \bm{H})\bigg\Vert_\sigma > 2t\bigg\}\right)  \\
	&\leq \sum_{l=0}^{\alpha-1}\mathbb{P}\left(\bigg\Vert \frac{1}{n_l}\sum_{k=1}^{n_l}(\bm{H}_k^l - \bm{H})\bigg\Vert_\sigma > 2t\right)  \\
	&\leq \sum_{l=0}^{\alpha-1}2d_Se^{-\frac{1}{2}\frac{n_lt^2}{2\bar{\mu} + 2K_{max}t/3}}\\
	&\leq 2\alpha d_Se^{-\frac{1}{2}\frac{(n-2\alpha)t^2}{2\alpha(\bar{\mu} + K_{max}t/3)}},
	\end{align*}
	where we use the fact that $n_l = \lceil\frac{n - l}{\alpha}\rceil \geq \frac{n}{\alpha} - 2$. Therefore, we get
	\begin{align*}
	\mathbb{P}(\Vert\hat{\bm{\Sigma}} - \bm{\Sigma}\Vert_\sigma > 2t) &= \mathbb{P}\left(\bigg\Vert\frac{1}{n}\sum_{k=1}^n\bm{H}_k - \bm{H}\bigg\Vert_\sigma > 2t\right)\\
	&\leq \mathbb{P}\left(\exists l, \bigg\Vert \frac{1}{n_l}\sum_{k=1}^{n_l}(\bm{H}_k^l - \bm{H})\bigg\Vert_\sigma > 2t\right) \\
	&\leq 2\alpha d_Se^{-\frac{1}{2}\frac{(n-2\alpha)t^2}{2\alpha(\bar{\mu} + K_{max}t/3)}}.
	\end{align*}
	Replacing $2t$ by $t$, we get
	\begin{align*}
	\mathbb{P}(\Vert\hat{\bm{\Sigma}} - \bm{\Sigma}\Vert_\sigma > t) \leq  2\alpha d_Se^{-\frac{1}{16}\frac{(n-2\alpha)t^2}{\alpha(\bar{\mu} + K_{max}t/6)}}.
	\end{align*}
	Now we assume
	\begin{align*}
	\frac{n/t_{mix}}{(\log (n/t_{mix}))^2} \geq 256\frac{K_{max}^2}{\bar{\mu}}\log\frac{d_St_{mix}}{\delta}
	\end{align*}
	and take
	\begin{align*}
	t = \sqrt{\frac{256\bar{\mu}\log\frac{d_St_{mix}}{\delta}(\log\frac{n}{t_{mix}})^2}{n/t_{mix}}}.
	\end{align*}
	Then we have
	\begin{align*}
	\alpha &\leq 4t_{mix}\log\frac{4K_{max}^3}{\bar{\mu}t} = 4t_{mix}\log \Bigg(\frac{K_{max}^{3}}{\bar{\mu}^{3/2}}\sqrt{\frac{n/t_{mix}}{16\log \frac{d_St_{mix}}{\delta}(\log\frac{n}{t_{mix}})^2}}\Bigg) \\
	&\leq 4t_{mix}\log \Bigg(\frac{K_{max}^{3}}{\bar{\mu}^{3/2}}\sqrt{\frac{n/t_{mix}}{16\log\frac{n}{t_{mix}}}}\Bigg) \leq 8t_{mix}\log\frac{n/t_{mix}}{\log\frac{n}{t_{mix}}} \leq 8t_{mix}\log \frac{n}{t_{mix}} \leq \frac{1}{4}n.
	\end{align*}
	Meanwhile, 
	\begin{align*}
	\frac{1}{6}K_{max}t = \frac{1}{6}K_{max}\sqrt{\frac{256\bar{\mu}\log\frac{d_St_{mix}}{\delta }(\log\frac{n}{t_{mix}})^2}{n/t_{mix}}} \leq \bar{\mu},
	\end{align*}
	so we have
	\begin{align*}
	\mathbb{P}(\Vert\hat{\bm{\Sigma}} - \bm{\Sigma}\Vert_\sigma > t) &\leq  2\alpha d_Se^{-\frac{1}{16}\frac{(n-2\alpha)t^2}{\alpha(\bar{\mu} + K_{max}t/6)}} \leq 16t_{mix}d_S\log\frac{n}{t_{mix}}e^{-\frac{1}{256}\frac{nt^2}{t_{mix}\bar{\mu}\log\frac{n}{t_{mix}}}}\\
	&= 16t_{mix}d_S\log\frac{n}{t_{mix}}e^{-\log\frac{d_St_{mix}}{\delta}\log\frac{n}{t_{mix}}}\\
	&\leq 16t_{mix}d_S\log\frac{n}{t_{mix}}e^{-(\log\frac{d_St_{mix}}{\delta } + \log\frac{n}{t_{mix}})}\\
	&= \frac{16\delta\log\frac{n}{t_{mix}}}{n/t_{mix}} \leq \delta,
	\end{align*}
	i.e., with probability at least $1-\delta$, we have
	\begin{align*}
	\Vert \hat{\bm{\Sigma}} - \bm{\Sigma}\Vert_\sigma \leq  \sqrt{\frac{256\bar{\mu}\log\frac{d_St_{mix}}{\delta}(\log\frac{n}{t_{mix}})^2}{n/t_{mix}}}.
	\end{align*}
	
	\textbf{Step 3:}
	
	Notice the relation 
	\begin{align*}
	\Vert (\hat{\bm{\Sigma}}^{-1} - \bm{\Sigma}^{-1})\bm{\Sigma}^{\frac{1}{2}}\Vert_\sigma& \leq \Vert \bm{\Sigma}^{-1}\Vert_\sigma\Vert \hat{\bm{\Sigma}} - \bm{\Sigma}\Vert_\sigma \Vert\hat{\bm{\Sigma}}^{-1}\bm{\Sigma}^{\frac{1}{2}}\Vert_\sigma \\
	&\leq  \big(\Vert (\bm{\Sigma}^{-1} - \hat{\bm{\Sigma}}^{-1})\bm{\Sigma}^{\frac{1}{2}}\Vert_\sigma + \Vert\bm{\Sigma}^{-\frac{1}{2}}\Vert_\sigma\big)\Vert \bm{\Sigma}^{-1}\Vert_\sigma\Vert \hat{\bm{\Sigma}} - \bm{\Sigma}\Vert_\sigma,
	\end{align*}
	i.e., 
	\begin{align*}
	\Vert (\hat{\bm{\Sigma}}^{-1} - \bm{\Sigma}^{-1})\bm{\Sigma}^{\frac{1}{2}}\Vert_\sigma  \leq \frac{\Vert\bm{\Sigma}^{-1}\Vert_\sigma^{\frac{3}{2}}\Vert \hat{\bm{\Sigma}} - \bm{\Sigma}\Vert_\sigma}{1 - \Vert\bm{\Sigma}^{-1}\Vert_\sigma\Vert \hat{\bm{\Sigma}} - \bm{\Sigma}\Vert_\sigma }.
	\end{align*}
	Therefore, if
	\begin{align*}
	\frac{n/t_{mix}}{(\log n/t_{mix})^2}\geq 1024\bar{\mu}\Vert\bm{\Sigma}^{-1}\Vert_\sigma^2\log\frac{d_St_{mix}}{\delta},
	\end{align*}
	then we have
	\begin{align*}
	\Vert\bm{\Sigma}^{-1}\Vert_\sigma\Vert \hat{\bm{\Sigma}} - \bm{\Sigma}\Vert_\sigma \leq \frac{1}{2}
	\end{align*}
	and
	\begin{align*}
	\Vert (\hat{\bm{\Sigma}}^{-1} - \bm{\Sigma}^{-1})\bm{\Sigma}^{\frac{1}{2}}\Vert_\sigma \leq 2\Vert \bm{\Sigma}^{-1}\Vert_\sigma^{\frac{3}{2}}\Vert \hat{\bm{\Sigma}} - \bm{\Sigma}\Vert_\sigma \leq 32\Vert \bm{\Sigma}^{-1}\Vert_\sigma^{\frac{3}{2}}\sqrt{\frac{\bar{\mu}\log\frac{d_St_{mix}}{\delta}(\log\frac{n}{t_{mix}})^2}{n/t_{mix}}}.
	\end{align*}
	In summary, if we have
	\begin{align*}
	\frac{n/t_{mix}}{(\log(n/t_{mix}))^2}&\geq 1024\bigg(\bar{\mu}\Vert\bm{\Sigma}^{-1}\Vert_\sigma^2 + \frac{K_{max}^2}{\bar{\mu}}\bigg)\log\frac{d_St_{mix}}{\delta},
	\end{align*}
	then with probability $1-\delta$, we have
	\begin{align*}
	&\Vert (\hat{\bm{\Sigma}}^{-1} - \bm{\Sigma}^{-1})\bm{\Sigma}^{\frac{1}{2}}\Vert_\sigma \leq 32\Vert \bm{\Sigma}^{-1}\Vert_\sigma^{\frac{3}{2}}\sqrt{\frac{\bar{\mu}\log\frac{d_St_{mix}}{\delta}(\log\frac{n}{t_{mix}})^2}{n/t_{mix}}}.
	\end{align*}

\paragraph{Proof of Lemma \ref{th:theorem-hat_F}} 
	\textbf{Step 1:}
	
	Denote $\bm{G}_i = \frac{\eta(a_i)}{\pi(a_i\vert s_i)}\phi(s_i)\circ \psi(a_i)\circ\phi(s_i')$ and let $\alpha= \lceil 2t_{mix}\log \frac{4K_{max}^{\frac{9}{2}}}{\bar{\lambda}t}\rceil + 1$. Then according to the result of Lemma \ref{mix}, we have that for arbitrary initial state distribution $\mu$, 
	\begin{align*}
	\Vert p^{\alpha, \pi}(s\vert \mu) - \xi(s)\Vert_{TV} \leq \frac{\bar{\lambda}}{2K_{max}^{3}}\wedge\frac{t}{2K_{max}^{3/2}}.
	\end{align*}
	For each $0\leq l\leq \alpha-1$ and $1\leq k\leq n_l = \lceil\frac{n - l}{\alpha}\rceil$, we define $\bm{G}_k^l = \bm{G}_{k\alpha + l}$. We also denote $\bm{G}$ as a random tensor independent with our data which is defined as
	\begin{align*}
	\bm{G} =\frac{\eta(A)}{\pi(A\vert S)}\phi(S)\circ \psi(A)\circ\phi(S'), S\sim \xi(\cdot), A\sim \pi(\cdot\vert S), S'\sim p(\cdot\vert S, A). 
	\end{align*}
	
	Then for any $u, w\in\mathbb{R}^{d_S}, v\in\mathbb{R}^{d_A}$, we have
	\begin{align*}
	\vert \bm{G}_k^l\times_1 u^\top\times_2 v^\top\times_3 w^\top\vert \leq \left\Vert \frac{\eta(a_{k\alpha + l})}{\pi(a_{k\alpha + l}\vert s_{k\alpha + l})}\phi(s_{k\alpha+l})\right\Vert\Vert \psi(a_{k\alpha+l})\Vert\Vert \phi(s_{k\alpha+l}')\Vert\leq K_{max}^{\frac{3}{2}}\kappa.
	\end{align*}
	Therefore, 
	\begin{align*}
	\vert \mathbb{E}[\bm{G}_k^l\times_1 u^\top\times_2 v^\top\times_3 w^\top\vert \mathcal{F}_{(k-1)\alpha + l}]\vert \leq K^{\frac{3}{2}}_{max}\kappa,
	\end{align*}
	which implies
	\begin{align*}
	\vert \bm{G}_k^l\times_1 u^\top\times_2 v^\top\times_3 w^\top - \mathbb{E}[\bm{G}_k^l\times_1 u^\top\times_2 v^\top\times_3 w^\top\vert \mathcal{F}_{(k-1)\alpha + l}]\vert \leq 2K^{\frac{3}{2}}_{max}\kappa.
	\end{align*}
	Meanwhile, we have
	\begin{align*}
	&\mathbb{E}[(\bm{G}_k^l\times_1 u^\top\times_2 v^\top\times_3 w^\top)^2\vert \mathcal{F}_{(k-1)\alpha + l}] \\
	=& \mathbb{E}[(\bm{G}_k^l\times_1 u^\top\times_2 v^\top\times_3 w^\top)^2 \\
	& - \mathbb{E}[(\bm{G}\times_1 u^\top\times_2 v^\top\times_3 w^\top)^2]\vert \mathcal{F}_{(k-1)\alpha + l}] + \mathbb{E}(\bm{G}\times_1 u^\top\times_2 v^\top\times_3 w^\top)^2]\\
	=& \int(\phi(s)^\top u)^2(\psi(a)^\top v)^2(\phi(s')^\top w)^2(p^{\alpha, \pi}(s\vert \mu_{(k-1)\alpha + l}) - \xi(s))\frac{\eta^2(a)}{\pi(a\vert s)}p(s'\vert s, a)dsdads' \\
	& + \mathbb{E}[(\bm{G}\times_1 u^\top\times_2 v^\top\times_3 w^\top)^2]\\
	\leq &\kappa\int(\phi(s)^\top u)^2(\psi(a)^\top v)^2(\phi(s')^\top w)^2\vert p^{\alpha, \pi}(s\vert \mu_{(k-1)\alpha + l}) - \xi(s)\vert\eta(a)p(s'\vert s, a)dsdads' + \kappa\bar{\lambda}\\
	\leq& 2\kappa\bar{\lambda},
	\end{align*}
	where $\mu_i$ is the state distribution at step $i$. Therefore, we have
	\begin{align*}
	&\mathbb{E}[(\bm{G}_k^l\times_1 u^\top\times_2 v^\top\times_3 w^\top - \mathbb{E}[\bm{G}_k^l\times_1 u^\top\times_2 v^\top\times_3 w^\top\vert \mathcal{F}_{(k-1)\alpha + l}])^2\vert \mathcal{F}_{(k-1)\alpha + l}] \\
	\leq &\mathbb{E}[(\bm{G}_k^l\times_1 u^\top\times_2 v^\top\times_3 w^\top)^2\vert \mathcal{F}_{(k-1)\alpha + l}] \leq 2\kappa\bar{\lambda}.
	\end{align*}
	
	Again we apply the matrix Bernstein's inequality \cite{tropp2011freedman} on $\bm{G}_k^l\times_1 u^\top\times_2 v^\top\times_3 w^\top$ (Note that $\bm{G}_k^l\times_1 u^\top\times_2 v^\top\times_3 w^\top$ is a scalar, which can be viewed as a $1\times 1$ matrix.), and get
	\begin{align*}
	& \mathbb{P}\left(\left\vert \frac{1}{n_l}\sum_{k=1}^{n_l}(\bm{G}_k^l\times_1 u^\top\times_2 v^\top\times_3 w^\top - \mathbb{E}[\bm{G}_k^l\times_1 u^\top\times_2 v^\top\times_3 w^\top \vert\mathcal{F}_{(k-1)\alpha + l}])\right\vert > t\right)\\
	\leq & 2e^{-\frac{1}{2}\frac{n_lt^2}{2\kappa\bar{\lambda} + \frac{2}{3}K_{max}^{3/2}\kappa t}}.
	\end{align*}
	
	\textbf{Step 2: }
	
	Now consider three $\frac{1}{4}$-nets over $S^{d_S-1}, S^{d_A - 1}, S^{d_S - 1}$, denoted as $\mathcal{N}_1, \mathcal{N}_2, \mathcal{N}_3$. By Lemma \ref{net_num}, we know that $\vert \mathcal{N}_1\vert \leq 9^{d_S}, \vert \mathcal{N}_2\vert\leq 9^{d_A}, \vert \mathcal{N}_3\vert\leq 9^{d_S}$. Then we can get a union bound by
	\begin{align*}
	& \mathbb{P}\Bigg(\exists u\in\mathcal{N}_1, v\in\mathcal{N}_2, w\in\mathcal{N}_3, \\
	& \quad \bigg\vert \frac{1}{n_l}\sum_{k=1}^{n_l}(\bm{G}_k^l\times_1 u^\top\times_2 v^\top\times_3 w^\top - \mathbb{E}[\bm{G}_k^l\times_1 u^\top\times_2 v^\top\times_3 w^\top \vert\mathcal{F}_{(k-1)\alpha + l}])\bigg\vert > t\Bigg)\\
	\leq &\sum_{u\in\mathcal{N}_1}\sum_{v\in\mathcal{N}_2}\sum_{w\in\mathcal{N}_3} \\
	& \qquad \mathbb{P}\left(\bigg\vert \frac{1}{n_l}\sum_{k=1}^{n_l}(\bm{G}_k^l\times_1 u^\top\times_2 v^\top\times_3 w^\top - \mathbb{E}[\bm{G}_k^l\times_1 u^\top\times_2 v^\top\times_3 w^\top \vert\mathcal{F}_{(k-1)\alpha + l}])\bigg\vert > t\right)\\
	=& 2\cdot 9^{2d_S + d_A}e^{-\frac{1}{2}\frac{n_lt^2}{2\kappa\bar{\lambda} + \frac{2}{3}K_{max}^{3/2}\kappa t} }\leq 2e^{-\frac{1}{2}\frac{n_lt^2}{2\kappa\bar{\lambda} + \frac{2}{3}K_{max}^{3/2}\kappa t} + 4 (d_S + d_A)\log 3}.
	\end{align*}
	Then according to Lemma \ref{net}, we know that
	\begin{align*}
	& \bigg\Vert\frac{1}{n_l}\sum_{k=1}^{n_l}(\bm{G}_k^l - \mathbb{E}[\bm{G}_k^l\vert \mathcal{F}_{(k-1)\alpha + l}])\bigg\Vert_\sigma \\
	\leq & \frac{64}{3}\max_{u\in\mathcal{N}_1, v\in\mathcal{N}_2, w\in\mathcal{N}_3}\bigg\vert\bigg\langle \frac{1}{n_l}\sum_{k=1}^{n_l}(\bm{G}_k^l - \mathbb{E}[\bm{G}_k^l\vert \mathcal{F}_{(k-1)\alpha + l}]), u\circ v\circ w\bigg\rangle\bigg\vert,
	\end{align*}
	which implies that
	\begin{align*}
	\mathbb{P}\left(\bigg\Vert \frac{1}{n_l}\sum_{k=1}^{n_l}(\bm{G}_k^l - \mathbb{E}[\bm{G}_k^l \vert\mathcal{F}_{(k-1)\alpha + l}])\bigg\Vert_\sigma > \frac{3}{64}t\right) &\leq 2e^{-\frac{1}{2}\frac{n_lt^2}{2\kappa\bar{\lambda} + \frac{2}{3}K_{max}^{3/2}\kappa t} + 4(d_S + d_A)\log 3} \\
	&\leq 2e^{-\frac{1}{2}\frac{n_lt^2}{2\kappa\bar{\lambda} + \frac{2}{3}K_{max}^{3/2}\kappa t} + 8(d_S + d_A)}.
	\end{align*}
	
	\textbf{Step 3: }
	
	Note that
	\begin{align*}
	\mathbb{E}[\bm{G}_k^l\vert \mathcal{F}_{(k-1)\alpha + l}] - \bm{F} = \int \phi(s)\circ\psi(a)\circ\phi(s') (p^{\alpha, \pi}(s\vert \mu_{(k-1)\alpha + l}) - \xi(s))\eta(a)p(s'\vert s,a)dsdads'.
	\end{align*}
	So we get
	\begin{align*}
	\Vert \mathbb{E}[\bm{G}_k^l\vert \mathcal{F}_{(k-1)\alpha + l}] - \bm{F}\Vert_\sigma\leq K_{max}^{\frac{3}{2}}\int\vert p^{\alpha, \pi}(s\vert \mu_{(k-1)\alpha + l}) - \xi(s)\vert dsdads' \leq t,
	\end{align*}
	which implies that 
	\begin{align*}
	\mathbb{P}\left(\bigg\Vert \frac{1}{n_l}\sum_{k=1}^{n_l}(\bm{G}_k^l - \bm{F})\bigg\Vert_\sigma > 2t\right) &\leq\mathbb{P}\left(\bigg\Vert \frac{1}{n_l}\sum_{k=1}^{n_l}(\bm{G}_k^l - \mathbb{E}[\bm{G}_k^l \vert\mathcal{F}_{(k-1)\alpha + l}])\bigg\Vert_\sigma > \frac{3}{64}t\right) \\
	&\leq 2e^{-\frac{1}{2}\frac{n_lt^2}{2\kappa\bar{\lambda} + \frac{2}{3}K_{max}^{3/2}\kappa t} + 8(d_S + d_A)}.
	\end{align*}
	Based on the fact that $n_l = \lceil\frac{n - l}{\alpha}\rceil \geq \frac{n}{\alpha} - 2$, and replace $2t$ by $t$, we further get
	\begin{align*}
	\mathbb{P}\left(\bigg\Vert \frac{1}{n_l}\sum_{k=1}^{n_l}(\bm{G}_k^l - \bm{F})\bigg\Vert_\sigma > t\right) \leq 2e^{-\frac{1}{16}\frac{(n - 2\alpha)t^2}{\alpha(\kappa\bar{\lambda} + \frac{1}{6}K_{max}^{3/2}\kappa t)} + 8(d_S + d_A)}.
	\end{align*}
	
	\textbf{Step 4:}
	
	Now, we get a union bound over $l$, and get
	\begin{align*}
	\mathbb{P}\left(\exists l, \bigg\Vert \frac{1}{n_l}\sum_{k=1}^{n_l}(\bm{G}_k^l - \bm{F})\bigg\Vert_\sigma > t\right) \leq &  \sum_{l=0}^{\alpha-1}\mathbb{P}\left(\bigg\Vert \frac{1}{n_l}\sum_{k=1}^{n_l}(\bm{G}_k^l - \bm{F})\bigg\Vert_\sigma > 2t\right) \\
	\leq & 2\alpha e^{-\frac{1}{16}\frac{(n - 2\alpha)t^2}{\alpha(\kappa\bar{\lambda} + \frac{1}{6}K_{max}^{3/2}\kappa t)} + 8(d_S + d_A)}.
	\end{align*}
	Such the result implies that
	\begin{align*}
	& \mathbb{P}\left(\bigg\Vert \frac{1}{n}\sum_{k=1}^{n}(\bm{G}_k - \bm{F})\bigg\Vert_\sigma > t\right) \leq \mathbb{P}\left(\exists l, \bigg\Vert \frac{1}{n_l}\sum_{k=1}^{n_l}(\bm{G}_k^l - \bm{F})\bigg\Vert_\sigma > t\right) \\
	\leq &  2\alpha e^{-\frac{1}{16}\frac{(n - 2\alpha)t^2}{\alpha(\kappa\bar{\lambda} + \frac{1}{6}K_{max}^{3/2}\kappa t)} + 8(d_S + d_A)}.
	\end{align*}
	Now we assume
	\begin{align*}
	\frac{n/t_{mix}}{(\log (n/t_{mix}))^2} \geq 1024\frac{\kappa K_{max}^3}{\bar{\lambda}}\left(\log\frac{t_{mix}}{\delta} + 8(d_S + d_A)\right)
	\end{align*}
	and take
	\begin{align*}
	t = \sqrt{\frac{1024\kappa\bar{\lambda}(\log\frac{t_{mix}}{\delta} + 8(d_S + d_A))(\log\frac{n}{t_{mix}})^2}{n/t_{mix}}}.
	\end{align*}
	Then we have
	\begin{align*}
	\alpha &\leq 4t_{mix}\log\frac{4K_{max}^{\frac{9}{2}}}{\bar{\lambda}t} = 4t_{mix}\log \left( \frac{K_{max}^{\frac{9}{2}}}{\bar{\lambda}^{3/2}}\sqrt{\frac{n/t_{mix}}{64\kappa(\log \frac{t_{mix}}{\delta} + 8(d_S + d_A))(\log\frac{n}{t_{mix}})^2}} \right) \\
	&\leq 4t_{mix}\log \left(\frac{K_{max}^{\frac{9}{2}}}{\bar{\lambda}^{3/2}}\sqrt{\frac{n/t_{mix}}{64\log\frac{n}{t_{mix}}}}\right) \leq 8t_{mix}\log\frac{n/t_{mix}}{\log\frac{n}{t_{mix}}} \leq 8t_{mix}\log \frac{n}{t_{mix}} \leq \frac{1}{4}n
	\end{align*}
	and
	\begin{align*}
	\frac{1}{6}K_{max}^{\frac{3}{2}}t = \frac{1}{6}K_{max}^{\frac{3}{2}}\sqrt{\frac{1024\bar{\lambda}\kappa(\log\frac{t_{mix}}{\delta} + d_S + d_A)(\log\frac{n}{t_{mix}})^2}{n/t_{mix}}} \leq \bar{\lambda}.
	\end{align*}
	So we have
	\begin{align*}
	\mathbb{P}(\Vert \frac{1}{n}\sum_{k=1}^{n}(\bm{G}_k - \bm{F})\Vert_\sigma > t)  &\leq  2\alpha e^{-\frac{1}{16}\frac{(n - 2\alpha)t^2}{\alpha(\kappa\bar{\lambda} + \frac{1}{6}K_{max}^{3/2}\kappa t)} + 8(d_S + d_A)} \\
	&\leq 16t_{mix}\log\frac{n}{t_{mix}} e^{-\frac{1}{1024}\frac{nt^2}{t_{mix}\kappa\bar{\lambda}\log \frac{n}{t_{mix}}} + 8(d_S + d_A)}\\
	&= 16t_{mix}\log\frac{n}{t_{mix}} e^{-(\log\frac{t_{mix}}{\delta} + 8(d_S + d_A))\log\frac{n}{t_{mix}}+ 8(d_S + d_A)}\\
	&\leq  16t_{mix}\log\frac{n}{t_{mix}} e^{-\log\frac{t_{mix}}{\delta}\log\frac{n}{t_{mix}}}\\
	&\leq  16t_{mix}\log\frac{n}{t_{mix}} e^{-\log\frac{t_{mix}}{\delta} - \log\frac{n}{t_{mix}}}\\
	&= \frac{16\delta\log\frac{n}{t_{mix}}}{n/t_{mix}} \leq \delta,
	\end{align*}
	i.e., with probability at least $1-\delta$, we have
	\begin{align*}
	\Vert \frac{1}{n}\sum_{k=1}^{n}(\bm{G}_k - \bm{F})\Vert_\sigma = \Vert\bar{\bm{F}} - \bm{F}\Vert_\sigma \leq  \sqrt{\frac{1024\kappa\bar{\lambda}(\log\frac{t_{mix}}{\delta} + 8(d_S + d_A))(\log\frac{n}{t_{mix}})^2}{n/t_{mix}}}.
	\end{align*}
	
	\textbf{Step 5:}
	According to our programming, we have
	\begin{align*}
	\Vert \hat{\bm{F}} - \bm{F}\Vert_\sigma \leq \Vert \hat{\bm{F}} - \bar{\bm{F}}\Vert_\sigma + \Vert\bar{\bm{F}} - \bm{F}\Vert_\sigma \leq 2\Vert\bar{\bm{F}} - \bm{F}\Vert _\sigma.
	\end{align*}
	So with probability at least $1-\delta$, we have
	\begin{align*}
	\Vert \hat{\bm{F}} -\bm{F}\Vert_\sigma \leq \sqrt{\frac{4096\kappa\bar{\lambda}(\log\frac{t_{mix}}{\delta} + 8(d_S + d_A))(\log\frac{n}{t_{mix}})^2}{n/t_{mix}}}.
	\end{align*}

\paragraph{Proof of Lemma \ref{im_main_thm2}}
	Given that
	\begin{align*}
	\frac{n/ t_{mix}}{(\log(n/t_{mix}))^2}&\geq 1024\left(\Vert\bm{\Sigma}^{-1}\Vert^2_\sigma \bar{\mu} + \frac{K_{max}^2}{\bar{\mu}} + \frac{\kappa K_{max}^3}{\bar{\lambda}}\right)\left(\log\frac{t_{mix}}{\delta} + 8(d_S + d_A)\right),
	\end{align*}
	then the assumptions in Lemma \ref{th:theorem-hat_F} and Lemma \ref{im_lm3} are satisfied simultaneously, and with probability at least $1-2\delta$, we have the following relations hold simultaneously, 
	\begin{align*}
	&\Vert \hat{\bm{F}} -\bm{F}\Vert_\sigma \leq 64\sqrt{\frac{\kappa\bar{\lambda}(\log\frac{t_{mix}}{\delta} + 8(d_S + d_A))(\log\frac{n}{t_{mix}})^2}{n/t_{mix}}}, \\
	&\Vert (\hat{\bm{\Sigma}}^{-1} - \bm{\Sigma}^{-1})\bm{\Sigma}^{\frac{1}{2}}\Vert_\sigma \leq 32\Vert \bm{\Sigma}^{-1}\Vert_\sigma^{\frac{3}{2}}\sqrt{\frac{\bar{\mu}\log\frac{d_St_{mix}}{\delta}(\log\frac{n}{t_{mix}})^2}{n/t_{mix}}}.
	\end{align*}
	
	So we have
	\begin{align*}
	& \Vert \hat{\bm{F}}\times_1 (\hat{\bm{\Sigma}}^{-1}\bm{\Sigma}^{\frac{1}{2}})^\top - \bm{F}\times_1(\bm{\Sigma}^{-1}\bm{\Sigma}^{\frac{1}{2}})^\top\Vert_\sigma \\ = & \Vert \hat{\bm{F}}\times_1 (\hat{\bm{\Sigma}}^{-1}\bm{\Sigma}^{\frac{1}{2}})^\top - \bm{F}\times_1 (\hat{\bm{\Sigma}}^{-1}\bm{\Sigma}^{\frac{1}{2}})^\top +  \bm{F}\times_1 (\hat{\bm{\Sigma}}^{-1}\bm{\Sigma}^{\frac{1}{2}})^\top - \bm{F}\times_1(\bm{\Sigma}^{-1}\bm{\Sigma}^{\frac{1}{2}})^\top\Vert_\sigma\\
	\leq & \Vert (\hat{\bm{F}} - \bm{F})\times_1 (\hat{\bm{\Sigma}}^{-1}\bm{\Sigma}^{\frac{1}{2}})^\top\Vert_\sigma + \Vert \bm{F}\times_1 ((\hat{\bm{\Sigma}}^{-1} - \bm{\Sigma}^{-1})^\top\bm{\Sigma}^{\frac{1}{2}})\Vert_\sigma\\
	\leq & \Vert \bm{F} - \hat{\bm{F}}\Vert_\sigma\Vert \hat{\bm{\Sigma}}^{-1}\bm{\Sigma}^{\frac{1}{2}}\Vert_\sigma + \Vert \bm{F}\Vert_\sigma\Vert (\hat{\bm{\Sigma}}^{-1} - \bm{\Sigma}^{-1})\bm{\Sigma}^{\frac{1}{2}}\Vert_\sigma\\
	\leq & \Vert \bm{F} - \hat{\bm{F}}\Vert_\sigma\Vert \bm{\Sigma}^{-\frac{1}{2}}\Vert_\sigma + \Vert \bm{F}\Vert_\sigma\Vert (\hat{\bm{\Sigma}}^{-1} - \bm{\Sigma}^{-1})\bm{\Sigma}^{\frac{1}{2}}\Vert_\sigma + \Vert \bm{F} - \hat{\bm{F}}\Vert_\sigma\Vert (\hat{\bm{\Sigma}}^{-1}-\bm{\Sigma}^{-1})\bm{\Sigma}^{\frac{1}{2}}\Vert_\sigma .
	\end{align*}
	Note that under our assumptions, we have
	\begin{align*}
	\Vert(\hat{\bm{\Sigma}}^{-1}-\bm{\Sigma}^{-1})\bm{\Sigma}^{\frac{1}{2}}\Vert_\sigma &\leq 32\Vert \bm{\Sigma}^{-1}\Vert_\sigma^{\frac{3}{2}}\sqrt{\frac{\bar{\mu}\log\frac{d_St_{mix}}{\delta}(\log\frac{n}{t_{mix}})^2}{n/t_{mix}}}\\
	&\leq 32\Vert \bm{\Sigma}^{-1}\Vert_\sigma^{\frac{3}{2}}\sqrt{\frac{\bar{\mu}\log\frac{d_St_{mix}}{\delta}}{1024\Vert\bm{\Sigma}^{-1}\Vert_\sigma^2\bar{\mu}(\log\frac{t_{mix}}{\delta}+8(d_S + d_A))}}\\
	&\leq \Vert \bm{\Sigma}^{-\frac{1}{2}}\Vert_\sigma.
	\end{align*}
	Meanwhile, we have
	\begin{align*}
	\bar{\lambda} &= \sup_{\Vert u\Vert\leq 1, \Vert v\Vert\leq 1, \Vert w\Vert\leq 1}\mathbb{E}[(\langle \phi(S)\circ \psi(A)\circ \phi(S'), u\circ v\circ w\rangle)^2] \\
	&\geq  \sup_{\Vert u\Vert\leq 1, \Vert v\Vert\leq 1, \Vert w\Vert\leq 1}(\mathbb{E}[\langle \phi(S)\circ \psi(A)\circ \phi(S'), u\circ v\circ w\rangle])^2\\
	&= \left( \sup_{\Vert u\Vert\leq 1, \Vert v\Vert\leq 1, \Vert w\Vert\leq 1}\mathbb{E}[\langle \phi(S)\circ \psi(A)\circ \phi(S'), u\circ v\circ w\rangle]\right)^2\\
	&= \Vert \bm{F}\Vert_\sigma^2.
	\end{align*}
	So we get
	\begin{align*}
	&\Vert \hat{\bm{F}}\times_1 (\hat{\bm{\Sigma}}^{-1}\bm{\Sigma}^{\frac{1}{2}})^\top - \bm{F}\times_1(\bm{\Sigma}^{-1}\bm{\Sigma}^{\frac{1}{2}})^\top\Vert_\sigma\\
	\leq &2\Vert \bm{F} - \hat{\bm{F}}\Vert_\sigma\Vert \bm{\Sigma}^{-\frac{1}{2}}\Vert_\sigma + \Vert \bm{F}\Vert_\sigma\Vert (\hat{\bm{\Sigma}}^{-1} - \bm{\Sigma}^{-1})\bm{\Sigma}^{\frac{1}{2}}\Vert_\sigma \\
	\leq &128\Vert \bm{\Sigma}^{-1}\Vert_\sigma^{\frac{1}{2}}\sqrt{\frac{\kappa\bar{\lambda}(\log\frac{t_{mix}}{\delta} + 8(d_S + d_A))(\log\frac{n}{t_{mix}})^2}{n/t_{mix}}} \\
	& + 32\Vert\bm{F}\Vert_\sigma \Vert \bm{\Sigma}^{-1}\Vert_\sigma^{\frac{3}{2}}\sqrt{\frac{\bar{\mu}\log\frac{d_St_{mix}}{\delta}(\log\frac{n}{t_{mix}})^2}{n/t_{mix}}}\\
	\leq& 256 \Vert\bm{\Sigma}^{-1}\Vert_\sigma^{\frac{1}{2}}\sqrt{\frac{\bar{\lambda}(\log\frac{t_{mix}}{\delta} + d_S + d_A)(\kappa + \bar{\mu}\Vert\bm{\Sigma}^{-1}\Vert_\sigma^2)(\log\frac{n}{t_{mix}})^2}{n/t_{mix}}}.
	\end{align*}
	Replacing $\delta$ by $\frac{1}{2}\delta$, then with probability at least $1-\delta$, we get
	\begin{align*}
	& \Vert \hat{\bm{F}}\times_1 (\hat{\bm{\Sigma}}^{-1}\bm{\Sigma}^{\frac{1}{2}})^\top - \bm{F}\times_1(\bm{\Sigma}^{-1}\bm{\Sigma}^{\frac{1}{2}})^\top\Vert_\sigma\\
	\leq & 256 \Vert\bm{\Sigma}^{-1}\Vert_\sigma^{\frac{1}{2}}\sqrt{\frac{\bar{\lambda}(\log\frac{2t_{mix}}{\delta} + d_S + d_A)(\kappa + \bar{\mu}\Vert\bm{\Sigma}^{-1}\Vert_\sigma^2)(\log\frac{n}{t_{mix}})^2}{n/t_{mix}}},
	\end{align*}
	which has finished the proof. 

\paragraph{Proof of Lemma \ref{feature_sin}}
	Note that for any vector $w\in\mathbb{R}^{d_S}$ such that $\Vert w\Vert = 1$, the columns of $\bm{U}_3$ span the row space of matrix $\bm{P}\times_1 w^\top$ and the columns of $\hat{\bm{U}}_3$ span the row space of matrix $\hat{\bm{P}}\times_1 w^\top$. So according to Wedin's lemma \cite{wedin1972perturbation}, we have
	\begin{align*}
	\Vert \sin\Theta(\bm{U}_3, \hat{\bm{U}}_3)\Vert\leq \frac{\Vert \bm{P}\times_1 w^\top - \hat{\bm{P}}\times_1 w^\top\Vert}{\sigma_m(\bm{P}\times_1 w^\top)} \leq \frac{\Vert \bm{P} - \hat{\bm{P}}\Vert_\sigma}{\sigma_m(\bm{P}\times_1 w^\top)} .
	\end{align*}
	Taking infimum over $w$, we get
	\begin{align*}
	\Vert \sin\Theta(\bm{U}_3, \hat{\bm{U}}_3)\Vert \leq  \frac{\Vert \bm{P} - \hat{\bm{P}}\Vert_\sigma}{\sup_{\Vert w\Vert = 1}\sigma_m(\bm{P}\times_1 w^\top)}  =  \frac{\Vert \bm{P} - \hat{\bm{P}}\Vert_\sigma}{\sigma},
	\end{align*}
	which has finished the proof. 

\paragraph{Proof of Theorem \ref{main_thm2}}

	According to the result of Lemma \ref{im_main_thm2}, we know that with probability at least $1-\delta$, we have
	\begin{align*}
	& \Vert \hat{\bm{F}}\times_1 (\hat{\bm{\Sigma}}^{-1}\bm{\Sigma}^{\frac{1}{2}})^\top - \bm{F}\times_1(\bm{\Sigma}^{-1}\bm{\Sigma}^{\frac{1}{2}})^\top\Vert_\sigma\\
	\leq & 256 \Vert\bm{\Sigma}^{-1}\Vert_\sigma^{\frac{1}{2}}\sqrt{\frac{\bar{\lambda}(\log\frac{2t_{mix}}{\delta} + d_S + d_A)(\kappa + \bar{\mu}\Vert\bm{\Sigma}^{-1}\Vert_\sigma^2)(\log\frac{n}{t_{mix}})^2}{n/t_{mix}}}
	\end{align*}
	and our result follows directly by noticing 
	\begin{align*}
	\Vert \hat{\bm{P}} - \bm{P}\Vert_\sigma &= \Vert \hat{\bm{F}}\times_1 \hat{\bm{\Sigma}}^{-1} - \bm{F}\times_1\bm{\Sigma}^{-1}\Vert_\sigma\\
	&=\Vert (\hat{\bm{F}}\times_1 (\hat{\bm{\Sigma}}^{-1}\bm{\Sigma}^{\frac{1}{2}})^\top - \bm{F}\times_1(\bm{\Sigma}^{-1}\bm{\Sigma}^{\frac{1}{2}})^\top)\times_1 \bm{\Sigma}^{-\frac{1}{2}}\Vert_\sigma\\
	&\leq \Vert (\hat{\bm{F}}\times_1 (\hat{\bm{\Sigma}}^{-1}\bm{\Sigma}^{\frac{1}{2}})^\top - \bm{F}\times_1(\bm{\Sigma}^{-1}\bm{\Sigma}^{\frac{1}{2}})^\top)\Vert_\sigma\Vert \bm{\Sigma}^{-1}\Vert_\sigma^{\frac{1}{2}}\\
	&\leq 256 \Vert\bm{\Sigma}^{-1}\Vert_\sigma\sqrt{\frac{\bar{\lambda}(\log\frac{2t_{mix}}{\delta} + d_S + d_A)(\kappa + \bar{\mu}\Vert\bm{\Sigma}^{-1}\Vert_\sigma^2)(\log\frac{n}{t_{mix}})^2}{n/t_{mix}}},
	\end{align*}
	which has finished the proof. 


\paragraph{Proof of Theorem \ref{dist}}

	We first prove
	\begin{align*}
	\textrm{dist}[(s, a), (s', a')] = \Vert \bm{C}\times_1 f(s)^\top\times_2 g(a)^\top - \bm{C}\times_1 f(s')^\top\times_2 g(a')^\top\Vert .
	\end{align*}
	For any $f\in \mathcal{H}_S$ such that $\Vert f\Vert_{\mathcal{H}_S}\leq 1$, we can always find some weight $\bm{w}$ such that $\Vert\bm{w}\Vert\leq 1, f = \bm{w}^\top \phi$, and
	\begin{align*}
	& \vert \langle p(\cdot\vert s, a), f(\cdot)\rangle -\langle p(\cdot\vert s', a'), f(\cdot)\rangle\vert \\ = & \vert \bm{w}^\top\langle p(\cdot\vert s, a), \phi(\cdot)\rangle -\bm{w}^\top\langle p(\cdot\vert s', a'), \phi(\cdot)\rangle\vert\\
	\leq & \Vert \langle p(\cdot\vert s, a), \phi(\cdot)\rangle - \langle p(\cdot\vert s', a'), \phi(\cdot)\rangle\Vert\\
	= & \Vert \bm{P}\times_1 \phi(s)^\top\times_2\psi(a)^\top - \bm{P}\times_1 \phi(s')^\top \times_2\psi(a')^\top\Vert\\
	= & \Vert \bm{C}\times_1 (\bm{U}_1^\top\phi(s))^\top \times_2(\bm{U}_2^\top\psi(a))^\top \times_3 \bm{U}_3 - \bm{C}\times_1 (\bm{U}_1^\top\phi(s'))^\top \times_2(\bm{U}_2^\top\psi(a'))^\top \times_3\bm{U}_3\Vert\\
	= & \Vert \bm{C}\times_1 (\bm{U}_1^\top\phi(s))^\top\times_2(\bm{U}_2^\top\psi(a))^\top - \bm{C}\times_1 (\bm{U}_1^\top\phi(s'))^\top\times_2(\bm{U}_2^\top\psi(a'))^\top\Vert\\
	= &\Vert \bm{C}\times_1 f(s)^\top\times_2 g(a)^\top - \bm{C}\times_1 f(s')^\top\times_2 g(a')^\top\Vert.
	\end{align*}
	Now we are ready to prove the main result. Notice that
	\begin{align*}
	\widehat{\dist}[(s, a), (s', a')] &=\Vert \hat{\bm{C}}\times_1 \hat{f}(s)^\top\times_2 \hat{g}(a)^\top - \hat{\bm{C}}\times_1 \hat{f}(s')^\top\times_2 \hat{g}(a')^\top\Vert\\
	&= \Vert \hat{\bm{C}}\times_1 (\hat{\bm{U}}_1^\top\phi(s))^\top \times_2 (\hat{\bm{U}}_2^\top\psi(a))^\top - \hat{\bm{C}}\times_1 (\hat{\bm{U}}_1^\top\phi(s'))^\top\times_2 (\hat{\bm{U}}_2^\top\psi(a'))^\top\Vert\\
	&= \Vert \hat{\bm{P}}\times_1 \phi(s)^\top\times_2 \psi(a)^\top\times_3\hat{\bm{U}}_3^\top - \hat{\bm{P}}\times_1 \phi(s')^\top\times_2 \psi(a')^\top\times_3\hat{\bm{U}}_3^\top\Vert\\
	&= \Vert \hat{\bm{\Phi}}(s, a) - \hat{\bm{\Phi}}(s', a')\Vert
	\end{align*}
	and for any orthogonal matrix $\bm{O}\in\mathbb{R}^{m\times m}$, we have
	\begin{align*}
	\dist[(s, a), (s', a')] &=\Vert \bm{C}\times_1 f(s)^\top\times_2 g(a)^\top - \bm{C}\times_1 f(s')^\top\times_2 g(a')^\top\Vert\\
	&= \Vert \bm{C}\times_1 (\bm{U}_1^\top\phi(s))^\top \times_2 (\bm{U}_2^\top\psi(a))^\top - \bm{C}\times_1 (\bm{U}_1^\top\phi(s'))^\top\times_2 (\bm{U}_2^\top\psi(a'))^\top\Vert\\
	&= \Vert \bm{P}\times_1 \phi(s)^\top\times_2 \psi(a)^\top\times_3\bm{U}_3^\top - \bm{P}\times_1 \phi(s')^\top\times_2 \psi(a')^\top\times_3\bm{U}_3^\top\Vert\\
	&= \Vert \bm{P}\times_1 \phi(s)^\top\times_2 \psi(a)^\top\times_3(\bm{U}_3 \bm{O})^\top - \bm{P}\times_1 \phi(s')^\top\times_2 \psi(a')^\top \times_3(\bm{U}_3 \bm{O})^\top\Vert\\
	&= \Vert \bm{O}^\top\bm{\Phi}(s, a) - \bm{O}^\top\bm{\Phi}(s', a')\Vert.
	\end{align*}
	So we have
	\begin{align*}
	\vert\widehat{\dist}[(s, a), (s', a')] - \dist[(s, a), (s', a')] \vert \leq \Vert \hat{\bm{\Phi}}(s, a) - \bm{O}^\top\bm{\Phi}(s, a)\Vert + \Vert \hat{\bm{\Phi}}(s', a') - \bm{O}^\top\bm{\Phi}(s', a')\Vert.
	\end{align*}
	It suffices to bound $\Vert \hat{\bm{\Phi}}(s, a) - \bm{O}^\top\bm{\Phi}(s, a)\Vert $, and $\Vert \hat{\bm{\Phi}}(s', a') - \bm{O}^\top\bm{\Phi}(s', a')\Vert$ can be bounded in the exactly same way. Notice that
	\begin{align*}
	& \Vert \hat{\bm{\Phi}}(s, a) - \bm{O}^\top\bm{\Phi}(s, a)\Vert \\
	\leq & \Vert  (\hat{\bm{P}} - \bm{P})\times_1 \phi(s)^\top\times_2 \psi(a)^\top\times_3\hat{\bm{U}}_3^\top + \bm{P}\times_1 \phi(s)^\top\times_2 \psi(a)^\top\times_3(\hat{\bm{U}}_3 - \bm{U}_3 \bm{O})^\top\Vert \\
	\leq & \Vert \bm{P} - \hat{\bm{P}}\Vert_\sigma K_{max} + \Vert\bm{P}\Vert_\sigma K_{max}\Vert \bm{U}_3^\top - \hat{\bm{U}}_3^\top \bm{O}\Vert.
	\end{align*}
	In particular, By Part 3, Lemma 1 in \cite{cai2018rate}, one can choose $\bm{O}$ such that
	\begin{align*}
	\Vert \bm{U}_3^\top - \hat{\bm{U}}_3^\top \bm{O}\Vert  \leq \sqrt{2}\Vert \sin\Theta(\bm{U}_3, \hat{\bm{U}}_3)\Vert.
	\end{align*}
	Then we have
	\begin{align*}
	\Vert \hat{\bm{\Phi}}(s, a) - \bm{O}^\top\bm{\Phi}(s, a)\Vert&\leq  \Vert \bm{P} - \hat{\bm{P}}\Vert_\sigma K_{max} + \sqrt{2}\Vert\bm{P}\Vert_\sigma K_{max}\Vert \sin\Theta(\bm{U}_3, \hat{\bm{U}}_3)\Vert.
	\end{align*}
	Now according to the result of Lemma \ref{feature_sin}, we know that
	\begin{align*}
	\Vert \sin\Theta(\bm{U}_3, \hat{\bm{U}}_3)\Vert\leq  \frac{\Vert \bm{P} - \hat{\bm{P}}\Vert_\sigma}{\sigma}.
	\end{align*}
	So we get
	\begin{align*}
	\Vert \hat{\bm{\Phi}}(s, a) - \bm{O}^\top\bm{\Phi}(s, a)\Vert \leq K_{max}\left(1 + \sqrt{2}\frac{\Vert\bm{P}\Vert_\sigma}{\sigma}\right)\Vert\hat{\bm{P}}-\bm{P}\Vert_\sigma.
	\end{align*}
	It follows that 
	\begin{align*}
	\vert\widehat{\dist}[(s, a), (s', a')] - \dist[(s, a), (s', a')] \vert&\leq\Vert \hat{\bm{\Phi}}(s, a) - \bm{O}^\top\bm{\Phi}(s, a)\Vert + \Vert \hat{\bm{\Phi}}(s', a') - \bm{O}^\top\bm{\Phi}(s', a')\Vert\\
	&\leq 2K_{max}\left(1 + \sqrt{2}\frac{\Vert\bm{P}\Vert_\sigma}{\sigma}\right)\Vert\hat{\bm{P}}-\bm{P}\Vert_\sigma,
	\end{align*}
	which has finished the proof. 

\paragraph{Proof of Theorem \ref{discrete}}
	Define $\hat{p}_d(\cdot\vert s, a) = \sum_{i=1}^{n_s}\sum_{j=1}^{n_a}\hat{q}_{ij}(\cdot)\bm{1}_{s\in \hat{A}_i}\bm{1}_{a\in \hat{B}_j}$, then we have
	\begin{align*}
	&\sum_{i, j}\int_{\hat{A}_i\times\hat{B}_j}\xi(s)\eta(a)\Vert p(\cdot\vert s, a) - \hat{q}_{ij}(\cdot)\Vert_{\mathcal{H}_S}^2dsda = \int \xi(s)\eta(a)\Vert p(\cdot\vert s, a) - \hat{p}_d(\cdot\vert s, a)\Vert_{\mathcal{H}_S}^2dsda\\
	=& \int \xi(s)\eta(a)\Vert\langle p(\cdot\vert s, a), \phi(\cdot)\rangle - \langle \hat{p}_d(\cdot\vert s, a), \phi(\cdot)\rangle\Vert^2dsda.
	\end{align*}
	Note that
	\begin{align*}
	&\int \xi(s)\eta(a)\Vert\langle p(\cdot\vert s, a), \phi(\cdot)\rangle - \langle \hat{p}_d(\cdot\vert s, a), \phi(\cdot)\rangle\Vert^2dsda\\
	=& \int \xi(s)\eta(a)\Vert\langle p(\cdot\vert s, a), \phi(\cdot)\rangle - \langle \hat{p}(\cdot\vert s, a), \phi(\cdot)\rangle + \langle \hat{p}(\cdot\vert s, a), \phi(\cdot)\rangle - \langle \hat{p}_d(\cdot\vert s, a), \phi(\cdot)\rangle\Vert^2dsda\\
	\leq &2\int \xi(s)\eta(a)\Vert\langle p(\cdot\vert s, a), \phi(\cdot)\rangle - \langle \hat{p}(\cdot\vert s, a), \phi(\cdot)\rangle\Vert^2dsda \\
	& + 2\int \xi(s)\eta(a)\Vert\langle \hat{p}(\cdot\vert s, a), \phi(\cdot)\rangle - \langle \hat{p}_d(\cdot\vert s, a), \phi(\cdot)\rangle\Vert^2dsda\\
	=& 2\int \xi(s)\eta(a)\Vert (\bm{P} - \hat{\bm{P}})\times_1 \phi(s)^\top\times_2\psi(a)^\top \Vert^2dsda \\
	& + 2\int \xi(s)\eta(a)\Vert\langle \hat{p}(\cdot\vert s, a), \phi(\cdot)\rangle - \langle \hat{p}_d(\cdot\vert s, a), \phi(\cdot)\rangle\Vert^2dsda.
	\end{align*}
	We have
	\begin{align*}
	&\int \xi(s)\eta(a)\Vert\langle \hat{p}(\cdot\vert s, a), \phi(\cdot)\rangle - \langle \hat{p}_d(\cdot\vert s, a), \phi(\cdot)\rangle\Vert^2dsda \\
	=& \int \xi(s)\eta(a)\Bigg\Vert\hat{\bm{C}}\times_1 \hat{f}(s)^\top\times_2 \hat{g}(a)^\top\times_3 \hat{\bm{U}}_3 - \sum_{i=1}^{n_s}\sum_{j=1}^{n_a}\bm{1}_{s\in \hat{A}_i}\bm{1}_{a\in \hat{B}_j}\hat{\bm{U}}_3\hat{\bm{z}}_{ij}\Bigg\Vert^2dsda\\
	=&\int \xi(s)\eta(a)\Bigg\Vert\hat{\bm{C}}\times_1 \hat{f}(s)^\top \times_2 \hat{g}(a)^\top - \sum_{i=1}^{n_s}\sum_{j=1}^{n_a}\bm{1}_{s\in \hat{A}_i}\bm{1}_{a\in \hat{B}_j}\hat{\bm{z}}_{ij}\Bigg\Vert^2dsda.
	\end{align*}
	For any orthogonal matrix $\bm{O}\in\mathbb{R}^{m\times m}$, because $\hat{A}_i, \hat{B}_j, \hat{\bm{z}}_{ij}$ is the minimizer of the above term, we have
	\begin{align*}
	&\int \xi(s)\eta(a)\Vert\hat{\bm{C}}\times_1 \hat{f}(s)^\top \times_2 \hat{g}(a)^\top - \sum_{i=1}^{n_s}\sum_{j=1}^{n_a}\bm{1}_{s\in \hat{A}_i}\bm{1}_{a\in \hat{B}_j}\hat{\bm{z}}_{ij}\Vert^2dsda\\
	\leq &\int \xi(s)\eta(a)\Vert\hat{\bm{C}}\times_1 \hat{f}(s)^\top \times_2 \hat{g}(a)^\top - \sum_{i=1}^{n_s}\sum_{j=1}^{n_a}\bm{1}_{s\in A_i}\bm{1}_{a\in B_j}\bm{O}\bm{z}_{ij}\Vert^2dsda\\
	\leq &2\int \xi(s)\eta(a)\Vert\hat{\bm{C}}\times_1 \hat{f}(s)^\top \times_2 \hat{g}(a)^\top - \bm{C}\times_1 f(s)^\top \times_2 g(a)^\top\times_3 \bm{O}\Vert^2dsda \\
	&+ 2\int \xi(s)\eta(a)\Vert\bm{C}\times_1 f(s)^\top\times_2 g(a)^\top\times_3 \bm{O} - \sum_{i=1}^{n_s}\sum_{j=1}^{n_a}\bm{1}_{s\in A_i}\bm{1}_{a\in B_j}\bm{O}\bm{z}_{ij}\Vert^2dsda\\
	=& 2\int \xi(s)\eta(a)\Vert\hat{\bm{C}}\times_1 \hat{f}(s)^\top\times_2 \hat{g}(a)^\top - \bm{C}\times_1 f(s)^\top\times_2 g(a)^\top\times_3 \bm{O}\Vert^2dsda \\
	&+ 2\int \xi(s)\eta(a)\Vert\bm{C}\times_1 f(s)^\top\times_2 g(a)^\top - \sum_{i=1}^{n_s}\sum_{j=1}^{n_a}\bm{1}_{s\in A_i}\bm{1}_{a\in B_j}\bm{z}_{ij}\Vert^2dsda\\
	=& 2\int \xi(s)\eta(a)\Vert\hat{\bm{C}}\times_1 \hat{f}(s)^\top\times_2 \hat{g}(a)^\top - \bm{C}\times_1 f(s)^\top\times_2 g(a)^\top\times_3 \bm{O}\Vert^2dsda \\
	&+ 2\int \xi(s)\eta(a)\Vert \langle p(\cdot\vert s, a), \phi(\cdot)\rangle - \langle p_d(\cdot\vert s, a), \phi(\cdot)\rangle\Vert^2dsda\\
	=& 2\int \xi(s)\eta(a)\Vert\hat{\bm{C}}\times_1 \hat{f}(s)^\top \times_2 \hat{g}(a)^\top - \bm{C}\times_1 f(s)^\top\times_2 g(a)^\top\times_3 \bm{O}\Vert^2dsda + 2L^*
	\end{align*}
	where we denote 
	\begin{align*}
	\bm{z}_{ij} = \bm{U}_3^\top\langle q^*_{ij}(\cdot), \phi(\cdot)\rangle, \qquad p_d(\cdot\vert s, a) = \sum_{i=1}^{n_s}\sum_{j=1}^{n_a}q^*_{ij}(\cdot)\bm{1}_{s\in A_i}\bm{1}_{a\in B_j}.
	\end{align*}
	Furthermore, we have
	\begin{align*}
	&\int \xi(s)\eta(a)\Vert\hat{\bm{C}}\times_1 \hat{f}(s)^\top\times_2 \hat{g}(a)^\top - \bm{C}\times_1 f(s)^\top\times_2 g(a)^\top\times_3 \bm{O}\Vert^2dsda \\
	=& \int \xi(s)\eta(a)\Vert\hat{\bm{P}}\times_1 \phi(s)^\top\times_2 \psi(a)^\top\times_3 \hat{\bm{U}}_3^\top - \bm{P}\times_1 \phi(s)^\top\times_2 \psi(a)^\top\times_3 (\bm{U}_3\bm{O}^\top)^\top\Vert^2dsda\\
	\leq & 2\int \xi(s)\eta(a)\Vert(\hat{\bm{P}} - \bm{P})\times_1 \phi(s)^\top\times_2 \psi(a)^\top\times_3 \hat{\bm{U}}_3^\top\Vert^2dsda \\
	&+ 2\int \xi(s)\eta(a)\Vert \bm{P}\times_1 \phi(s)^\top\times_2 \psi(a)^\top\times_3 (\hat{\bm{U}}_3 - \bm{U}_3\bm{O}^\top)^\top\Vert^2dsda\\
	\leq & 2\int \xi(s)\eta(a)\Vert(\hat{\bm{P}} - \bm{P})\times_1 \phi(s)^\top\times_2 \psi(a)^\top\Vert^2dsda \\
	& + 2\int \xi(s)\eta(a)\Vert \bm{P}\times_1 \phi(s)^\top\times_2 \psi(a)^\top\times_3 (\hat{\bm{U}}_3 - \bm{U}_3\bm{O}^\top)^\top\Vert^2dsda.
	\end{align*}
	Therefore, 
	\begin{align*}
	&\int \xi(s)\eta(a)\Vert\langle p(\cdot\vert s, a), \phi(\cdot)\rangle - \langle \hat{p}_d(\cdot\vert s, a), \phi(\cdot)\rangle\Vert^2dsda\\
	\leq & 4L^* + 10\int \xi(s)\eta(a)\Vert(\hat{\bm{P}} - \bm{P})\times_1 \phi(s)^\top\times_2 \psi(a)^\top\Vert^2dsda \\
	&+ 8\int \xi(s)\eta(a)\Vert \bm{P}\times_1 \phi(s)^\top\times_2 \psi(a)^\top\times_3 (\hat{\bm{U}}_3 - \bm{U}_3\bm{O}^\top)^\top\Vert^2dsda\\
	=& 10\sum_{k=1}^{d_S} \!\int\! \xi(s)\eta(a)\big( (\hat{\bm{P}} - \bm{P})_{::k}\times_1 \phi(s)^\top\times_2\psi(a)^\top \big)^2dsda \\
	&+8\int \xi(s)\eta(a)\Vert \bm{P}\times_1 \phi(s)^\top\times_2 \psi(a)^\top\times_3 (\hat{\bm{U}}_3 \!-\! \bm{U}_3\bm{O}^\top)^\top\Vert^2dsda+ 4L^*\\
	=& 10\sum_{k=1}^{d_S}\int \xi(s)\eta(a) \psi(a)^\top(\hat{\bm{P}} - \bm{P})_{::k}^\top\phi(s)\phi(s)^\top(\hat{\bm{P}} - \bm{P})_{::k}\psi(a) dsda \\ & \quad  +8\int \xi(s)\eta(a)\Vert \bm{P}\times_1 \phi(s)^\top\times_2 \psi(a)^\top\times_3 (\hat{\bm{U}}_3 - \bm{U}_3\bm{O}^\top)^\top\Vert^2dsda
	+ 4L^*.
	\end{align*}
	Note that
	\begin{align*}
	\int \xi(s)\phi(s)\phi(s)^\top ds = \bm{\Sigma}, \qquad \int \eta(a)\psi(a)\psi(a)^\top da = I_{d_A}
	\end{align*}
	So we have
	\begin{align*}
	&\int \xi(s)\eta(a) \psi(a)^\top(\hat{\bm{P}} - \bm{P})_{::k}^\top\phi(s)\phi(s)^\top(\hat{\bm{P}} - \bm{P})_{::k}\psi(a)dsda \\
	=& \int \eta(a) \psi(a)^\top(\hat{\bm{P}} - \bm{P})_{::k}^\top\bigg(\int \xi(s)\phi(s)\phi(s)^\top ds\bigg)(\hat{\bm{P}} - \bm{P})_{::k}\psi(a)da\\
	=&\int \eta(a) \psi(a)^\top(\hat{\bm{P}} - \bm{P})_{::k}^\top\bm{\Sigma}(\hat{\bm{P}} - \bm{P})_{::k}\psi(a)da\\
	=& \int tr\big[\eta(a) \psi(a)^\top(\hat{\bm{P}} - \bm{P})_{::k}^\top\bm{\Sigma}(\hat{\bm{P}} - \bm{P})_{::k}\psi(a)\big]da\\
	=& tr\bigg[(\hat{\bm{P}} - \bm{P})_{::k}^\top\bm{\Sigma}(\hat{\bm{P}} - \bm{P})_{::k}\int\eta(a)\psi(a)\psi(a)^\top da\bigg]\\
	=&tr\big[(\hat{\bm{P}} - \bm{P})_{::k}^\top\bm{\Sigma}(\hat{\bm{P}} - \bm{P})_{::k}\big]\\
	=& \Vert \bm{\Sigma}^{\frac{1}{2}}(\bm{\hat{P}} - \bm{P})_{::k}\Vert_F^2,
	\end{align*}
	which implies
	\begin{align*}
	& \sum_{k=1}^{d_S}\int \xi(s)\eta(a) \psi(a)^\top(\hat{\bm{P}} - \bm{P})_{::k}^\top\phi(s)\phi(s)^\top(\hat{\bm{P}} - \bm{P})_{::k}\psi(a) dsda \\
	\leq &  \sum_{k=1}^{d_S}\Vert \bm{\Sigma}^{\frac{1}{2}}(\hat{\bm{P}} - \bm{P})_{::k}\Vert_F^2 = \Vert (\bm{P} - \hat{\bm{P}})\times_1 \bm{\Sigma}^{\frac{1}{2}}\Vert^2_F.
	\end{align*}
	With a similar argument, we also have
	\begin{align*}
	\int \xi(s)\eta(a)\Vert \bm{P}\times_1 \phi(s)^\top\times_2 \psi(a)^\top\times_3 (\hat{\bm{U}}_3 - \bm{U}_3\bm{O}^\top)^\top\Vert^2dsda = \Vert \bm{P}\times_1\bm{\Sigma}^{\frac{1}{2}}\times_3 (\hat{\bm{U}}_3 - \bm{U}_3\bm{O}^\top)^\top\Vert_F^2.
	\end{align*}
	Therefore, 
	\begin{align*}
	& \int \xi(s)\eta(a)\Vert\langle p(\cdot\vert s, a), \phi(\cdot)\rangle - \langle \hat{p}_d(\cdot\vert s, a), \phi(\cdot)\rangle\Vert^2dsda \\ \leq & 10 \Vert (\hat{\bm{P}} - \bm{P})\times_1\bm{\Sigma}^{\frac{1}{2}}\Vert_F^2  +8\Vert \bm{P}\times_1\bm{\Sigma}^{\frac{1}{2}}\times_3 (\hat{\bm{U}}_3 - \bm{U}_3\bm{O}^\top)^\top\Vert_F^2+ 4L^*.
	\end{align*}
	Note that 
	\begin{align*}
	(\hat{\bm{P}} - \bm{P})\times_1\bm{\Sigma}^{\frac{1}{2}} = \hat{\bm{F}}\times_1(\hat{\bm{\Sigma}}^{-1}\bm{\Sigma}^{\frac{1}{2}})^\top - \bm{F}\times_1(\bm{\Sigma}^{-1}\bm{\Sigma}^{\frac{1}{2}})^\top.
	\end{align*}
	According to the result of Lemma \ref{norm}, and note that $\textrm{Tucker-Rank}(\hat{\bm{F}}\times_1 (\hat{\bm{\Sigma}}^{-1}\bm{\Sigma}^{\frac{1}{2}})^\top - \bm{F}\times_1(\bm{\Sigma}^{-1}\bm{\Sigma}^{\frac{1}{2}})^\top) \leq (2r, 2l, 2m)$, we have
	\begin{align*}
	& \Vert \hat{\bm{F}}\times_1 (\hat{\bm{\Sigma}}^{-1}\bm{\Sigma}^{\frac{1}{2}})^\top - \bm{F}\times_1(\bm{\Sigma}^{-1}\bm{\Sigma}^{\frac{1}{2}})^\top\Vert_F \\
	\leq & 2\sqrt{\frac{rlm}{\max\{r, l, m\}}}\Vert \hat{\bm{F}}\times_1 (\hat{\bm{\Sigma}}^{-1}\bm{\Sigma}^{\frac{1}{2}})^\top - \bm{F}\times_1(\bm{\Sigma}^{-1}\bm{\Sigma}^{\frac{1}{2}})^\top\Vert_\sigma.
	\end{align*}
	For the second term, similarly we have
	\begin{align*}
	\Vert \bm{P}\times_1\bm{\Sigma}^{\frac{1}{2}}\times_3 (\hat{\bm{U}}_3 - \bm{U}_3\bm{O}^\top)^\top\Vert_F \leq 2\sqrt{\frac{rlm}{\max\{r, l, m\}}}\Vert \bm{P}\times_1\bm{\Sigma}^{\frac{1}{2}}\times_3 (\hat{\bm{U}}_3 - \bm{U}_3\bm{O}^\top)^\top\Vert_\sigma.
	\end{align*}
	Note that
	\begin{align*}
	\Vert \bm{P}\times_1\bm{\Sigma}^{\frac{1}{2}}\times_3 (\hat{\bm{U}}_3 - \bm{U}_3\bm{O}^\top)^\top\Vert_\sigma \leq \Vert \bm{F}\Vert_\sigma\Vert \bm{\Sigma}^{-\frac{1}{2}}\Vert_\sigma\Vert \hat{\bm{U}}_3 - \bm{U}_3\bm{O}^\top\Vert_\sigma \leq  \sqrt{\bar{\lambda}}\Vert\bm{\Sigma}^{-\frac{1}{2}}\Vert_\sigma\Vert \hat{\bm{U}}_3 - \bm{U}_3\bm{O}^\top\Vert_\sigma .
	\end{align*}
	In particular, by Part 3, Lemma 1 in \cite{cai2018rate}, we can choose $\bm{O}$ such that $\Vert \hat{\bm{U}}_3 - \bm{U}_3\bm{O}^\top\Vert_\sigma \leq \sqrt{2}\Vert\sin\Theta(\bm{U}_3, \hat{\bm{U}}_3)\Vert_\sigma$
	and according to the result of Lemma \ref{feature_sin}, we know that
	\begin{align*}
	\Vert\sin\Theta(\bm{U}_3, \hat{\bm{U}}_3)\Vert_\sigma \leq \frac{\Vert \bm{P} - \hat{\bm{P}}\Vert_\sigma}{\sigma}.
	\end{align*}
	By Theorem \ref{main_thm2} and Lemma \ref{im_main_thm2}, we know that with probability at least $1-\delta$, we have
	\begin{align*}
	& \Vert \hat{\bm{F}}\times_1 (\hat{\bm{\Sigma}}^{-1}\bm{\Sigma}^{\frac{1}{2}})^\top - \bm{F}\times_1(\bm{\Sigma}^{-1}\bm{\Sigma}^{\frac{1}{2}})^\top\Vert_\sigma\\
	\leq & 256 \Vert\bm{\Sigma}^{-1}\Vert_\sigma^{\frac{1}{2}}\sqrt{\frac{\bar{\lambda}(\log\frac{2t_{mix}}{\delta} + d_S + d_A)(\kappa + \bar{\mu}\Vert\bm{\Sigma}^{-1}\Vert_\sigma^2)(\log\frac{n}{t_{mix}})^2}{n/t_{mix}}}\\
	& \Vert \hat{\bm{P}} - \bm{P}\Vert_\sigma \\
	\leq & 256 \Vert\bm{\Sigma}^{-1}\Vert_\sigma\sqrt{\frac{\bar{\lambda}(\log\frac{2t_{mix}}{\delta} + d_S + d_A)(\kappa + \bar{\mu}\Vert\bm{\Sigma}^{-1}\Vert_\sigma^2)(\log\frac{n}{t_{mix}})^2}{n/t_{mix}}},
	\end{align*}
	which implies that with probability at least $1-\delta$, 
	\begin{align*}
	&\int \xi(s)\eta(a)\Vert\langle p(\cdot\vert s, a), \phi(\cdot)\rangle - \langle \hat{p}_d(\cdot\vert s, a), \phi(\cdot)\rangle\Vert^2dsda \\
	\leq &2^{22}\Vert\bm{\Sigma}^{-1}\Vert_\sigma\frac{rlm}{\max\{r, l, m\}}\left(1 + \frac{2\bar{\lambda}\Vert \bm{\Sigma}^{-1}\Vert_\sigma^2}{\sigma^2}\right)\frac{\bar{\lambda}(\log\frac{2t_{mix}}{\delta} + d_S + d_A)(\kappa + \bar{\mu}\Vert\bm{\Sigma}^{-1}\Vert_\sigma^2)(\log\frac{n}{t_{mix}})^2}{n/t_{mix}} + 4L^*.
	\end{align*}
	Then we get the desired result.

\paragraph{Proof of Theorem \ref{partition}}
	For each $k\in[n_s], l\in[n_a]$, define
	\begin{align*}
	\delta_{kl}^2 = \min_{(i, j)\neq (k, l)}\Vert q_{kl}(\cdot) - q_{ij}(\cdot)\Vert_{\mathcal{H}_S}^2 =\min_{(i, j)\neq (k, l)}\Vert \bm{z}_{kl} - \bm{z}_{ij}\Vert^2, \qquad \bar{\Psi}(s, a) = \sum_{i=1}^{n_s}\sum_{j=1}^{n_a}\hat{\bm{z}}_{ij}\bm{1}_{s\in \hat{A}_i}\bm{1}_{a\in \hat{B}_j}.
	\end{align*}
	where we still use the notation $\bm{z}_{ij} = \bm{U}_3^\top\langle q_{ij}(\cdot), \phi(\cdot)\rangle$. For an arbitrary orthogonal matrix $\bm{O}\in\mathbb{R}^{m\times m}$, let
	\begin{align*}
	S_{kl} = \{(s, a)\in A_{k}\times B_l\vert \Vert \bm{O}\bar{\Psi}(s, a) - \bm{z}_{kl}\Vert \geq \frac{\delta_{kl}}{2}\},
	\end{align*}
	then we have
	\begin{align*}
	\sum_{k=1}^{n_s}\sum_{l=1}^{n_a}(\xi\times\eta)(S_{kl})\delta_{kl}^2 &\leq 4\sum_{k=1}^{n_s}\sum_{l=1}^{n_a}\int_{S_{kl}}\xi(s)\eta(a)\Vert \bm{O}\bar{\Psi}(s, a) - \bm{z}_{kl}\Vert^2 dsda\\
	& \leq 4\int\xi(s)\eta(a)\Bigg\Vert \bm{O}\bar{\Psi}(s, a) - \sum_{k=1}^{n_s}\sum_{l=1}^{n_a}\bm{z}_{kl}\bm{1}_{s\in A_k}\bm{1}_{a\in B_l}\Bigg\Vert^2dsda \\
	&= 4\Bigg\Vert  \bm{O}\bar{\Psi}(\cdot) - \sum_{k=1}^{n_s}\sum_{l=1}^{n_a}\bm{z}_{kl}\bm{1}_{A_k\times B_l}(\cdot)\Bigg\Vert_{L^2(\xi\times \eta)}^2.
	\end{align*}
	Note that
	\begin{align*}
	&\Bigg\Vert  \bm{O}\bar{\Psi}(\cdot) - \sum_{k=1}^{n_s}\sum_{l=1}^{n_a}\bm{z}_{kl}\bm{1}_{A_k\times B_l}(\cdot)\Bigg\Vert_{L^2(\xi\times \eta)}^2 \\
	\leq& \Vert  \bm{O}\bar{\Psi}(\cdot) - \hat{\bm{C}}\times_1 \hat{f}^\top(\cdot)\times_2 \hat{g}^\top(\cdot)\times_3 \bm{O}\Vert_{L^2(\xi\times\eta) }\\
	& + \Bigg\Vert\hat{\bm{C}}\times_1 \hat{f}(\cdot)^\top\times_2 \hat{g}(\cdot)^\top\times_3 \bm{O} - \sum_{k=1}^{n_s}\sum_{l=1}^{n_a}\bm{z}_{kl}\bm{1}_{A_k\times B_l}(\cdot)\Bigg\Vert_{L^2(\xi\times\eta)}\\
	=& \Vert  \bar{\Psi}(\cdot) - \hat{\bm{C}}\times_1 \hat{f}(\cdot)^\top \times_2 \hat{g}(\cdot)^\top \Vert_{L^2(\xi\times\eta) }\\
	& + \Bigg\Vert\hat{\bm{C}}\times_1 \hat{f}(\cdot)^\top \times_2 \hat{g}(\cdot)^\top \times_3 \bm{O} - \sum_{k=1}^{n_s}\sum_{l=1}^{n_a}\bm{z}_{kl}\bm{1}_{A_k\times B_l}(\cdot)\Bigg\Vert_{L^2(\xi\times\eta)}.
	\end{align*}
	We use the fact that $\hat{A}_i, \hat{B}_j, \hat{\bm{z}}_{ij}$ is the solution to (\ref{prob}), and derive
	\begin{align*}
	& \Vert \bar{\Psi}(\cdot) - \hat{\bm{C}}\times_1 \hat{f}(\cdot)^\top \times_2 \hat{g}(\cdot)^\top \Vert_{L^2(\xi\times\eta)}\\ 
	=& \Bigg\Vert \sum_{i=1}^{n_s}\sum_{j=1}^{n_a}\hat{\bm{z}}_{ij}\bm{1}_{\hat{A}_i\times \hat{B}_j}(\cdot) - \hat{\bm{C}}\times_1 \hat{f}(\cdot)^\top\times_2 \hat{g}(\cdot)^\top\Bigg\Vert_{L^2(\xi\times\eta)} \\
	\leq&\Bigg\Vert \sum_{i=1}^{n_s}\sum_{j=1}^{n_a}\bm{O}^{\top}\bm{z}_{ij}\bm{1}_{A_i\times B_j}(\cdot) - \hat{\bm{C}}\times_1 \hat{f}(\cdot)^\top\times_2 \hat{g}(\cdot)^\top\Bigg\Vert_{L^2(\xi\times\eta)}\\
	= & \Vert\bm{C}\times_1 f(\cdot)^\top\times_2 g(\cdot)^\top\times_3 \bm{O}^\top - \hat{\bm{C}}\times_1 \hat{f}(\cdot)^\top\times_2 \hat{g}(\cdot)^\top \Vert_{L^2(\xi\times\eta)} \\
	= & \Vert\bm{P}\times_1 \phi(\cdot)^\top \times_2 \psi(\cdot)^\top \times_3 (\bm{U}_3\bm{O})^\top - \hat{\bm{P}}\times_1 \phi(\cdot)^\top\times_2 \psi(\cdot)^\top\times_3\hat{\bm{U}}_3^\top\Vert_{L^2(\xi\times\eta)}.
	\end{align*}
	Here we use the fact that
	\begin{align*}
	\bm{C}\times_1 f(s)^\top\times_2 g(a)^\top = \bm{U}_3^\top \langle p(\cdot\vert s, a), \phi(\cdot)\rangle = \sum_{i=1}^{n_s}\sum_{j=1}^{n_a}\bm{z}_{ij}\bm{1}_{A_i\times B_j}(s, a)
	\end{align*}
	because of Assumption \ref{block}. Therefore, we have
	\begin{align*}
	& \Bigg\Vert \bm{O}\bar{\Psi}(\cdot) - \sum_{k=1}^{n_s}\sum_{l=1}^{n_a}\bm{z}_{kl}\bm{1}_{A_k\times B_l}(\cdot)\Bigg\Vert_{L^2(\xi\times\eta)}\\
	\leq & 2\Vert\bm{P}\times_1 \phi(\cdot)^\top\times_2 \psi(\cdot)^\top\times_3 (\bm{U}_3\bm{O})^\top - \hat{\bm{P}}\times_1 \phi(\cdot)^\top\times_2 \psi(\cdot)^\top\times_3\hat{\bm{U}}_3^\top\Vert_{L^2(\xi\times\eta)} 
	\end{align*}
	and
	\begin{align*}
	& \sum_{k=1}^{n_s}\sum_{l=1}^{n_a}(\xi\times\eta)(S_{kl})\delta_{kl}^2\\ \leq & 16\Vert\bm{P}\times_1 \phi(\cdot)^\top\times_2 \psi(\cdot)^\top\times_3 (\bm{U}_3\bm{O})^\top - \hat{\bm{P}}\times_1 \phi(\cdot)^\top\times_2 \psi(\cdot)^\top\times_3\hat{\bm{U}}_3^\top\Vert_{L^2(\xi\times\eta)} ^2\\
	\leq & 32\Vert(\bm{P} - \hat{\bm{P}})\times_1 \phi(\cdot)^\top\times_2 \psi(\cdot)^\top\times_3 \hat{\bm{U}}_3^\top\Vert_{L^2(\xi\times\eta)} ^2 \\
	& + 32\Vert\bm{P}\times_1 \phi(\cdot)^\top\times_2 \psi(\cdot)^\top\times_3 (\bm{U}_3\bm{O} - \hat{\bm{U}}_3)^\top\Vert_{L^2(\xi\times\eta)} ^2\\
	\leq & 32\Vert(\bm{P} - \hat{\bm{P}})\times_1 \phi(\cdot)^\top\times_2 \psi(\cdot)^\top\Vert_{L^2(\xi\times\eta)} ^2 \\
	& + 32\Vert\bm{P}\times_1 \phi(\cdot)^\top\times_2 \psi(\cdot)^\top\times_3 (\bm{U}_3\bm{O} - \hat{\bm{U}}_3)^\top\Vert_{L^2(\xi\times\eta)} ^2\\
	= & 32\Vert (\bm{P} - \hat{\bm{P}})\times_1\bm{\Sigma}^{\frac{1}{2}}\Vert^2_F + 32\Vert\bm{P}\times_1 \bm{\Sigma}^{\frac{1}{2}}\times_3 (\bm{U}_3\bm{O} - \hat{\bm{U}}_3)^\top\Vert_F^2.
	\end{align*}
	Again by Part 3, Lemma 1 in \cite{cai2018rate}, we can choose $\bm{O}$ such that
	\begin{align*}
	\Vert \bm{U}_3\bm{O}^\top - \hat{\bm{U}_3}\Vert \leq \sqrt{2}\Vert \sin\Theta(\bm{U}_3, \hat{\bm{U}}_3)\Vert.
	\end{align*}
	With a similar argument in the proof of Theorem \ref{discrete}, we know when
	\begin{align*}
	\frac{n/ t_{mix}}{(\log(n/t_{mix}))^2}&\geq 1024\left(\Vert\bm{\Sigma}^{-1}\Vert^2_\sigma \bar{\mu} + \frac{K_{max}^2}{\bar{\mu}} + \frac{\kappa K_{max}^3}{\bar{\lambda}}\right)\left(\log\frac{t_{mix}}{\delta} + 8(d_S + d_A)\right),
	\end{align*}
	with probability at least $1-\delta$ we have
	\begin{align*}
	&32\Vert (\bm{P} - \hat{\bm{P}})\times_1\bm{\Sigma}^{\frac{1}{2}}\Vert^2_F + 32\Vert\bm{P}\times_1 \bm{\Sigma}^{\frac{1}{2}}\times_3 (\bm{U}_3\bm{O}^\top - \hat{\bm{U}}_3)^\top\Vert_F^2\\
	\leq &2^{24}\Vert\bm{\Sigma}^{-1}\Vert_\sigma\frac{rlm}{\max\{r, l, m\}}\left(1 + \frac{2\bar{\lambda}\Vert \bm{\Sigma}^{-1}\Vert_\sigma^2}{\sigma^2}\right)\frac{\bar{\lambda}(\log\frac{2t_{mix}}{\delta} + d_S + d_A)(\kappa + \bar{\mu}\Vert\bm{\Sigma}^{-1}\Vert_\sigma^2)(\log\frac{n}{t_{mix}})^2}{n/t_{mix}}.
	\end{align*}
	Now we choose sufficiently large $n$ such that 
	\begin{align*}
	32\Vert (\bm{P} - \hat{\bm{P}})\times_1\bm{\Sigma}^{\frac{1}{2}}\Vert^2_F + 32\Vert\bm{P}\times_1 \bm{\Sigma}^{\frac{1}{2}}\times_3 (\bm{U}_3\bm{O}^\top - \hat{\bm{U}}_3)^\top\Vert_F^2 < \Delta_1^2.
	\end{align*}
	Then for any $1\leq k\leq n_s, 1\leq l\leq n_a$ we always have
	\begin{align*}
	(\xi\times\eta)(S_{kl})\delta_{kl}^2\leq \sum_{k=1}^{n_s}\sum_{l=1}^{n_a}(\xi\times\eta)(S_{kl})\delta_{kl}^2 < \Delta_1^2 \leq (\xi\times\eta)(A_k\times B_l)\delta_{kl}^2,
	\end{align*}
	i.e., $(\xi\times\eta)(S_{kl}) \leq (\xi\times\eta)(A_k\times B_l)$, and $(\xi\times\eta)((A_k\times B_l)\setminus S_{kl})\neq \emptyset$. Now, for any $(i, j)\neq (k, l)$, we are going to show that $\forall (s, a)\in (A_i\times B_j)\setminus S_{ij}, (s', a')\in (A_k\times B_l)\setminus S_{kl}$, we always have $\bar{\bm{\Psi}}(s, a)\neq \bar{\bm{\Psi}}(s', a')$. Suppose the claim is not true, then we can find the corresponding $(s, a), (s', a')$, such that $\bar{\bm{\Psi}}(s, a)= \bar{\bm{\Psi}}(s', a')$. However, then we have
	\begin{align*}
	\max\{\delta_{ij}, \delta_{kl}\} \leq \Vert \bm{z}_{ij} - \bm{z}_{kl}\Vert \leq \Vert \bm{z}_{ij} - \bm{O}\bar{\bm{\Psi}}(s, a)\Vert + \Vert \bm{O}\bar{\bm{\Psi}}(s', a')- \bm{z}_{kl}\Vert \leq \frac{\delta_{ij}}{2} + \frac{\delta_{kl}}{2},
	\end{align*}
	which leads to a contradiction. 
	
	Furthermore, notice that for any $(s, a), (s', a') \in (A_i\times B_j)\setminus S_{ij}$, we always have $\bar{\bm{\Psi}}(s, a) = \bar{\bm{\Psi}}(s', a')$, otherwise $\bar{\bm{\Psi}}(s, a)$ will take more than $n_sn_a$ values. 
	
	The above two claims show that we can find two one-to-one mappings $\sigma_1: [n_s]\rightarrow[n_s], \sigma_2:[n_a]\rightarrow[n_a]$, such that for any $1\leq k\leq n_s, 1\leq l\leq n_a$, we have 
	\begin{align*}
	(A_k\times B_l)\setminus S_{kl} \subseteq \hat{A}_{\sigma_1(k)}\times \hat{B}_{\sigma_2(l)}. 
	\end{align*} 
	Without loss of generality we can assume that $\sigma_1(k) = k, \sigma_2(l) = l$, which is always possible after we rearrange the indexes of $\hat{A}_k, \hat{B}_l$. Then we have
	\begin{align*}
	(A_k\times B_l)\setminus S_{kl} \subseteq  (\hat{A}_k\times \hat{B}_l),
	\end{align*}
	which implies
	\begin{align*}
	(A_k\times B_l)\setminus (\hat{A}_k\times \hat{B}_l) \subseteq S_{kl}.
	\end{align*}
	Therefore, for $n$ sufficiently large, we have
	\begin{align*}
	&\sum_{ i=1}^{n_s}\sum_{j=1}^{n_a}\frac{(\xi\times\eta)((A_i\times B_j)\setminus (\hat{A}_i \times \hat{B}_j))}{(\xi\times\eta)(A_i\times B_j)} \\
	\leq &\sum_{ i=1}^{n_s}\sum_{j=1}^{n_a}\frac{(\xi\times\eta)(S_{ij})\delta_{ij}^2}{\Delta_1^2}\\
	\leq & 2^{24} \frac{1}{\Delta_1^2}\Vert\bm{\Sigma}^{-1}\Vert_\sigma\frac{rlm}{\max\{r, l, m\}}\left(1 + \frac{2\bar{\lambda}\Vert \bm{\Sigma}^{-1}\Vert_\sigma^2}{\sigma^2}\right)\frac{\bar{\lambda}(\log\frac{2t_{mix}}{\delta} + d_S + d_A)(\kappa + \bar{\mu}\Vert\bm{\Sigma}^{-1}\Vert_\sigma^2)(\log\frac{n}{t_{mix}})^2}{n/t_{mix}},
	\end{align*}
	which has finished the proof. 

\paragraph{Proof of Theorem \ref{partition2}}
	According to the result of Theorem \ref{partition}, we can find some mapping $\sigma_1, \sigma_2$, such that
	\begin{align*}
	&\sum_{i=1}^{n_s}\sum_{j=1}^{n_a}\frac{(\xi\times\eta)((A_i\times B_j)\setminus (\hat{A}_{\sigma_1(i)}\times \hat{B}_{\sigma_2(j)}))}{(\xi\times\eta)(A_i\times B_j)}
	\\
	\leq &2^{24} \frac{1}{\Delta_1^2}\Vert\bm{\Sigma}^{-1}\Vert_\sigma\frac{rlm}{\max\{r, l, m\}}\left(1 + \frac{2\bar{\lambda}\Vert \bm{\Sigma}^{-1}\Vert_\sigma^2}{\sigma^2}\right)\frac{\bar{\lambda}(\log\frac{2t_{mix}}{\delta} + d_S + d_A)(\kappa + \bar{\mu}\Vert\bm{\Sigma}^{-1}\Vert_\sigma^2)}{(n/t_{mix})(\log\frac{n}{t_{mix}})^{-2}} .
	\end{align*}
	Without loss of generality we assume that $\sigma_1(k) = k, \sigma_2(l) = l, \forall k\in[n_s], l\in[n_a]$. In this case, the result changes into 
	\begin{align*}
	&\sum_{i=1}^{n_s}\sum_{j=1}^{n_a}\frac{(\xi\times\eta)((A_i\times B_j)\setminus (\hat{A}_{i}\times \hat{B}_{j}))}{(\xi\times\eta)(A_i\times B_j)}
	\\
	\leq &2^{24} \frac{1}{\Delta_1^2}\Vert\bm{\Sigma}^{-1}\Vert_\sigma\frac{rlm}{\max\{r, l, m\}}\left(1 + \frac{2\bar{\lambda}\Vert \bm{\Sigma}^{-1}\Vert_\sigma^2}{\sigma^2}\right)\frac{\bar{\lambda}(\log\frac{2t_{mix}}{\delta} + d_S + d_A)(\kappa + \bar{\mu}\Vert\bm{\Sigma}^{-1}\Vert_\sigma^2)}{(n/t_{mix})(\log\frac{n}{t_{mix}})^{-2}} =:\varepsilon.
	\end{align*}
	Based on that, we want to further bound $\sum_{i=1}^{n_s}\sum_{j=1}^{n_a}\frac{(\xi\times\eta)((\hat{A}_i\times \hat{B}_j)\setminus (A_i \times B_j))}{(\xi\times\eta)(\hat{A}_i\times \hat{B}_j)}$. Notice that
	\begin{align*}
	\sum_{i=1}^{n_s}\sum_{j=1}^{n_a} (\xi\times\eta)((\hat{A}_i\times \hat{B}_j)\setminus (A_i\times B_j)) &= \sum_{i=1}^{n_s}\sum_{j=1}^{n_a} (\xi\times\eta)((A_i\times B_j)\setminus (\hat{A}_i\times \hat{B}_j)) \\
	&= \sum_{i=1}^{n_s}\sum_{j=1}^{n_a} \frac{(\xi\times\eta)((A_i\times B_j)\setminus (\hat{A}_i\times \hat{B}_j))}{(\xi\times\eta)(A_i\times B_j)}(\xi\times\eta)(A_i\times B_j) \\
	&\leq \bar{c}\sum_{i=1}^{n_s}\sum_{j=1}^{n_a} \frac{(\xi\times\eta)((A_i\times B_j)\setminus (\hat{A}_i\times \hat{B}_j))}{(\xi\times\eta)(A_i\times B_j)} \leq \bar{c}\varepsilon
	\end{align*}
	and 
	\begin{align*}
	(\xi\times\eta)(A_i\times B_j) \leq (\xi\times\eta)(\hat{A}_i\times \hat{B}_j) +  (\xi\times\eta)((A_i\times B_j)\setminus(\hat{A}_i\times \hat{B}_j)) \leq  (\xi\times\eta)(\hat{A}_i\times \hat{B}_j)+\bar{c}\varepsilon.
	\end{align*}
	So we have
	\begin{align*}
	\sum_{i=1}^{n_s}\sum_{j=1}^{n_a}\frac{(\xi\times\eta)((\hat{A}_i\times \hat{B}_j)\setminus (A_i \times B_j))}{(\xi\times\eta)(\hat{A}_i\times \hat{B}_j)}  &\leq \sum_{i=1}^{n_s}\sum_{j=1}^{n_a}\frac{(\xi\times\eta)((\hat{A}_i\times \hat{B}_j)\setminus (A_i \times B_j))}{(\xi\times\eta)(A_i\times B_j) - \bar{c}\varepsilon} \\
	&\leq \sum_{i=1}^{n_s}\sum_{j=1}^{n_a}\frac{(\xi\times\eta)((\hat{A}_i\times \hat{B}_j)\setminus (A_i \times B_j))}{\underline{c} - \bar{c}\varepsilon}\\
	& = \sum_{i=1}^{n_s}\sum_{j=1}^{n_a}\frac{(\xi\times\eta)((A_i\times B_j)\setminus (\hat{A}_i \times \hat{B}_j))}{\underline{c} - \bar{c}\varepsilon} \\
	& \leq \frac{\bar{c}}{\underline{c}-\bar{c}\varepsilon} \sum_{i=1}^{n_s}\sum_{j=1}^{n_a}\frac{(\xi\times\eta)((A_i\times B_j)\setminus (\hat{A}_i \times \hat{B}_j))}{(\xi\times\eta)(A_i\times B_j) } \\
	&\leq \frac{\bar{c}}{\underline{c}-\bar{c}\varepsilon}M( \{\hat{A}_i\}_{i=1}^{n_s}, \{\hat{B}_j\}_{j=1}^{n_a}).
	\end{align*}
	Therefore, when $n$ is sufficiently large such that $\bar{c}\varepsilon \leq \frac{1}{2}\underline{c}$, we have
	\begin{align*}
	\sum_{i=1}^{n_s}\sum_{j=1}^{n_a}\frac{(\xi\times\eta)((\hat{A}_i\times \hat{B}_j)\setminus (A_i \times B_j))}{(\xi\times\eta)(\hat{A}_i\times \hat{B}_j)} \leq 2\frac{\bar{c}}{\underline{c}}M(\{\hat{A}_i\}_{i=1}^{n_s}, \{\hat{B}_j\}_{j=1}^{n_a}).
	\end{align*}
	
	Now we are ready to prove the main result. By definition, the transition dynamic can be written as
	\begin{align*}
	\hat{p}(s'\vert s, a) = \sum_{i=1}^{n_s}\sum_{j=1}^{n_a}\sum_{k=1}^{n_s}\frac{1}{\xi(\hat{A}_k)}\hat{q}(k\vert i, j)\bm{1}_{s\in \hat{A}_i}\bm{1}_{a\in \hat{B}_j}\bm{1}_{s'\in \hat{A}_k}.
	\end{align*}
	Now, without loss of generality, we can assume that the groundtruth transition dynamic also takes the form
	\begin{align*}
	p(s'\vert s, a) = \sum_{i=1}^{n_s}\sum_{j=1}^{n_a}\sum_{k=1}^{n_s}\frac{1}{\xi(A_k)}q(k\vert i, j)\bm{1}_{s\in A_i}\bm{1}_{a\in B_j}\bm{1}_{s'\in A_k}.
	\end{align*}
	which is due to the fact that the policy and reward are the same in the same block, so for a general transition dynamic satisfying assumption \ref{block}, we can simply set $q(k\vert i, j) = \int_{A_k} q^*_{ij}(s')ds$, then the transition dynamic provided by the above formula will lead to exactly the same $H$-step value. 
	
	Because the infimum is taken over all possible $\hat{q}$, we can choose $\hat{q}$ exactly to be $q$, and get
	\begin{align*}
	\hat{p}^\pi(s', a'\vert s, a) = \sum_{i=1}^{n_s}\sum_{j=1}^{n_a}\sum_{k=1}^{n_s}\sum_{l=1}^{n_a}\frac{q^\pi(k, l\vert i, j)\bm{1}_{s\in \hat{A}_i}\bm{1}_{a\in \hat{B}_j}\bm{1}_{s'\in \hat{A}_k}\bm{1}_{a'\in \hat{B}_l}}{\xi(\hat{A}_k)\eta(\hat{B}_l)},
	\end{align*}
	where
	\begin{align*}
	q^\pi(k, l\vert i, j) = \pi(l\vert k)q(k\vert i, j).
	\end{align*}
	For each single path $(s_1, a_1), (s_2, a_2), \cdots, (s_H, a_H)$, consider a series of auxiliary random variables $T_1, T_2,\cdots, T_H$, which are defined inductively by 
	\begin{align*}
	T_0 = 0, T_h = \max\left\{T_{h-1}, \bm{1}_{(s_h, a_h)\not\in \bigcup_{i=1}^{n_s}(A_{i}\cap \hat{A}_i)\times \bigcup_{j=1}^{n_a}(B_j\cap \hat{B}_j)}\right\}.
	\end{align*}
	Then for each reward function $r_h, h\in[H]$ and any initial distribution $\mu$, we have
	\begin{align*}
	\mathbb{E}^\pi_\mu[r_h(s_h, a_h)] = \mathbb{E}^\pi_\mu[r_h(s_h, a_h)\bm{1}_{T_h = 0}] + \mathbb{E}^\pi_\mu[r_h(s_h, a_h)\bm{1}_{T_h = 1}], \\
	\hat{\mathbb{E}}^\pi_\mu[r_h(s_h, a_h)] = \hat{\mathbb{E}}^\pi_\mu[r_h(s_h, a_h)\bm{1}_{T_h = 0}] + \hat{\mathbb{E}}^\pi_\mu[r_h(s_h, a_h)\bm{1}_{T_h = 1}]
	\end{align*}
	and
	\begin{align*}
	& \vert \mathbb{E}^\pi_\mu[r_h(s_h, a_h)] -  \hat{\mathbb{E}}^\pi_\mu[r_h(s_h, a_h)]\vert \\
	\leq & \vert \mathbb{E}^\pi_\mu[r_h(s_h, a_h)\bm{1}_{T_h = 0}] -  \hat{\mathbb{E}}^\pi_\mu[r_h(s_h, a_h)\bm{1}_{T_h = 0}]\vert + (\mathbb{P}^\pi_\mu(T_h = 1) + \hat{\mathbb{P}}^\pi_\mu(T_h = 1)).
	\end{align*}
	For the second term, note that
	\begin{align*}
	&\mathbb{P}^\pi_\mu(T_{h+1} = 1) \\
	=& \mathbb{P}^\pi_\mu(T_{h+1} = 1\vert T_h = 1)\mathbb{P}^\pi_\mu(T_{h} = 1) + \mathbb{P}^\pi_\mu(T_{h+1} = 1\vert T_h = 0)\mathbb{P}^\pi_\mu(T_{h} = 0)   \\
	=&  \mathbb{P}^\pi_\mu(T_{h} = 1) + \sum_{i=1}^{n_s}\sum_{j=1}^{n_a}\mathbb{P}^\pi_\mu(s_{h+1}\in A_i, a_{h+1}\in B_j, T_{h+1}=1\vert T_h = 0)\\
	=&\mathbb{P}^\pi_\mu(T_{h} = 1)  + \sum_{i=1}^{n_s}\sum_{j=1}^{n_a}\mathbb{P}^\pi_\mu((s_{h+1}, a_{h+1})\in (A_i\times B_j)\setminus (\hat{A}_i\times \hat{B}_j)\vert s_{h+1}\in A_i, a_{h+1}\in B_j)\\
	& \quad\quad\quad \cdot \mathbb{P}^\pi_\mu(s_{h+1}\in A_i, a_{h+1}\in B_j\vert T_h = 0)\\
	=&  \mathbb{P}^\pi_\mu(T_{h} = 1) + \sum_{i=1}^{n_s}\sum_{j=1}^{n_a}\frac{(\xi\times\eta)((A_i\times B_j)\setminus (\hat{A}_i \times \hat{B}_j))}{(\xi\times\eta)(A_i\times B_j)}\mathbb{P}^\pi_{\mu}(s_{h+1}\in A_i, a_{h+1}\in B_j\vert T_h = 0)\\
	\leq& \mathbb{P}^\pi_\mu(T_{h} = 1) + \max_{1\leq i\leq n_s, 1\leq j\leq n_a}\frac{(\xi\times\eta)((A_i\times B_j)\setminus (\hat{A}_i \times \hat{B}_j))}{(\xi\times\eta)(A_i\times B_j)},
	\end{align*}
	which implies 
	\begin{align*}
	\mathbb{P}^\pi_\mu(T_{h} = 1) \leq H\max_{1\leq i\leq n_s, 1\leq j\leq n_a}\frac{(\xi\times\eta)((A_i\times B_j)\setminus (\hat{A}_i \times \hat{B}_j))}{(\xi\times\eta)(A_i\times B_j)}.
	\end{align*}
	Similarly, we have
	\begin{align*}
	\hat{\mathbb{P}}^\pi_\mu(T_{h} = 1) \leq H\max_{1\leq i\leq n_s, 1\leq j\leq n_a}\frac{(\xi\times\eta)((\hat{A}_i\times \hat{B}_j)\setminus (A_i \times B_j))}{(\xi\times\eta)(\hat{A}_i\times \hat{B}_j)}.
	\end{align*}
	For the first term, we consider a discrete MDP over $(A_i\cap \hat{A}_i) \times (B_j\cap\hat{B}_j), 1\leq i\leq n_s, 1\leq j\leq n_a$ plus an absorbing state $C_0$, and two corresponding transition probabilities
	\begin{align*}
	&\mathbb{P}((A_k\cap \hat{A}_k)\times (B_l\cap\hat{B}_l)\vert (A_i\cap \hat{A}_i)\times(B_j\cap \hat{B}_j)) = q^\pi(k, l\vert i, j)\frac{(\xi\times\eta)((A_k\cap \hat{A}_k)\times(B_l\cap \hat{B}_l))}{(\xi\times\eta)(A_k\times B_l)},\\
	&\mathbb{P}(C_0\vert (A_i\cap \hat{A}_i)\times(B_j\cap \hat{B}_j)) = 1 - \sum_{k=1}^{n_s}\sum_{l=1}^{n_a}\mathbb{P}((A_k\cap \hat{A}_k)\times (B_l\cap\hat{B}_l)\vert (A_i\cap \hat{A}_i)\times(B_j\cap \hat{B}_j))\\
	&\hat{\mathbb{P}}((A_k\cap \hat{A}_k)\times (B_l\cap\hat{B}_l)\vert (A_i\cap \hat{A}_i)\times(B_j\cap \hat{B}_j)) = q^\pi(k, l\vert i, j)\frac{(\xi\times\eta)((A_k\cap \hat{A}_k)\times(B_l\cap \hat{B}_l))}{(\xi\times\eta)(\hat{A}_k\times \hat{B}_l)},\\
	&\hat{\mathbb{P}}(C_0\vert (A_i\cap \hat{A}_i)\times(B_j\cap \hat{B}_j)) = 1 - \sum_{k=1}^{n_s}\sum_{l=1}^{n_a}\hat{\mathbb{P}}((A_k\cap \hat{A}_k)\times (B_l\cap\hat{B}_l)\vert (A_i\cap \hat{A}_i)\times(B_j\cap \hat{B}_j)).
	\end{align*}
	We use $\bm{P}, \hat{\bm{P}}$ to denote the two corresponding transition matrices restricted on $(A_i\cap \hat{A}_i) \times (B_j\cap\hat{B}_j), 1\leq i\leq n_s, 1\leq j\leq n_a$ (i.e., the entries related with $C_0$ are not included, so $\bm{P}, \hat{\bm{P}}$ are sub-matrices of two stochastic matrices). For two different distributions $\mu, \tilde{\mu}$ over $(A_i\cap \hat{A}_i) \times (B_j\cap\hat{B}_j), 1\leq i\leq n_s, 1\leq j\leq n_a$, we have 
	\begin{align*}
	\Vert \mu\bm{P} - \tilde{\mu}\hat{\bm{P}}\Vert_1 &\leq \Vert \mu\bm{P} - \mu\hat{\bm{P}} \Vert_1 + \Vert\mu\hat{\bm{P}} - \tilde{\mu}\hat{\bm{P}}\Vert_1\\
	&\leq \Vert \mu\Vert_1 \max_{(i, j),(k,l)} \vert(\bm{P} - \hat{\bm{P}})_{(i,j),(k,l)}\vert + \Vert \mu - \tilde{\mu}\Vert_1\max_{(i,j), (k, l)}\vert \hat{\bm{P}}_{(i,j),(k,l)}\vert\\
	&\leq \max_{(i, j), (k,l)} \vert(\bm{P} - \hat{\bm{P}})_{(i,j),(k,l)}\vert + \Vert \mu - \tilde{\mu}\Vert_1.
	\end{align*}
	Note that
	\begin{align*}
	(\bm{P} - \hat{\bm{P}})_{(i, j),(k, l)} = \left(\frac{(\xi\times\eta)((A_k\cap \hat{A}_k)\times(B_l\cap\hat{B}_l))}{(\xi\times\eta)(A_k\times B_l)} - \frac{(\xi\times\eta)((A_k\cap \hat{A}_k)\times(B_l\cap\hat{B}_l))}{(\xi\times\eta)(\hat{A}_k\times \hat{B}_l)}\right)q^\pi(k, l\vert i, j),
	\end{align*}
	so we have
	\begin{align*}
	& \max_{(i, j), (k,l)} \vert(\bm{P} - \hat{\bm{P}})_{(i,j),(k,l)}\vert \\ \leq & \max_{1\leq i\leq n_s, 1\leq j\leq n_a} \left\vert \frac{(\xi\times\eta)((A_i\cap \hat{A}_i)\times(B_j\cap\hat{B}_j))}{(\xi\times\eta)(A_i\times B_j)} - \frac{(\xi\times\eta)((A_i\cap \hat{A}_i)\times(B_j\cap\hat{B}_j))}{(\xi\times\eta)(\hat{A}_i\times \hat{B}_j)}\right\vert \\
	\leq & \max_{1\leq i\leq n_s, 1\leq j\leq n_a}\frac{(\xi\times\eta)((A_i\times B_j)\setminus (\hat{A}_i \times \hat{B}_j))}{(\xi\times\eta)(A_i\times B_j)}+ \max_{1\leq i\leq n_s, 1\leq j\leq n_a}\frac{(\xi\times\eta)((\hat{A}_i\times \hat{B}_j)\setminus (A_i \times B_j))}{(\xi\times\eta)(\hat{A}_i\times \hat{B}_j)}.
	\end{align*}
	So we get 
	\begin{align*}
	&\Vert \mu\bm{P}^h - \mu\hat{\bm{P}}^h\Vert_1 \\
	\leq & \max_{\begin{subarray}{c} 1\leq i\leq n_s \\ 1\leq j\leq n_a \end{subarray}}\frac{(\xi\times\eta)((A_i\times B_j)\setminus (\hat{A}_i \times \hat{B}_j))}{(\xi\times\eta)(A_i\times B_j)}+ \max_{\begin{subarray}{c} 1\leq i\leq n_s \\ 1\leq j\leq n_a \end{subarray}} \frac{(\xi\times\eta)((\hat{A}_i\times \hat{B}_j)\setminus (A_i \times B_j))}{(\xi\times\eta)(\hat{A}_i\times \hat{B}_j)}\\
	& + \Vert \mu\bm{P}^{h-1} - \mu\hat{\bm{P}}^{h-1}\Vert_1\\
	\leq &\cdots\\
	\leq &H \Bigg(\max_{1\leq i\leq n_s, 1\leq j\leq n_a}\frac{(\xi\times\eta)((A_i\times B_j)\setminus (\hat{A}_i \times \hat{B}_j))}{(\xi\times\eta)(A_i\times B_j)}\\
	& \quad + \max_{1\leq i\leq n_s, 1\leq j\leq n_a}\frac{(\xi\times\eta)((\hat{A}_i\times \hat{B}_j)\setminus (A_i \times B_j))}{(\xi\times\eta)(\hat{A}_i\times\hat{B}_j)}\Bigg),
	\end{align*}
	Note that the transition dynamic of the original state-action space can be embedded into the discrete MDP mentioned above according to the following mapping: Denote $\tilde{s}_h$ as the state of the discrete MDP at step $h$, then we let $\tilde{s}_h = C_0$ if $T_h = 1$, and let $\tilde{s}_h = (A_i\times B_j)\cap (\hat{A}_i\times \hat{B}_j)$ if $T_h = 0$ and $(s_h, a_h) \in (A_i\times B_j)\cap (\hat{A}_i\times \hat{B}_j)$. Because the reward is only related with blocks and is bounded between $[0, 1]$, so we have
	\begin{align*}
	&\vert \mathbb{E}^\pi_\mu[r_h(S_h, A_h)\bm{1}_{T_h = 0}] -  \hat{\mathbb{E}}^\pi_\mu[r_h(S_h, A_h)\bm{1}_{T_h = 0}]\vert \\
	\leq& \Vert \mu\bm{P}^h - \mu\hat{\bm{P}}^h\Vert_1\\
	\leq& H \left( \max_{1\leq i\leq n_s, 1\leq j\leq n_a}\frac{(\xi\times\eta)((A_i\times B_j)\setminus (\hat{A}_i \times \hat{B}_j))}{(\xi\times\eta)(A_i\times B_j)}+ \max_{1\leq i\leq n_s, 1\leq j\leq n_a}\frac{(\xi\times\eta)((\hat{A}_i\times \hat{B}_j)\setminus (A_i \times B_j))}{(\xi\times\eta)(\hat{A}_i\times \hat{B}_j)}\right).
	\end{align*}
	In summary, we get
	\begin{align*}
	& \vert \mathbb{E}^\pi[r_h(S_h, A_h)] -  \hat{\mathbb{E}}^\pi[r_h(S_h, A_h)]\vert \\
	\leq& 2H \left( \max_{1\leq i\leq n_s, 1\leq j\leq n_a}\frac{(\xi\times\eta)((A_i\times B_j)\setminus (\hat{A}_i \times \hat{B}_j))}{(\xi\times\eta)(A_i\times B_j)}+ \max_{1\leq i\leq n_s, 1\leq j\leq n_a}\frac{(\xi\times\eta)((\hat{A}_i\times \hat{B}_j)\setminus (A_i \times B_j))}{(\xi\times\eta)(\hat{A}_i\times \hat{B}_j)}\right)\\
	\leq &4H\frac{\bar{c}}{\underline{c}}M(\{\hat{A}_i\}_{i=1}^{n_s}, \{\hat{B}_j\}_{j=1}^{n_a})
	\end{align*}
	and
	\begin{align*}
	&\left\vert \sum_{h=1}^H\mathbb{E}^\pi[r_h(S_h, A_h)] -  \sum_{h=1}^H\hat{\mathbb{E}}^\pi[r_h(S_h, A_h)]\right\vert \\
	\leq & \sum_{h=1}^H\vert\mathbb{E}^\pi[r_h(S_h, A_h)] - \hat{\mathbb{E}}^\pi[r_h(S_h, A_h)]\vert\\
	\leq & 4H^2\frac{\bar{c}}{\underline{c}}M(\{\hat{A}_i\}_{i=1}^{n_s}, \{\hat{B}_j\}_{j=1}^{n_a})
	\end{align*}
	which finishes the proof. 
\end{document}